\documentclass[10pt,journal,compsoc]{IEEEtran}
\usepackage{amsmath,amsfonts}
\usepackage{algorithmic}
\usepackage{algorithm}
\usepackage{array}
\usepackage[caption=false,font=normalsize,labelfont=sf,textfont=sf]{subfig}
\usepackage{textcomp}
\usepackage{stfloats}
\usepackage{url}
\usepackage{verbatim}
\usepackage{graphicx}
\usepackage{cite}
\usepackage{booktabs}
\usepackage{graphicx}
\usepackage{multirow}
\usepackage{tablefootnote}
\usepackage{xcolor}
\usepackage{balance}
\usepackage{hyperref}
\usepackage{threeparttable}
\newcommand{\RE}{\textcolor{black}}

\begin{document}

\title{Unifying Large Language Models and Knowledge Graphs: A Roadmap}

\author{Shirui Pan, \emph{Senior Member, IEEE},~Linhao~Luo,\\~Yufei~Wang,~Chen~Chen,~Jiapu~Wang,~Xindong~Wu, \textit{Fellow, IEEE}
\IEEEcompsocitemizethanks{
\IEEEcompsocthanksitem Shirui Pan is with the School of Information and Communication Technology and Institute for Integrated and Intelligent Systems (IIIS), Griffith University, Queensland, Australia. 
Email: s.pan@griffith.edu.au;
\IEEEcompsocthanksitem Linhao Luo and Yufei~Wang are with the Department of Data Science and AI, Monash University, Melbourne, Australia. E-mail: linhao.luo@monash.edu, garyyufei@gmail.com.
\IEEEcompsocthanksitem Chen Chen is with the Nanyang Technological University, Singapore. E-mail: s190009@ntu.edu.sg.
\IEEEcompsocthanksitem Jiapu Wang is with the Faculty of Information Technology, Beijing University of Technology, Beijing, China. E-mail: jpwang@emails.bjut.edu.cn.
\IEEEcompsocthanksitem Xindong Wu is with the Key Laboratory of Knowledge Engineering with Big Data (the Ministry of Education of China), Hefei University of Technology, Hefei, China, and also with the Research Center for Knowledge Engineering, Zhejiang Lab, Hangzhou, China.
Email: xwu@hfut.edu.cn.
\IEEEcompsocthanksitem Shirui Pan and Linhao Luo contributed equally to this work. \IEEEcompsocthanksitem Corresponding Author: Xindong Wu.
}
}

\markboth{Journal of \LaTeX\ Class Files,~Vol.~??, No.~??, MONTH~20YY}%
{Shell \MakeLowercase{\textit{et al.}}: A Sample Article Using IEEEtran.cls for IEEE Journals}

\IEEEpubid{0000--0000/00\$00.00~\copyright~2023 IEEE}

\IEEEtitleabstractindextext{
\begin{abstract}
    Large language models (LLMs), such as ChatGPT and GPT4, are making new waves in the field of natural language processing and artificial intelligence, due to their emergent ability and generalizability. However, LLMs are black-box models, which often fall short of capturing and accessing factual knowledge. In contrast, Knowledge Graphs (KGs), Wikipedia and Huapu for example, are structured knowledge models that explicitly store rich factual knowledge. KGs can enhance LLMs by providing external knowledge for inference and interpretability. Meanwhile, KGs are difficult to construct and evolve by nature, which challenges the existing methods in KGs to generate new facts and represent unseen knowledge. Therefore, it is complementary to unify LLMs and KGs together and simultaneously leverage their advantages. 
    In this article, we present a forward-looking roadmap for the unification of LLMs and KGs. Our roadmap consists of three general frameworks, namely, \textit{1) KG-enhanced LLMs,} which incorporate KGs during the pre-training and inference phases of LLMs, or for the purpose of enhancing understanding of the knowledge learned by LLMs; \textit{2) LLM-augmented KGs,} that leverage LLMs for different KG tasks such as embedding, completion, construction, graph-to-text generation, and question answering; and \textit{3) Synergized LLMs + KGs}, in which LLMs and KGs play equal roles and work in a mutually beneficial way to enhance both LLMs and KGs for bidirectional reasoning driven by both data and knowledge. We review and summarize existing efforts within these three frameworks in our roadmap and pinpoint their future research directions.
\end{abstract}

\begin{IEEEkeywords} Natural Language Processing, Large Language Models, Generative Pre-Training, Knowledge Graphs, Roadmap, Bidirectional Reasoning.
\end{IEEEkeywords}
}

\maketitle

\section{Introduction}
Large language models (LLMs)\footnote{LLMs are also known as pre-trained language models (PLMs).} (e.g., BERT \cite{devlin2018bert}, RoBERTA \cite{liu2019roberta}, and T5 \cite{raffel2020exploring}), pre-trained on the large-scale corpus, have shown great performance in various natural language processing (NLP) tasks, such as question answering \cite{su2019generalizing}, machine translation \cite{lewis2020bart}, and text generation \cite{li2021pretrained}. Recently, the dramatically increasing model size further enables the LLMs with the emergent ability \cite{weiemergent}, paving the road for applying LLMs as Artificial General Intelligence (AGI). Advanced LLMs like ChatGPT\footnote{\url{https://openai.com/blog/chatgpt}} and PaLM2\footnote{\url{https://ai.google/discover/palm2}}, with billions of parameters, exhibit great potential in many complex practical tasks, such as education \cite{malinka2023educational}, code generation \cite{li2022cctest} and recommendation \cite{liu2023chatgpt}.

\begin{figure}
    \includegraphics[width=\columnwidth]{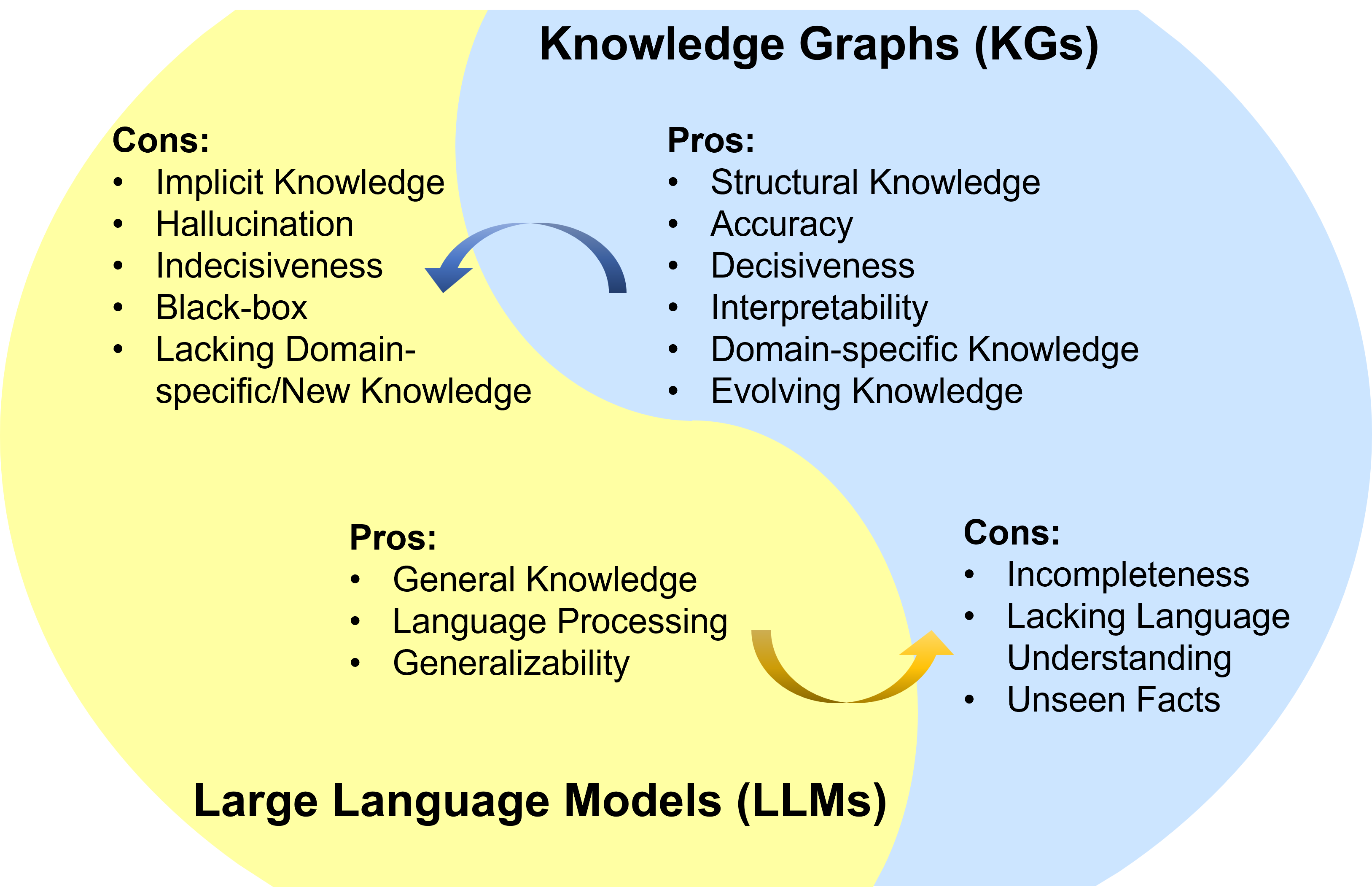}
    \caption{Summarization of the pros and cons for LLMs and KGs. LLM pros: \emph{General Knowledge} \cite{zhao2023survey}, \emph{Language Processing} \cite{qiu2020pre}, \emph{Generalizability} \cite{yang2023harnessing}; LLM cons: \emph{Implicit Knowledge} \cite{petroni2019language}, \emph{Hallucination} \cite{ji2023survey}, \emph{Indecisiveness} \cite{zhang2022survey}, \emph{Black-box} \cite{danilevsky2020survey}, \emph{Lacking Domain-specific/New Knowledge} \cite{wang2023robustness}. KG pros: \emph{Structural Knowledge} \cite{ji2021survey}, \emph{Accuracy} \cite{vrandevcic2014wikidata}, \emph{Decisiveness} \cite{hu2018state}, \emph{Interpretability} \cite{zhang2021neural}, \emph{Domain-specific Knowledge} \cite{abu2021domain}, \emph{Evolving Knowledge} \cite{Mitchell2015NeverEndingL}; KG cons: \emph{Incompleteness} \cite{zhong2023comprehensive}, \emph{Lacking Language Understanding} \cite{yao2019kg}, \emph{Unseen Facts} \cite{luo2023normalizing}. Pros. and Cons. are selected based on their representativeness. Detailed discussion can be found in \emph{Appendix \ref{app:pros_and_cons}.}}\label{fig:LLM_vs_kg}
\end{figure}

Despite their success in many applications, LLMs have been criticized for their lack of factual knowledge.  Specifically, LLMs memorize facts and knowledge contained in the training corpus \cite{petroni2019language}. However, further studies reveal that LLMs are not able to recall facts and often experience hallucinations by generating statements that are factually incorrect \cite{bang2023multitask,ji2023survey}. For example, LLMs might say ``Einstein discovered gravity in 1687'' when asked, ``When did Einstein discover gravity?'', which contradicts the fact that Isaac Newton formulated the gravitational theory. This issue severely impairs the trustworthiness of LLMs.

As black-box models, LLMs are also criticized for their lack of interpretability. LLMs represent knowledge implicitly in their parameters. It is difficult to interpret or validate the knowledge obtained by LLMs. Moreover, LLMs perform reasoning by a probability model, which is an indecisive process \cite{zhang2022survey}.  The specific patterns and functions LLMs used to arrive at predictions or decisions are not directly accessible or explainable to humans \cite{danilevsky2020survey}. Even though some LLMs are equipped to explain their predictions by applying chain-of-thought \cite{wang2022self}, their reasoning explanations also suffer from the hallucination issue \cite{golovneva2022roscoe}. This severely impairs the application of LLMs in high-stakes scenarios, such as medical diagnosis and legal judgment. For instance, in a medical diagnosis scenario, LLMs may incorrectly diagnose a disease and provide explanations that contradict medical commonsense. This raises another issue that LLMs trained on general corpus might not be able to generalize well to specific domains or new knowledge due to the lack of domain-specific knowledge or new training data \cite{wang2023robustness}.

To address the above issues, a potential solution is to incorporate knowledge graphs (KGs) into LLMs. Knowledge graphs (KGs), storing enormous facts in the way of triples, i.e., $(head~entity, relation, tail~entity)$, are a structured and decisive manner of knowledge representation (e.g., Wikidata \cite{vrandevcic2014wikidata}, YAGO \cite{suchanek2007yago}, and NELL \cite{carlson2010toward}). KGs are crucial for various applications as they offer accurate explicit knowledge \cite{ji2021survey}. 
Besides, they are renowned for their symbolic reasoning ability \cite{zhang2021neural}, which generates interpretable results. 
 KGs can also actively evolve with new knowledge continuously added in \cite{Mitchell2015NeverEndingL}. Additionally, experts can construct domain-specific KGs to provide precise and dependable domain-specific knowledge \cite{abu2021domain}. 
 
 Nevertheless, KGs are difficult to construct \cite{zhong2023comprehensive}, and current approaches in KGs \cite{bordes2013translating,wan2021reasoning,luo2023normalizing} are inadequate in handling the incomplete and dynamically changing nature of real-world KGs. These approaches fail to effectively model unseen entities and represent new facts. In addition, they often ignore the abundant textual information in KGs. Moreover, existing methods in KGs are often customized for specific KGs or tasks, which are not generalizable enough. Therefore, it is also necessary to utilize LLMs to address the challenges faced in KGs. We summarize the pros and cons of LLMs and KGs in Fig. \ref{fig:LLM_vs_kg}, respectively.

Recently, the possibility of unifying LLMs with KGs has attracted increasing attention from researchers and practitioners. LLMs and KGs are inherently interconnected and can mutually enhance each other. In \emph{KG-enhanced LLMs}, KGs can not only be incorporated into the pre-training and inference stages of LLMs to provide external knowledge \cite{zhang-etal-2019-ernie,DBLP:conf/aaai/LiuZ0WJD020,DBLP:conf/aaai/LiuW0PY21}, but also used for analyzing LLMs and providing interpretability \cite{petroni2019language,lin-etal-2019-kagnet,dai2021knowledge}. In \emph{LLM-augmented KGs}, LLMs have been used in various KG-related tasks, e.g., KG embedding \cite{wang-etal-2021-kepler}, KG completion \cite{yao2019kg}, KG construction \cite{melnyk2021grapher}, KG-to-text generation \cite{ke-etal-2021-jointgt}, and KGQA \cite{jiang2023unikgqa}, to improve the performance and facilitate the application of KGs. In \emph{Synergized LLM + KG}, researchers marries the merits of LLMs and KGs to mutually enhance performance in knowledge representation \cite{yasunaga2022deep} and reasoning \cite{choudhary2023complex,wang2023unifying}.
Although there are some surveys on knowledge-enhanced LLMs \cite{zhen2022survey,wei2021knowledge,yin2022survey}, which mainly focus on using KGs as an external knowledge to enhance LLMs, they ignore other possibilities of integrating KGs for LLMs and the potential role of LLMs in KG applications.

In this article, we present a forward-looking roadmap for unifying both LLMs and KGs, to leverage their respective strengths and overcome the limitations of each approach, for various downstream tasks. We propose detailed categorization, conduct comprehensive reviews, and pinpoint emerging directions in these fast-growing fields. 
Our main contributions are summarized as follows:
\begin{enumerate}
    \item \textbf{Roadmap.} We present a forward-looking roadmap for integrating LLMs and KGs. Our roadmap, consisting of three general frameworks to unify LLMs and KGs, namely, \textit{KG-enhanced LLMs}, \textit{LLM-augmented KGs}, and \textit{Synergized LLMs + KGs}, provides guidelines for the unification of these two distinct but complementary technologies.
    \item \textbf{Categorization and review.} For each integration framework of our roadmap, we present a detailed categorization and novel taxonomies of research on unifying LLMs and KGs. 
    In each category, we review the research from the perspectives of different integration strategies and tasks, which provides more insights into each framework.
    \item \textbf{Coverage of emerging advances.} We cover the advanced techniques in both LLMs and KGs. We include the discussion of state-of-the-art LLMs like ChatGPT and GPT-4 as well as the novel KGs e.g., multi-modal knowledge graphs.
    \item \textbf{Summary of challenges and future directions.} We highlight the challenges in existing research and present several promising future research directions.
\end{enumerate}

The rest of this article is organized as follows. Section \ref{sec:background} first explains the background of LLMs and KGs. Section \ref{sec:roadmap}  introduces the roadmap and the overall categorization of this article. Section \ref{sec:KG_for_LLM} presents the different KGs-enhanced LLM approaches. Section \ref{sec:LLM_for_KG} describes the possible LLM-augmented KG methods. Section \ref{sec:unification_LLM_kg} shows the approaches of synergizing LLMs and KGs. Section \ref{sec:challenge} discusses the challenges and future research directions. Finally, Section \ref{sec:conclusion} concludes this paper.

\section{Background}\label{sec:background}
In this section, we will first briefly introduce a few representative large language models (LLMs) and discuss the prompt engineering that efficiently uses LLMs for varieties of applications. Then, we illustrate the concept of knowledge graphs (KGs) and present different categories of KGs.
\begin{figure*}
    \centering
    \includegraphics[width=.8\linewidth]{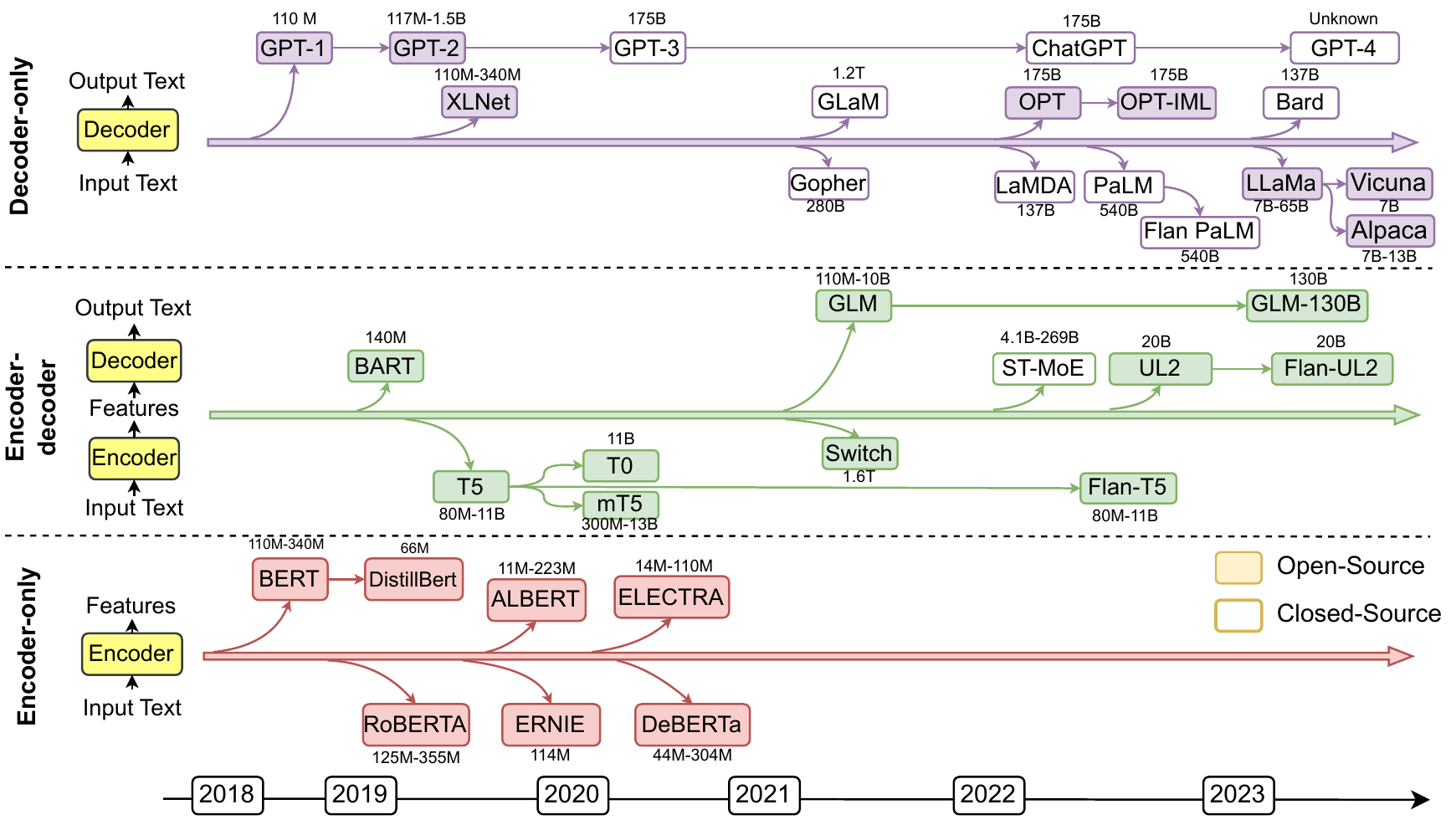}
    \caption{Representative large language models (LLMs) in recent years. Open-source models are represented by solid squares, while closed source models are represented by hollow squares.}
    \label{tab:LLM_types}
\end{figure*}
\begin{figure}[h]
    \centering
    \includegraphics[width=.8\columnwidth]{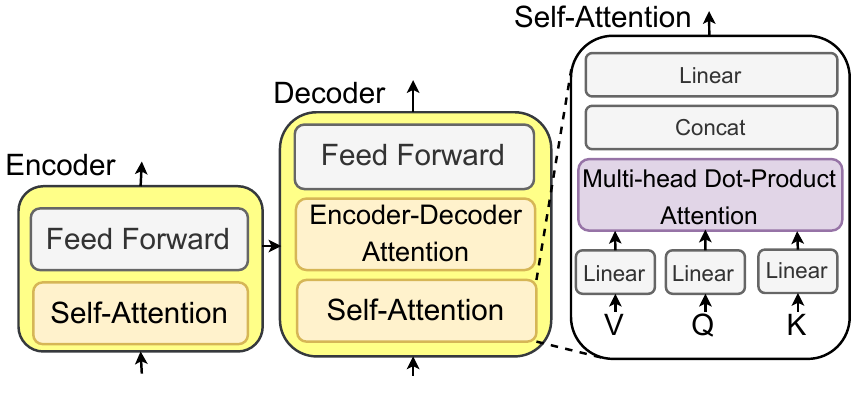}
    \caption{An illustration of the Transformer-based LLMs with self-attention mechanism.}
    \label{fig:self-attn}
\end{figure}

\subsection{Large Language models (LLMs)}
Large language models (LLMs) pre-trained on large-scale corpus have shown great potential in various NLP tasks \cite{yang2023harnessing}. As shown in Fig. \ref{fig:self-attn}, most LLMs derive from the Transformer design \cite{vaswani2017attention}, which contains the encoder and decoder modules empowered by a self-attention mechanism. Based on the architecture structure, LLMs can be categorized into three groups: \emph{1)~encoder-only LLMs},~\emph{2)~encoder-decoder LLMs},~and \emph{3)~decoder-only LLMs}. As shown in Fig. \ref{tab:LLM_types}, we summarize several representative LLMs with different model architectures, model sizes, and open-source availabilities.

\subsubsection{Encoder-only LLMs.}
Encoder-only large language models only use the encoder to encode the sentence and understand the relationships between words. The common training paradigm for these model is to predict the mask words in an input sentence. This method is unsupervised and can be trained on the large-scale corpus. Encoder-only LLMs like BERT \cite{devlin2018bert}, ALBERT \cite{lanalbert}, RoBERTa \cite{liu2019roberta}, and ELECTRA \cite{clark2020electra} require adding an extra prediction head to resolve downstream tasks. These models are most effective for tasks that require understanding the entire sentence, such as text classification \cite{yao2019kg} and named entity recognition \cite{hakala2019biomedical}.

\subsubsection{Encoder-decoder LLMs.}
Encoder-decoder large language models adopt both the encoder and decoder module. The encoder module is responsible for encoding the input sentence into a hidden-space, and the decoder is used to generate the target output text. The training strategies in encoder-decoder LLMs can be more flexible. For example, T5 \cite{raffel2020exploring} is pre-trained by masking and predicting spans of masking words. UL2 \cite{tay2022ul2} unifies several training targets such as different masking spans and masking frequencies. Encoder-decoder LLMs (e.g., T0 \cite{sanhmultitask}, ST-MoE \cite{zoph2022st}, and GLM-130B \cite{zeng2023glm-130b}) are able to directly resolve tasks that generate sentences based on some context, such as summariaztion, translation, and question answering \cite{xue2021mt5}.

\subsubsection{Decoder-only LLMs.}
Decoder-only large language models only adopt the decoder module to generate target output text. The training paradigm for these models is to predict the next word in the sentence. Large-scale decoder-only LLMs can generally perform downstream tasks from a few examples or simple instructions, without adding prediction heads or finetuning \cite{brown2020language}. Many state-of-the-art LLMs (e.g., Chat-GPT \cite{ouyang2022training} and GPT-4\footnote{\url{https://openai.com/product/gpt-4}}) follow the decoder-only architecture. However, since these models are closed-source, it is challenging for academic researchers to conduct further research. Recently, Alpaca\footnote{\url{https://github.com/tatsu-lab/stanford_alpaca}} and Vicuna\footnote{\url{https://lmsys.org/blog/2023-03-30-vicuna/}} are released as open-source decoder-only LLMs. These models are finetuned based on LLaMA \cite{touvron2023llama} and achieve comparable performance with ChatGPT and GPT-4.

\subsubsection{Prompt Engineering}
\begin{figure}
    \centering
    \includegraphics[width=.85\columnwidth]{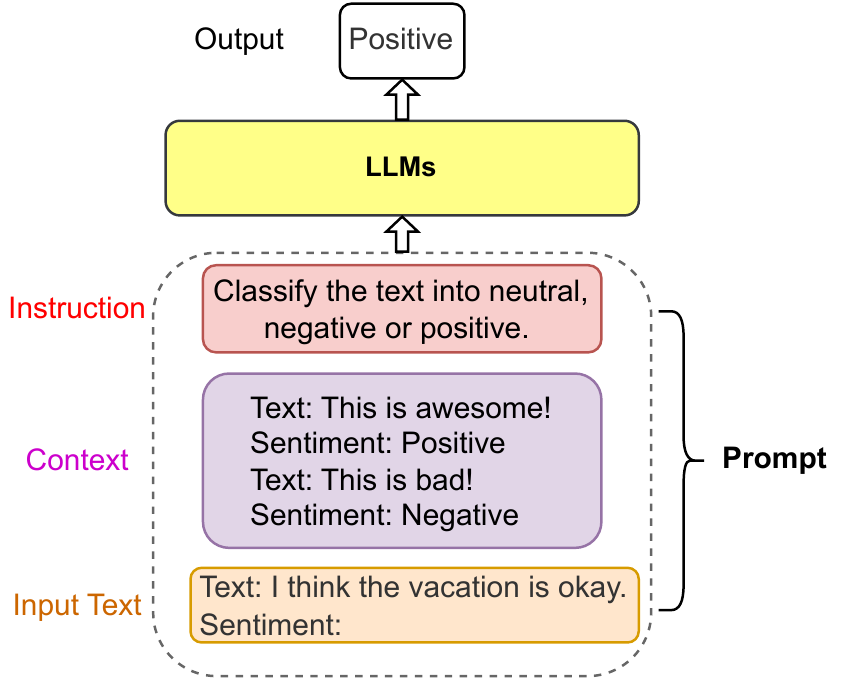}
    \caption{An example of sentiment classification prompt.}
    \label{fig:prompts}
\end{figure}

Prompt engineering is a novel field that focuses on creating and refining prompts to maximize the effectiveness of large language models (LLMs) across various applications and research areas \cite{Saravia_Prompt_Engineering_Guide_2022}.
As shown in Fig. \ref{fig:prompts}, a prompt is a sequence of natural language inputs for LLMs that are specified for the task, such as sentiment classification. A prompt could contain several elements, i.e., \emph{1)~Instruction}, \emph{2)~Context}, and \emph{3)~Input Text}. \emph{Instruction} is a short sentence that instructs the model to perform a specific task. \emph{Context} provides the context for the input text or few-shot examples. \emph{Input Text} is the text that needs to be processed by the model.

Prompt engineering seeks to improve the capacity of large large language models (e.g., ChatGPT) in diverse complex tasks such as question answering, sentiment classification, and common sense reasoning. Chain-of-thought (CoT) prompt \cite{weichain} enables complex reasoning capabilities through intermediate reasoning steps. 
Prompt engineering also enables the integration of structural data like knowledge graphs (KGs) into LLMs. Li et al. \cite{li2023graph} simply linearizes the KGs and uses templates to convert the KGs into passages.
Mindmap \cite{wen2023mindmap} designs a KG prompt to convert graph structure into a mind map that enables LLMs to perform reasoning on it.
Prompt offers a simple way to utilize the potential of LLMs without finetuning. Proficiency in prompt engineering leads to a better understanding of the strengths and weaknesses of LLMs.

\begin{figure}
    \centering
    \includegraphics[width=\columnwidth]{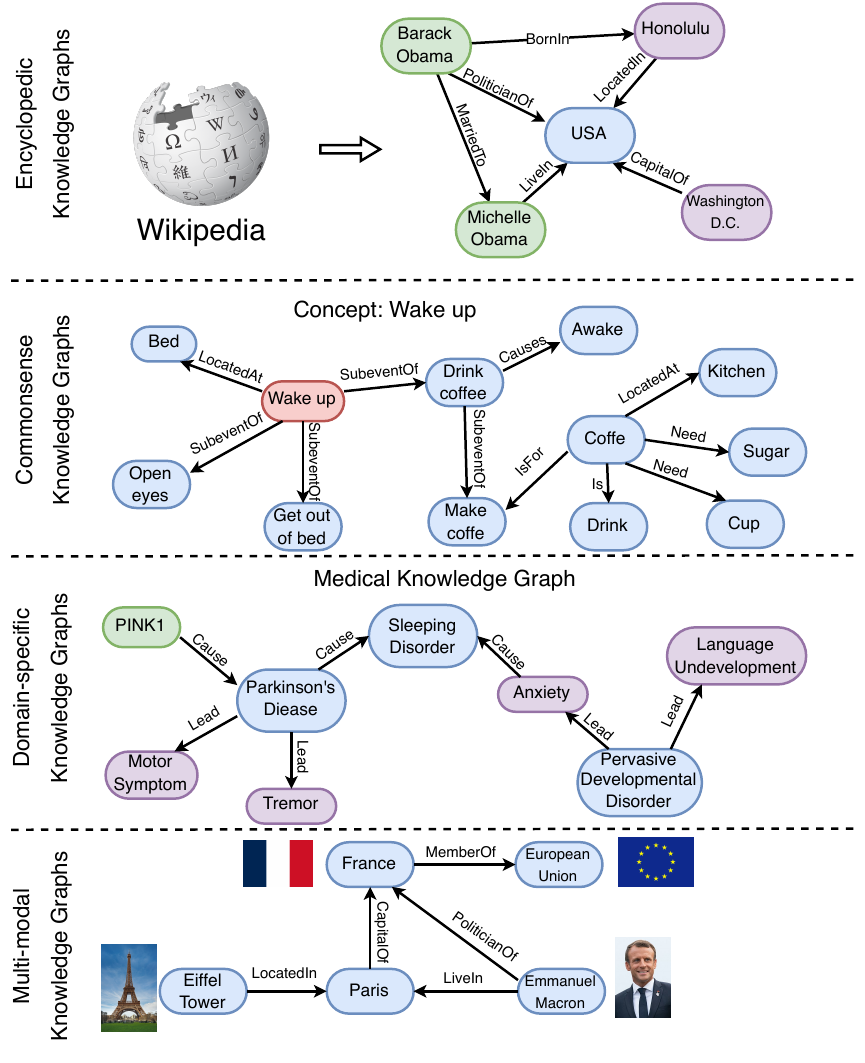}
    \caption{Examples of different categories' knowledge graphs, i.e., \emph{encyclopedic KGs},  \emph{commonsense KGs},  \emph{domain-specific KGs}, and  \emph{multi-modal KGs}.}
    \label{fig:kgs}
\end{figure}

\subsection{Knowledge Graphs (KGs)}


Knowledge graphs (KGs) store structured knowledge as a collection of triples $\mathcal{KG}=\{(h,r,t)\subseteq \mathcal{E}\times \mathcal{R}\times \mathcal{E}\}$, where $\mathcal{E}$ and $\mathcal{R}$ respectively denote the set of entities and relations. Existing knowledge graphs (KGs) can be classified into four groups based on the stored information: \emph{1)~encyclopedic KGs}, \emph{2)~commonsense KGs}, \emph{3)~domain-specific KGs}, and \emph{4)~multi-modal KGs}. We illustrate the examples of KGs of different categories in Fig. \ref{fig:kgs}.

\subsubsection{Encyclopedic Knowledge Graphs.}
Encyclopedic knowledge graphs are the most ubiquitous KGs, which represent the general knowledge in real-world. Encyclopedic knowledge graphs are often constructed by integrating information from diverse and extensive sources, including human experts, encyclopedias, and databases. Wikidata \cite{vrandevcic2014wikidata} is one of the most widely used encyclopedic knowledge graphs, which incorporates varieties of knowledge extracted from articles on Wikipedia. Other typical encyclopedic knowledge graphs, like Freebase \cite{bollacker2008freebase}, Dbpedia \cite{auer2007dbpedia}, and YAGO \cite{suchanek2007yago} are also derived from Wikipedia. In addition, NELL \cite{carlson2010toward} is a continuously improving encyclopedic knowledge graph, which automatically extracts knowledge from the web, and uses that knowledge to improve its performance over time. There are several encyclopedic knowledge graphs available in languages other than English such as CN-DBpedia \cite{xu2017cn} and Vikidia \cite{hai2022vikidia}. The largest knowledge graph, named Knowledge Occean (KO)\footnote{\url{https://ko.zhonghuapu.com/}}, currently contains 4,8784,3636 entities and
17,3115,8349 relations in both English and Chinese.

\subsubsection{Commonsense Knowledge Graphs.}
Commonsense knowledge graphs formulate the knowledge about daily concepts, e.g., objects, and events, as well as their relationships \cite{ilievski2021cskg}. Compared with encyclopedic knowledge graphs, commonsense knowledge graphs often model the tacit knowledge extracted from text such as \textit{(Car, UsedFor, Drive)}. ConceptNet \cite{speer2017conceptnet} contains a wide range of commonsense concepts and relations, which can help computers understand the meanings of words people use. ATOMIC \cite{ji2020language,hwang2021comet} and ASER \cite{zhang2020aser} focus on the causal effects between events, which can be used for commonsense reasoning. Some other commonsense knowledge graphs, such as TransOMCS \cite{zhang2021transomcs} and CausalBanK \cite{ijcai2020-guided} are automatically constructed to provide commonsense knowledge.

\subsubsection{Domain-specific Knowledge Graphs}
Domain-specific knowledge graphs are often constructed to represent knowledge in a specific domain, e.g., medical, biology, and finance \cite{abu2021domain}. Compared with encyclopedic knowledge graphs, domain-specific knowledge graphs are often smaller in size, but more accurate and reliable. For example, UMLS \cite{bodenreider2004unified} is a domain-specific knowledge graph in the medical domain, which contains biomedical concepts and their relationships. In addition, there are some domain-specific knowledge graphs in other domains, such as finance \cite{liu2019anticipating}, geology \cite{zhu2017intelligent}, biology \cite{choi2019inference}, chemistry \cite{farazi2020knowledge} and genealogy \cite{wu-genealogy-23}.

\begin{figure*}
    \centering
    \includegraphics[width=.9\textwidth]{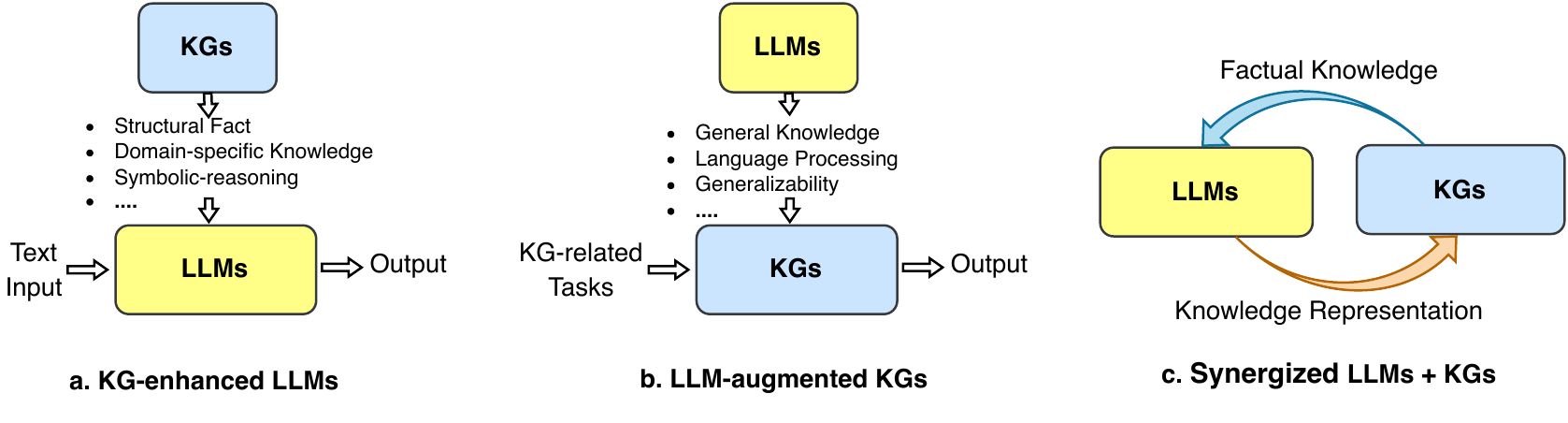}
    \caption{The general roadmap of unifying KGs and LLMs. (a.) KG-enhanced LLMs. (b.) LLM-augmented KGs. (c.) Synergized LLMs + KGs.}
    \label{fig:roadmap}
\end{figure*}

\subsubsection{Multi-modal Knowledge Graphs.}
Unlike conventional knowledge graphs that only contain textual information, multi-modal knowledge graphs represent facts in multiple modalities such as images, sounds, and videos \cite{zhu2022multi}. For example, IMGpedia \cite{ferrada2017imgpedia}, MMKG \cite{liu2019mmkg}, and Richpedia \cite{wang2020richpedia} incorporate both the text and image information into the knowledge graphs. These knowledge graphs can be used for various multi-modal tasks such as image-text matching \cite{shi2019knowledge}, visual question answering \cite{shah2019kvqa}, and recommendation \cite{sun2020multi}.

\begin{table}[]
    \centering
    \caption{Representative applications of using LLMs and KGs.}
    \label{tab:applications}
    \resizebox{\columnwidth}{!}{%
        \begin{tabular}{@{}c|c|c|c|c@{}}
            \toprule
            Name          & Category              & LLMs       & KGs        & URL                             \\ \midrule
            ChatGPT/GPT-4 & Chat Bot              & \checkmark &            & \url{https://shorturl.at/cmsE0} \\
            ERNIE 3.0     & Chat Bot              & \checkmark & \checkmark & \url{https://shorturl.at/sCLV9} \\
            Bard          & Chat Bot              & \checkmark & \checkmark & \url{https://shorturl.at/pDLY6} \\
            Firefly       & Photo Editing         & \checkmark &            & \url{https://shorturl.at/fkzJV} \\
            AutoGPT       & AI Assistant          & \checkmark &            & \url{https://shorturl.at/bkoSY} \\
            Copilot       & Coding Assistant      & \checkmark &            & \url{https://shorturl.at/lKLUV} \\
            New Bing      & Web Search            & \checkmark &            & \url{https://shorturl.at/bimps} \\
            Shop.ai       & Recommendation        & \checkmark &            & \url{https://shorturl.at/alCY7} \\
            Wikidata      & Knowledge Base        &            & \checkmark & \url{https://shorturl.at/lyMY5} \\
            KO            & Knowledge Base        &            & \checkmark & \url{https://shorturl.at/sx238} \\
            OpenBG     & Recommendation        &            & \checkmark & \url{https://shorturl.at/pDMV9} \\
            Doctor.ai     & Health Care Assistant & \checkmark & \checkmark & \url{https://shorturl.at/dhlK0} \\ \bottomrule
        \end{tabular}%
    }
\end{table}

\subsection{Applications}
LLMs as KGs have been widely applied in various real-world applications. We summarize some representative applications of using LLMs and KGs in Table \ref{tab:applications}. ChatGPT/GPT-4 are LLM-based chatbots that can communicate with humans in a natural dialogue format. To improve knowledge awareness of LLMs, ERNIE 3.0 and Bard incorporate KGs into their chatbot applications. Instead of Chatbot. Firefly develops a photo editing application that allows users to edit photos by using natural language descriptions. Copilot, New Bing, and Shop.ai adopt LLMs to empower their applications in the areas of coding assistant, web search, and recommendation, respectively. Wikidata and KO are two representative knowledge graph applications that are used to provide external knowledge.
OpenBG \cite{ICDE2023_OpenBG} is a knowledge graph designed for recommendation. Doctor.ai develops a health care assistant that incorporates LLMs and KGs to provide medical advice.

\section{Roadmap \& Categorization}\label{sec:roadmap}

In this section, we first present a road map of explicit frameworks that unify LLMs and KGs. Then, we present the categorization of research on unifying LLMs and KGs.


\subsection{Roadmap}
The roadmap of unifying KGs and LLMs is illustrated in Fig. \ref{fig:roadmap}. In the roadmap, we identify three frameworks for the unification of LLMs and KGs, including KG-enhanced LLMs, LLM-augmented KGs, and Synergized LLMs + KGs. The KG-enhanced LLMs and LLM-augmented KGs are two parallel frameworks that aim to enhance the capabilities of LLMs and KGs, respectively. Building upon these frameworks, Synergized LLMs + KGs is a unified framework that aims to synergize LLMs and KGs to mutually enhance each other.

\subsubsection{KG-enhanced LLMs}
LLMs are renowned for their ability to learn knowledge from large-scale corpus and achieve state-of-the-art performance in various NLP tasks. However, LLMs are often criticized for their hallucination issues \cite{ji2023survey}, and lacking of interpretability. To address these issues, researchers have proposed to enhance LLMs with knowledge graphs (KGs).

KGs store enormous knowledge in an explicit and structured way, which can be used to enhance the knowledge awareness of LLMs. Some researchers have proposed to incorporate KGs into LLMs during the pre-training stage, which can help LLMs learn knowledge from KGs \cite{zhang-etal-2019-ernie,rosset2020knowledge}. Other researchers have proposed to incorporate KGs into LLMs during the inference stage. By retrieving knowledge from KGs, it can significantly improve the performance of LLMs in accessing domain-specific knowledge \cite{NEURIPS2020_6b493230}. To improve the interpretability of LLMs, researchers also utilize KGs to interpret the facts \cite{petroni2019language} and the reasoning process of LLMs \cite{lin-etal-2019-kagnet}.




\subsubsection{LLM-augmented KGs}
KGs store structure knowledge playing an essential role in many real-word applications \cite{ji2021survey}. Existing methods in KGs fall short of handling incomplete KGs \cite{bordes2013translating} and processing text corpus to construct KGs \cite{zhu2023llms}. With the generalizability of LLMs, many researchers are trying to harness the power of LLMs for addressing KG-related tasks.

The most straightforward way to apply LLMs as text encoders for KG-related tasks. Researchers take advantage of LLMs to process the textual corpus in the KGs and then use the representations of the text to enrich KGs representation \cite{zhang2020pretrain}. Some studies also use LLMs to process the original corpus and extract relations and entities for KG construction \cite{kumar2020building}. Recent studies try to design a KG prompt that
can effectively convert structural KGs into a format that can be comprehended by LLMs. In this way, LLMs can be directly applied to KG-related tasks, e.g., KG completion \cite{GenKGC} and KG reasoning \cite{chen2023incorporating}.


\begin{figure}
    \centering
    \includegraphics[]{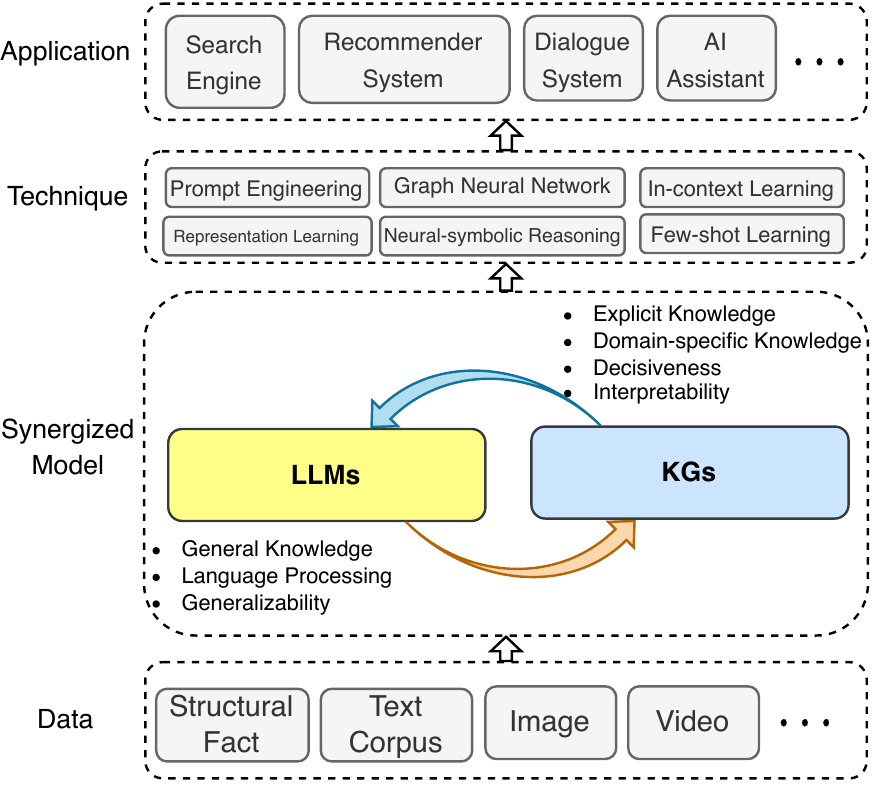}
    \caption{The general framework of the \textit{Synergized LLMs + KGs}, which contains four layers: \emph{1)~Data}, \emph{2)~Synergized Model}, \emph{3)~Technique}, and \emph{4)~Application}.}
    \label{fig:unify_framework}
\end{figure}

\subsubsection{Synergized LLMs + KGs}
The synergy of LLMs and KGs has attracted increasing attention from researchers these years  \cite{wang-etal-2021-kepler,ke-etal-2021-jointgt}. LLMs and KGs are two inherently complementary techniques, which should be unified into a general framework to mutually enhance each other.

To further explore the unification, we propose a unified framework of the synergized LLMs + KGs in Fig. \ref{fig:unify_framework}. The unified framework contains four layers: \emph{1)~Data}, \emph{2)~Synergized Model}, \emph{3)~Technique}, and \emph{4)~Application}. In the \emph{Data} layer, LLMs and KGs are used to process the textual and structural data, respectively. With the development of multi-modal LLMs \cite{zhu2023minigpt} and KGs \cite{challenge-mm-3}, this framework can be extended to process multi-modal data, such as video, audio, and images. In the \emph{Synergized Model} layer, LLMs and KGs could synergize with each other to improve their capabilities. In \emph{Technique} layer, related techniques that have been used in LLMs and KGs can be incorporated into this framework to further enhance the performance. In the \emph{Application} layer, LLMs and KGs can be integrated to address various real-world applications, such as search engines \cite{thoppilan2022lamda}, recommender systems \cite{liu2023chatgpt}, and AI assistants \cite{sun2021ernie}.

\begin{figure*}
    \centering
    \includegraphics[width=0.75\linewidth]{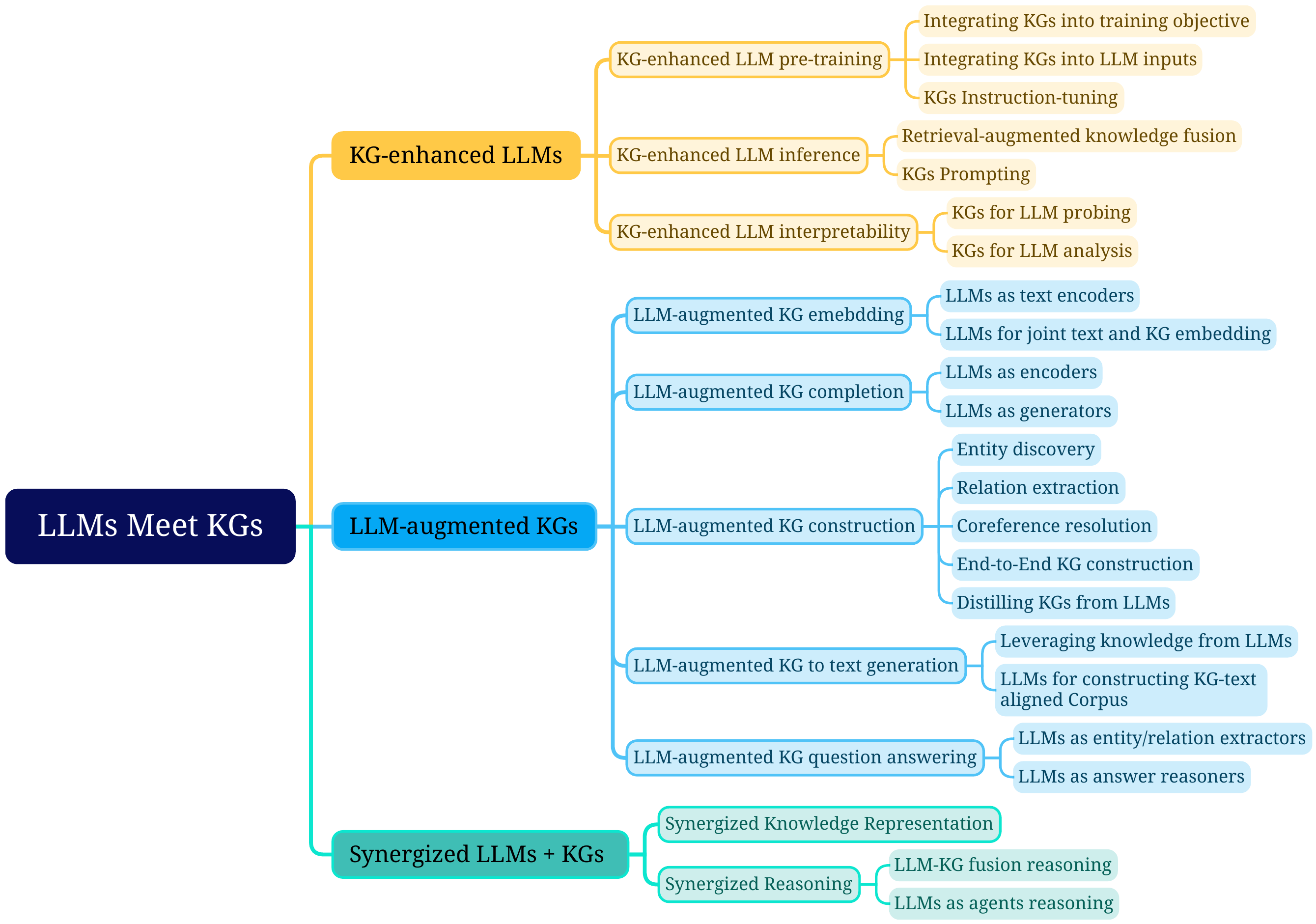}
    \caption{Fine-grained categorization of research on unifying large language models (LLMs) with knowledge graphs (KGs).}
    \label{fig:taxonomy}
\end{figure*}

\subsection{Categorization}
To better understand the research on unifying LLMs and KGs, we further provide a fine-grained categorization for each framework in the roadmap. Specifically, we focus on different ways of integrating KGs and LLMs, i.e., KG-enhanced LLMs, KG-augmented LLMs, and Synergized LLMs + KGs. The fine-grained categorization of the research is illustrated in Fig. \ref{fig:taxonomy}.

\textbf{KG-enhanced LLMs.} Integrating KGs can enhance the performance and interpretability of LLMs in various downstream tasks. We categorize the research on KG-enhanced LLMs into three groups:
\begin{enumerate}
    \item \emph{KG-enhanced LLM pre-training} includes works that apply KGs during the pre-training stage and improve the knowledge expression of LLMs.
    \item \emph{KG-enhanced LLM inference} includes research that utilizes KGs during the inference stage of LLMs, which enables LLMs to access the latest knowledge without retraining.
    \item \emph{KG-enhanced LLM interpretability} includes works that use KGs to understand the knowledge learned by LLMs and interpret the reasoning process of LLMs.
\end{enumerate}

\textbf{LLM-augmented KGs.} LLMs can be applied to augment various KG-related tasks. We categorize the research on LLM-augmented KGs into five groups based on the task types:
\begin{enumerate}
    \item \emph{LLM-augmented KG embedding} includes studies that apply LLMs to enrich representations of KGs by encoding the textual descriptions
          of entities and relations.
    \item \emph{LLM-augmented KG completion} includes papers that utilize LLMs to encode text or generate facts for better KGC performance.
    \item \emph{LLM-augmented KG construction} includes works that apply LLMs to address the entity discovery, coreference resolution, and relation extraction tasks for KG construction.
    \item \emph{LLM-augmented KG-to-text Generation} includes research that utilizes LLMs to generate natural language that describes the facts from KGs.
    \item \emph{LLM-augmented KG question answering} includes studies that apply LLMs to bridge the gap between natural language questions and retrieve answers from KGs.
\end{enumerate}

\textbf{Synergized LLMs + KGs.} The synergy of LLMs and KGs aims to integrate LLMs and KGs into a unified framework to mutually enhance each other. In this categorization, we review the recent attempts of Synergized LLMs + KGs from the perspectives of \emph{knowledge representation} and \emph{reasoning}.

In the following sections (Sec \ref{sec:KG_for_LLM}, \ref{sec:LLM_for_KG}, and \ref{sec:unification_LLM_kg}), we will provide details on these categorizations.

\section{KG-enhanced LLMs}\label{sec:KG_for_LLM}
Large language models (LLMs) achieve promising results in many natural language processing tasks. However, LLMs have been criticized for their lack of practical knowledge and tendency to generate factual errors during inference. To address this issue, researchers have proposed integrating knowledge graphs (KGs) to enhance LLMs. In this section, we first introduce the KG-enhanced LLM pre-training, which aims to inject knowledge into LLMs during the pre-training stage. Then, we introduce the KG-enhanced LLM inference, which enables LLMs to consider the latest knowledge while generating sentences. Finally, we introduce the KG-enhanced LLM interpretability, which aims to improve the interpretability of LLMs by using KGs. Table \ref{tab:summary_of_kgs_for_LLMs} summarizes the typical methods that integrate KGs for LLMs. 

\begin{table}[]
    \centering
    \caption{Summary of KG-enhanced LLM methods.}
    \label{tab:summary_of_kgs_for_LLMs}
    \resizebox{\columnwidth}{!}{%
    \begin{threeparttable}[b]
        \begin{tabular}{@{}c|llll@{}}
            \toprule
            Task                                           & Method                                             & Year                 & KG & Technique                                      \\ \midrule
            \multirow{17}{*}{KG-enhanced LLM pre-training}     & ERNIE \cite{zhang-etal-2019-ernie}                        & 2019      & \textbf{E}           & Integrating KGs into Training Objective        \\
                                                           & GLM  \cite{shen-etal-2020-exploiting}              & 2020  & \textbf{C}               & Integrating KGs into Training Objective        \\
                                                           & Ebert \cite{zhang2020bert}                         & 2020  & \textbf{D}             & Integrating KGs into Training Objective        \\
                                                           & KEPLER  \cite{wang-etal-2021-kepler}                      & 2021 &   \textbf{E}              & Integrating KGs into Training Objective        \\
                                                           & Deterministic LLM \cite{li-etal-2022-pre-training} & 2022 & \textbf{E}                & Integrating KGs into Training Objective        \\
                                                           & KALA \cite{kang2022kala}                           & 2022              & \textbf{D}    & Integrating KGs into Training Objective        \\
                                                           & WKLM \cite{Xiong2020Pretrained}                    & 2020       &\textbf{E}          & Integrating KGs into Training Objective        \\
            \cmidrule(l){2-5}
                                                           & K-BERT \cite{DBLP:conf/aaai/LiuZ0WJD020}           & 2020    & \textbf{E} + \textbf{D}             & Integrating KGs into Language Model Inputs     \\
                                                           & CoLAKE \cite{sun-etal-2020-colake}                 & 2020   & \textbf{E}              & Integrating KGs into Language Model Inputs     \\
                                                           & ERNIE3.0 \cite{sun2021ernie}                       & 2021  & \textbf{E} + \textbf{D}               & Integrating KGs into Language Model Inputs     \\
                                                           & DkLLM \cite{DBLP:conf/aaai/Zhang0HQTH022}          & 2022  & \textbf{E}               & Integrating KGs into Language Model Inputs     \\ \cmidrule(l){2-5}
                                                           & KP-PLM \cite{wang2022knowledge} & 2022 & \textbf{E} & KGs Instruction-tuning \\
         & OntoPrompt \cite{ye2022ontology}    & 2022 & \textbf{E} + \textbf{D} & KGs Instruction-tuning\\
         & ChatKBQA \cite{luo2023chatkbqa}     & 2023 & \textbf{E} & KGs Instruction-tuning\\
         & RoG \cite{luo_rog}                  & 2023 & \textbf{E} & KGs Instruction-tuning \\ \midrule
            \multirow{8}{*}{KG-enhanced LLM inference}      
                                                           & KGLM \cite{logan-etal-2019-baracks}                & 2019   & \textbf{E}              & Retrival-augmented knowledge fusion            \\
                                                           & REALM \cite{10.5555/3524938.3525306}               & 2020   & \textbf{E}              & Retrival-augmented knowledge fusion            \\
                                                           & RAG \cite{NEURIPS2020_6b493230}                    & 2020 & \textbf{E}                & Retrival-augmented knowledge fusion            \\
                                                           & EMAT \cite{wu-etal-2022-efficient}                 & 2022  & \textbf{E}               & Retrival-augmented knowledge fusion            \\\cmidrule(l){2-5}
                                                           & Li et al. \cite{li2023graph}    & 2023 & \textbf{C} & KGs Prompting \\
         & Mindmap \cite{wen2023mindmap}       & 2023  & \textbf{E} + \textbf{D} & KGs Prompting \\
         & ChatRule \cite{luo2023chatrule}     & 2023 & \textbf{E} + \textbf{D} & KGs Prompting \\
         & CoK \cite{wang2023boosting}         & 2023 & \textbf{E} + \textbf{C} + \textbf{D} & KGs Prompting  \\  \midrule
            \multirow{10}{*}{KG-enhanced LLM interpretability} & LAMA \cite{petroni2019language}                    & 2019 & \textbf{E} & KGs for LLM probing                            \\
                                                           & LPAQA \cite{jiang2020can}                          & 2020  & \textbf{E}               & KGs for LLM probing                            \\
                                                           & Autoprompt \cite{shin2020autoprompt}       & 2020 & \textbf{E}                         & KGs for LLM probing                            \\
                                                           & MedLAMA \cite{meng2021rewire}                      & 2022  & \textbf{D}               & KGs for LLM probing                            \\
                                                           & LLM-facteval \cite{luo2023systematic}                   & 2023 &  \textbf{E} + \textbf{D}                & KGs for LLM probing                            \\
            \cmidrule(l){2-5}
                                                           & KagNet \cite{lin-etal-2019-kagnet}                        & 2019  & \textbf{C}              & KGs for LLM analysis                           \\
                                                           & Interpret-lm \cite{swamy2021interpreting}          & 2021  & \textbf{E}              & KGs for LLM analysis                           \\
                                                           & knowledge-neurons \cite{dai2021knowledge}
                                                           & 2021  & \textbf{E}                                              & KGs for LLM analysis                                                  \\
                                                           & Shaobo et al. \cite{li2022pre}                     & 2022  & \textbf{E}                & KGs for LLM analysis                           \\ \bottomrule
        \end{tabular}%
    \begin{tablenotes}
        \item \textbf{E}: Encyclopedic Knowledge Graphs, \textbf{C}: Commonsense Knowledge Graphs, \textbf{D}: Domain-Specific Knowledge Graphs.
      \end{tablenotes}
     \end{threeparttable}
    }
\end{table}

\subsection{KG-enhanced LLM Pre-training}\label{sec:kgs_for_LLMpretraining}
Existing large language models mostly rely on unsupervised training on the large-scale corpus. While these models may exhibit impressive performance on downstream tasks, they often lack practical knowledge relevant to the real world. Previous works that integrate KGs into large language models can be categorized into three parts: \emph{1) Integrating KGs into training objective}, \emph{2) Integrating KGs into LLM inputs}, and \emph{3) KGs Instruction-tuning}.

\subsubsection{Integrating KGs into Training Objective}

The research efforts in this category focus on designing novel knowledge-aware training objectives. An intuitive idea is to expose more knowledge entities in the pre-training objective. GLM~\cite{shen-etal-2020-exploiting} leverages the knowledge graph structure to assign a masking probability. Specifically, entities that can be reached within a certain number of hops are considered to be the most important entities for learning, and they are given a higher masking probability during pre-training. Furthermore,  E-BERT~\cite{zhang2020bert} further controls the balance between the token-level and entity-level training losses.
The training loss values are used as indications of the learning process for token and entity, which dynamically determines their ratio for the next training epochs. SKEP~\cite{tian-etal-2020-skep} also follows a similar fusion to inject sentiment knowledge during LLMs pre-training. SKEP first determines words with positive and negative sentiment by utilizing PMI along with a predefined set of seed sentiment words. Then, it assigns a higher masking probability to those identified sentiment words in the word masking objective.

\begin{figure}[]
    \centering
    \includegraphics[width=.8\columnwidth]{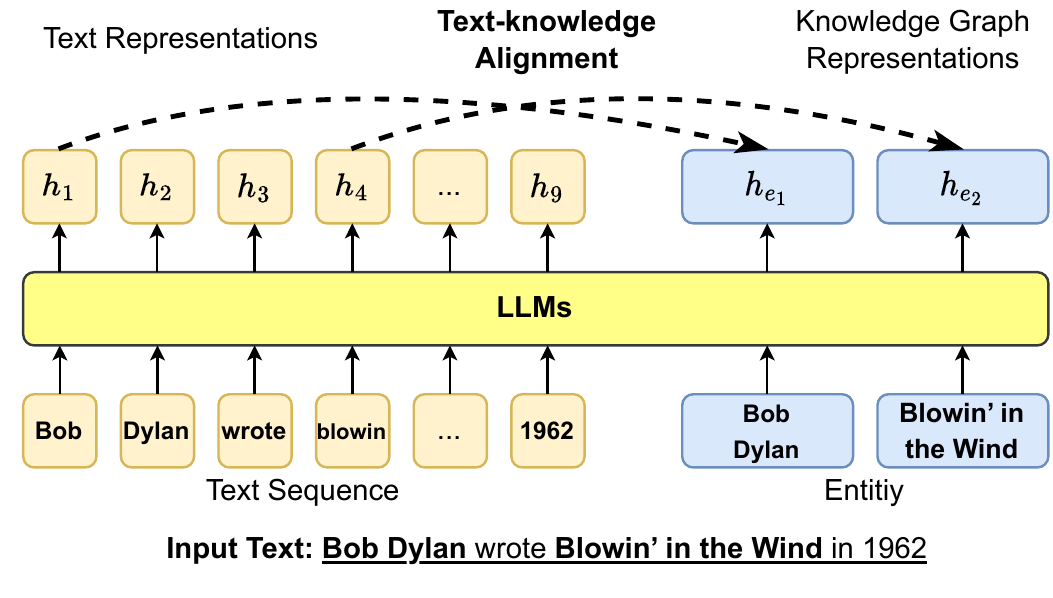}
    \caption{Injecting KG information into LLMs training objective via text-knowledge alignment loss, where $h$ denotes the hidden representation generated by LLMs.}
    \label{fig:KG_LLM_objective}
\end{figure}

The other line of work explicitly leverages the connections with knowledge and input text. As shown in Fig.~\ref{fig:KG_LLM_objective}, ERNIE~\cite{zhang-etal-2019-ernie} proposes a novel word-entity alignment training objective as a pre-training objective. Specifically, ERNIE feeds both sentences and corresponding entities mentioned in the text into LLMs, and then trains the LLMs to predict alignment links between textual tokens and entities in knowledge graphs. Similarly, KALM~\cite{rosset2020knowledge} enhances the input tokens by incorporating entity embeddings and includes an entity prediction pre-training task in addition to the token-only pre-training objective. This approach aims to improve the ability of LLMs to capture knowledge related to entities. Finally, KEPLER~\cite{wang-etal-2021-kepler} directly employs both knowledge graph embedding training objective and Masked token pre-training objective into a shared transformer-based encoder. Deterministic LLM~\cite{li-etal-2022-pre-training} focuses on pre-training language models to capture \emph{deterministic} factual knowledge. It only masks the span that has a deterministic entity as the question and introduces additional clue contrast learning and clue classification objective. WKLM~\cite{Xiong2020Pretrained} first replaces entities in the text with other same-type entities and then feeds them into LLMs. The model is further pre-trained to distinguish whether the entities have been replaced or not.

\subsubsection{Integrating KGs into LLM Inputs}
\begin{figure}[]
    \centering
    \includegraphics[width=0.7\columnwidth]{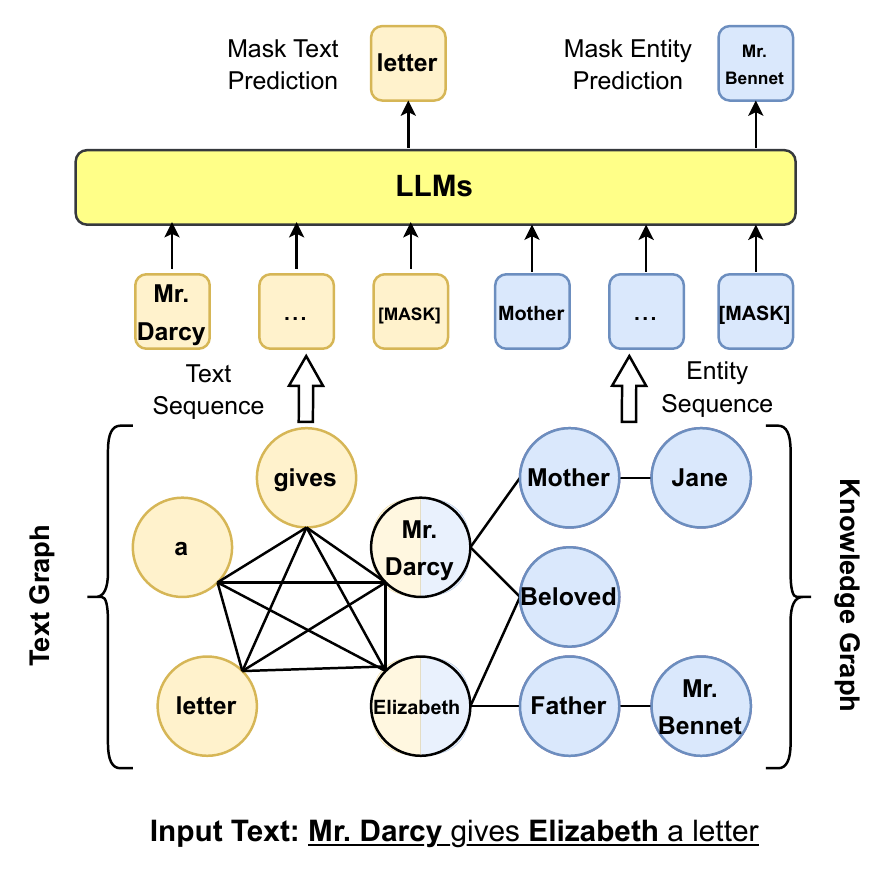}
    \caption{Injecting KG information into LLMs inputs using graph structure.}
    \label{fig:KG_LLM_input}
\end{figure}

As shown in Fig. \ref{fig:KG_LLM_input}, this kind of research focus on introducing relevant knowledge sub-graph into the inputs of LLMs. Given a knowledge graph triple and the corresponding sentences, ERNIE 3.0~\cite{sun2021ernie} represents the triple as a sequence of tokens and directly concatenates them with the sentences. It further randomly masks either the relation token in the triple or tokens in the sentences
to better combine knowledge with textual representations. However, such
direct knowledge triple concatenation method allows the tokens in the sentence to intensively interact with the tokens in the knowledge sub-graph, which could result in \emph{Knowledge Noise}~\cite{DBLP:conf/aaai/LiuZ0WJD020}. To solve this issue, K-BERT~\cite{DBLP:conf/aaai/LiuZ0WJD020} takes the first step to inject the knowledge triple into the sentence via a \emph{visible matrix} where only the knowledge entities have access to the knowledge triple information, while the tokens in the sentences can only see each other in the self-attention module. To further reduce \emph{Knowledge Noise}, Colake~\cite{sun-etal-2020-colake} proposes a unified word-knowledge graph (shown in Fig. \ref{fig:KG_LLM_input}) where the tokens in the input sentences form a fully connected word graph where tokens aligned with knowledge entities are connected with their neighboring entities.

The above methods can indeed inject a large amount of knowledge into LLMs. However, they mostly focus on popular entities and overlook the low-frequent and long-tail ones. DkLLM~\cite{DBLP:conf/aaai/Zhang0HQTH022} aims to improve the LLMs representations towards those entities. DkLLM first proposes a novel measurement to determine long-tail entities and then replaces these selected entities in the text with pseudo token embedding as new input to the large language models. Furthermore, Dict-BERT~\cite{yu-etal-2022-dict} proposes to leverage external dictionaries to solve this issue. Specifically, Dict-BERT improves the representation quality of rare words by appending their definitions from the dictionary at the end of input text and trains the language model to locally align rare word representations in input sentences and dictionary definitions as well as to discriminate whether the input text and definition are correctly mapped.

\subsubsection{KGs Instruction-tuning}\label{sec:kg_instruction_tuning}
Instead of injecting factual knowledge into LLMs, the KGs Instruction-tuning aims to fine-tune LLMs to better comprehend the structure of KGs and effectively follow user instructions to conduct complex tasks. KGs Instruction-tuning utilizes both facts and the structure of KGs to create instruction-tuning datasets. LLMs finetuned on these datasets can extract both factual and structural knowledge from KGs, enhancing the reasoning ability of LLMs.
KP-PLM \cite{wang2022knowledge} first designs several prompt templates to transfer structural graphs into natural language text. Then, two self-supervised tasks are proposed to finetune LLMs to further leverage the knowledge from these prompts. OntoPrompt \cite{ye2022ontology} proposes an ontology-enhanced prompt-tuning that can place knowledge of entities into the context of LLMs, which are further finetuned on several downstream tasks. ChatKBQA \cite{luo2023chatkbqa} finetunes LLMs on KG structure to generate logical queries, which can be executed on KGs to obtain answers. To better reason on graphs, RoG \cite{luo_rog} presents a planning-retrieval-reasoning framework. RoG is finetuned on KG structure to generate relation paths grounded by KGs as faithful plans. These plans are then used to retrieve valid reasoning paths from the KGs for LLMs to conduct faithful reasoning and generate interpretable results.

KGs Instruction-tuning can better leverage the knowledge from KGs for downstream tasks. However, it requires retraining the models, which is time-consuming and requires lots of resources.

\subsection{KG-enhanced LLM Inference}\label{sec:LLM_inference}
\label{kg4LLMinference}
The above methods could effectively fuse knowledge into LLMs. However, real-world knowledge is subject to change and the limitation of these approaches is that they do not permit updates to the incorporated knowledge without retraining the model. As a result, they may not generalize well to the unseen knowledge during inference~\cite{mccoy-etal-2019-right}. Therefore, considerable research has been devoted to keeping the knowledge space and text space separate and injecting the knowledge while inference. These methods mostly focus on the Question Answering (QA) tasks, because QA requires the model to capture both textual semantic meanings and up-to-date real-world knowledge.

\subsubsection{Retrieval-Augmented Knowledge Fusion}

\begin{figure}[]
    \centering
    \includegraphics[width=\columnwidth]{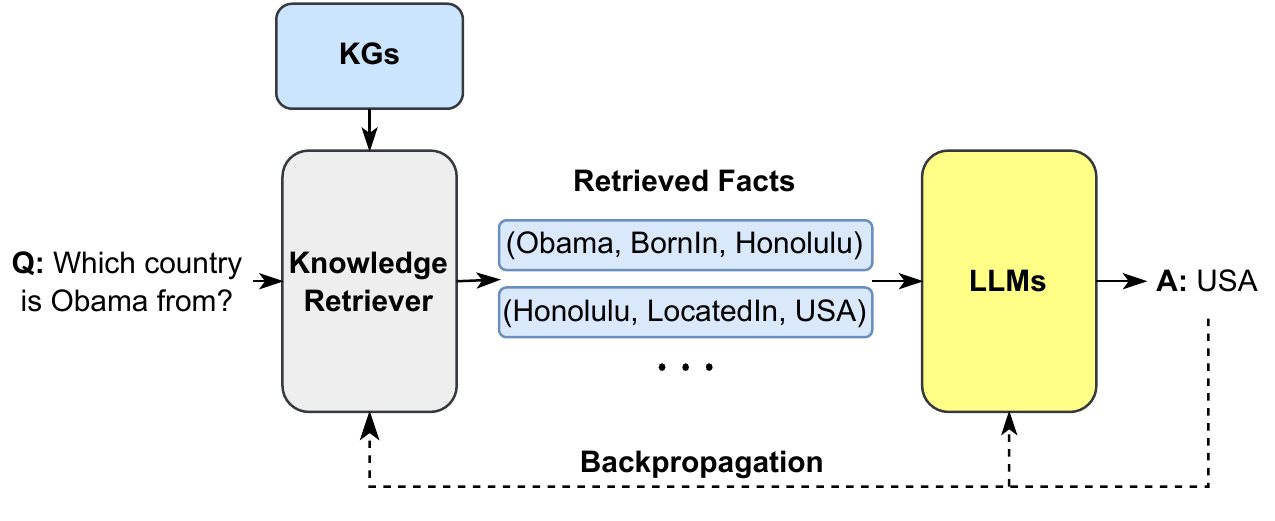}
    \caption{Retrieving external knowledge to enhance the LLM generation.}
    \label{fig:retrivekg}
\end{figure}

Retrieval-Augmented Knowledge Fusion is a popular method to inject knowledge into LLMs during inference. The key idea is to retrieve relevant knowledge from a large corpus and then fuse the retrieved knowledge into LLMs. As shown in Fig.~\ref{fig:retrivekg}, RAG~\cite{NEURIPS2020_6b493230} proposes to combine non-parametric and parametric modules to handle the external knowledge. Given the input text, RAG first searches for relevant KG in the non-parametric module via MIPS to obtain several documents. RAG then treats these documents as hidden variables $z$ and feeds them into the output generator, empowered by Seq2Seq LLMs, as additional context information. The research indicates that using different retrieved documents as conditions at different generation steps performs better than only using a single document to guide the whole generation process.
The experimental results show that RAG outperforms other parametric-only and non-parametric-only baseline models in open-domain QA. RAG can also generate more specific, diverse, and factual text than other parameter-only baselines. Story-fragments \cite{wilmot-keller-2021-memory} further improves architecture by adding an additional module to determine salient knowledge entities and fuse them into the generator to improve the quality of generated long stories. EMAT~\cite{wu-etal-2022-efficient} further improves the efficiency of such a system by encoding external knowledge into a key-value memory and exploiting the
fast maximum inner product search for memory querying. REALM~\cite{10.5555/3524938.3525306} proposes a novel knowledge retriever to help the model to retrieve and attend over documents from a large corpus during the pre-training stage and successfully improves the performance of open-domain question answering. KGLM~\cite{logan-etal-2019-baracks} selects the facts from a knowledge graph using the current context to generate factual sentences. With the help of an external knowledge graph, KGLM could describe facts using out-of-domain words or phrases.

\subsubsection{KGs Prompting}\label{sec:kg_prompting}
To better feed the KG structure into the LLM during inference, KGs prompting aims to design a crafted prompt that converts structured KGs into text sequences, which can be fed as context into LLMs. In this way, LLMs can better take advantage of the structure of KGs to perform reasoning. Li et al. \cite{li2023graph} adopt the pre-defined template to convert each triple into a short sentence, which can be understood by LLMs for reasoning. Mindmap \cite{wen2023mindmap} designs a KG prompt to convert graph structure into a mind map that enables LLMs to perform reasoning by consolidating the facts in KGs and the implicit knowledge from LLMs. 
ChatRule \cite{luo2023chatrule} samples several relation paths from KGs, which are verbalized and fed into LLMs. Then, LLMs are prompted to generate meaningful logical rules that can be used for reasoning. CoK \cite{wang2023boosting} proposes a chain-of-knowledge prompting that uses a sequence of triples to elicit the reasoning ability of LLMs to reach the final answer.

KGs prompting presents a simple way to synergize LLMs and KGs. By using the prompt, we can easily harness the power of LLMs to perform reasoning based on KGs without retraining the models. However, the prompt is usually designed manually, which requires lots of human effort.

\subsection{Comparison between KG-enhanced LLM Pre-training and Inference}
KG-enhanced LLM Pre-training methods commonly enrich large-amount of unlabeled corpus with semantically relevant real-world knowledge. These methods allow the knowledge representations to be aligned with appropriate linguistic context and explicitly train LLMs to leverage those knowledge from scratch. When applying the resulting LLMs to downstream knowledge-intensive tasks, they should achieve optimal performance. In contrast, KG-enhanced LLM inference methods only present the knowledge to LLMs in the inference stage and the underlying LLMs may not be trained to fully leverage these knowledge when conducting downstream tasks, potentially resulting in sub-optimal model performance.

However, real-world knowledge is dynamic and requires frequent updates. Despite being effective, the KG-enhanced LLM Pre-training methods never permit knowledge updates or editing without model re-training. As a result, the KG-enhanced LLM Pre-training methods could generalize poorly to recent or unseen knowledge. KG-enhanced LLM inference methods can easily maintain knowledge updates by changing the inference inputs. These methods help improve LLMs performance on new knowledge and domains.

In summary, when to use these methods depends on the application scenarios. If one wishes to apply LLMs to handle time-insensitive knowledge in particular domains (e.g., commonsense and reasoning knowledge), KG-enhanced LLM Pre-training methods should be considered. Otherwise, KG-enhanced LLM inference methods can be used to handle open-domain knowledge with frequent updates.

\begin{figure}[]
    \centering
    \includegraphics[width=.7\columnwidth]{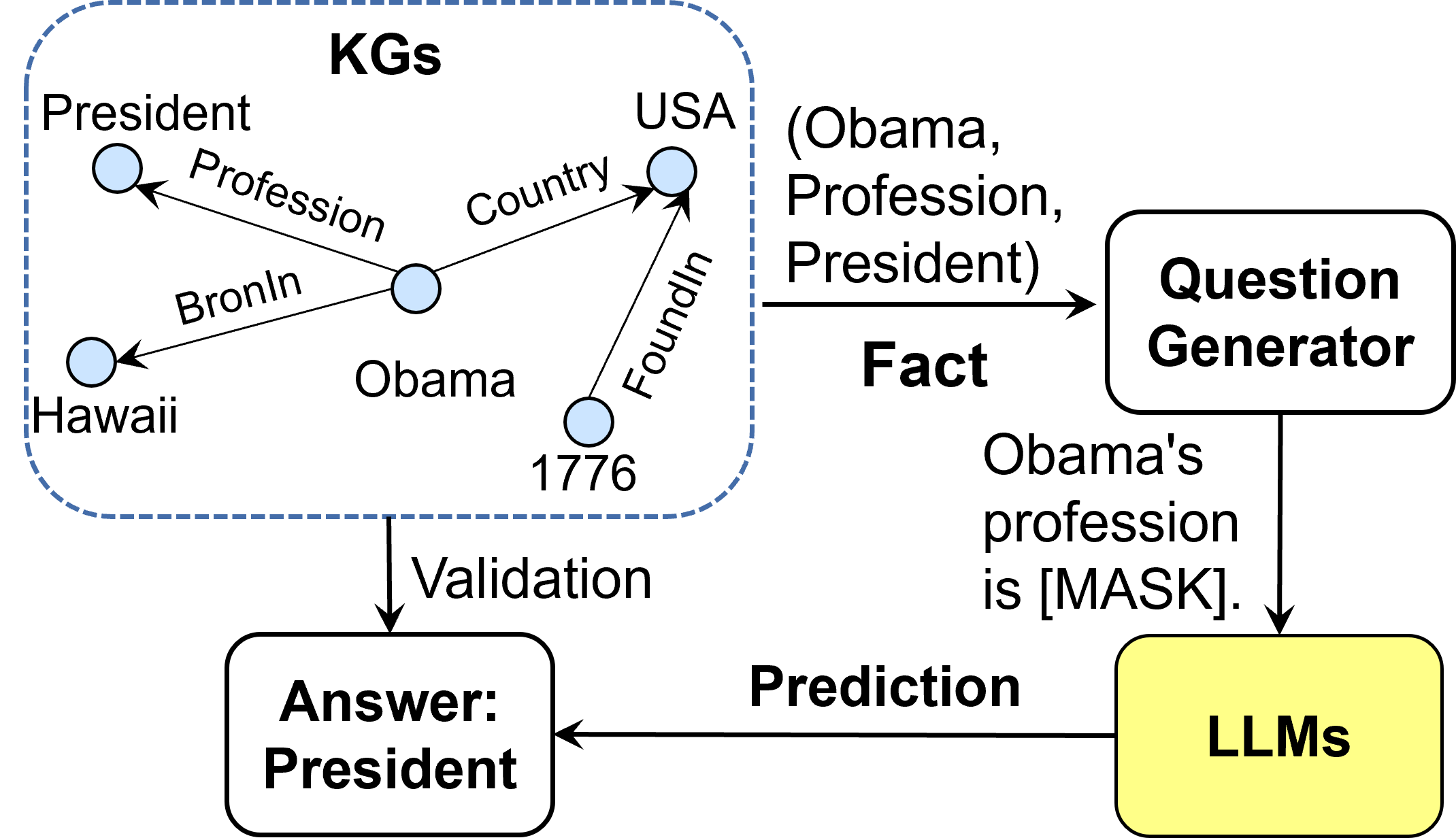}
    \caption{The general framework of using knowledge graph for language model probing.}
    \label{fig: KG for LMP}
\end{figure}

\subsection{KG-enhanced LLM Interpretability}
\label{LLMinterpretability}
Although LLMs have achieved remarkable success in many NLP tasks, they are still criticized for their lack of interpretability. The large language model (LLM) interpretability refers to the understanding and explanation of the inner workings and decision-making processes of a large language model \cite{danilevsky2020survey}. This can improve the trustworthiness of LLMs and facilitate their applications in high-stakes scenarios such as medical diagnosis and legal judgment.
Knowledge graphs (KGs) represent the knowledge structurally and can provide good interpretability for the reasoning results. Therefore, researchers try to utilize KGs to improve the interpretability of LLMs, which can be roughly grouped into two categories: \emph{1) KGs for language model probing}, and \emph{2) KGs for language model analysis}.

\subsubsection{KGs for LLM Probing}
The large language model (LLM) probing aims to understand the knowledge stored in LLMs. LLMs, trained on large-scale corpus, are often known as containing enormous knowledge. However, LLMs store the knowledge in a hidden way, making it hard to figure out the stored knowledge. Moreover, LLMs suffer from the hallucination problem \cite{ji2023survey}, which results in generating statements that contradict facts. This issue significantly affects the reliability of LLMs. Therefore, it is necessary to probe and verify the knowledge stored in LLMs.

LAMA \cite{petroni2019language} is the first work to probe the knowledge in LLMs by using KGs. As shown in Fig. \ref{fig: KG for LMP}, LAMA first converts the facts in KGs into cloze statements by a pre-defined prompt template and then uses LLMs to predict the missing entity. The prediction results are used to evaluate the knowledge stored in LLMs. For example, we try to probe whether LLMs know the fact \textit{(Obama, profession, president)}. We first convert the fact triple into a cloze question ``Obama's profession is $\_$." with the object masked. Then, we test if the LLMs can predict the object ``president" correctly.

However, LAMA ignores the fact that the prompts are inappropriate. For example, the prompt \textit{``Obama worked as a \_"} may be more favorable to the prediction of the blank by the language models than \textit{``Obama is a \_ by profession"}. Thus, LPAQA \cite{jiang2020can} proposes a mining and paraphrasing-based method to automatically generate high-quality and diverse prompts for a more accurate assessment of the knowledge contained in the language model. Moreover, Adolphs et al. \cite{adolphs2021query} attempt to use examples to make the language model understand the query, and experiments obtain substantial improvements for BERT-large on the T-REx data. Unlike using manually defined prompt templates, Autoprompt \cite{shin2020autoprompt} proposes an automated method, which is based on the gradient-guided search to create prompts. LLM-facteval \cite{luo2023systematic} designs a systematic framework that automatically generates probing questions from KGs. The generated questions are then used to evaluate the factual knowledge stored in LLMs.

Instead of probing the general knowledge by using the encyclopedic and commonsense knowledge graphs, BioLAMA \cite{sung2021can} and MedLAMA \cite{meng2021rewire} probe the medical knowledge in LLMs by using medical knowledge graphs.
Alex et al. \cite{mallen2022not} investigate the capacity of LLMs to retain less popular factual knowledge. They select unpopular facts from Wikidata knowledge graphs which have low-frequency clicked entities. These facts are then used for the evaluation, where the results indicate that LLMs encounter difficulties with such knowledge, and that scaling fails to appreciably improve memorization of factual knowledge in the tail.

\begin{figure}[]
    \centering
    \includegraphics[width=0.7\columnwidth]{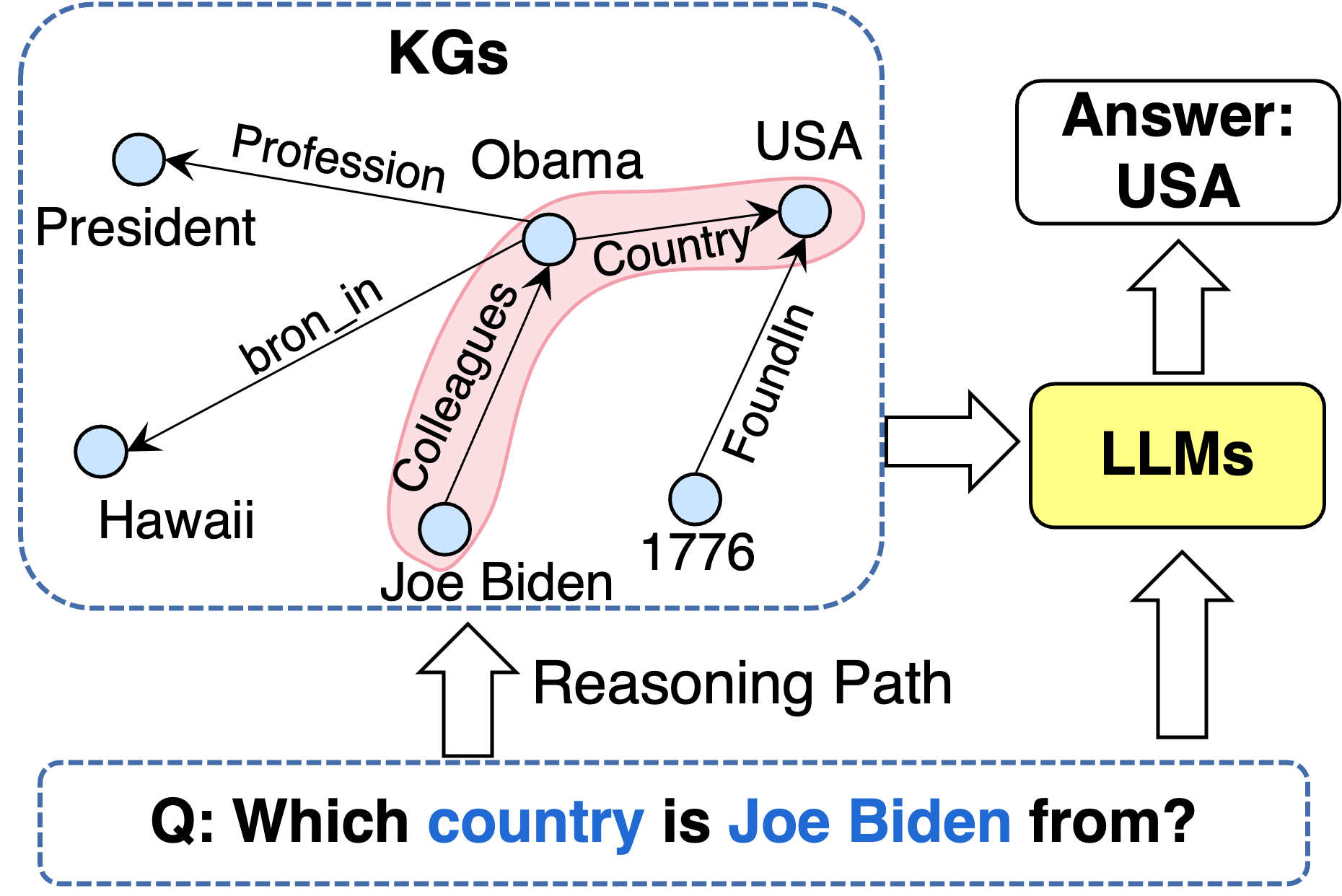}
    \caption{The general framework of using knowledge graph for language model analysis.}
    \label{fig: KG for LM}
\end{figure}


\subsubsection{KGs for LLM Analysis}

Knowledge graphs (KGs) for pre-train language models (LLMs) analysis aims to answer the following questions such as ``how do LLMs generate the results?'', and ``how do the function and structure work in LLMs?''.
To analyze the inference process of LLMs, as shown in Fig. \ref{fig: KG for LM}, KagNet \cite{lin-etal-2019-kagnet} and QA-GNN \cite{yasunaga-etal-2021-qa} make the results generated by LLMs at each reasoning step grounded by knowledge graphs. In this way, the reasoning process of LLMs can be explained by extracting the graph structure from KGs. Shaobo et al. \cite{li2022pre} investigate how LLMs generate the results correctly. They adopt the causal-inspired analysis from facts extracted from KGs. This analysis quantitatively measures the word patterns that LLMs depend on to generate the results. The results show that LLMs generate the missing factual more by the positionally closed words rather than the knowledge-dependent words. Thus, they claim that LLMs are inadequate to memorize factual knowledge because of the inaccurate dependence.
To interpret the training of LLMs, Swamy et al. \cite{swamy2021interpreting} adopt the language model during pre-training to generate knowledge graphs. The knowledge acquired by LLMs during training can be unveiled by the facts in KGs explicitly. To explore how implicit knowledge is stored in parameters of LLMs, Dai~et al. \cite{dai2021knowledge} propose the concept of \emph{knowledge neurons}. Specifically, activation of the identified knowledge neurons is highly correlated with knowledge expression. Thus, they explore the knowledge and facts represented by each neuron by suppressing and amplifying knowledge neurons.

\section{LLM-augmented KGs}\label{sec:LLM_for_KG}
Knowledge graphs are famous for representing knowledge in a structural manner. They have been applied in many downstream tasks such as question answering, recommendation, and web search. However, the conventional KGs are often incomplete and existing methods often lack considering textual information. To address these issues, recent research has explored integrating LLMs to augment KGs to consider the textual information and improve the performance in downstream tasks. In this section, we will introduce the recent research on LLM-augmented KGs. We will introduce the methods that integrate LLMs for KG embedding, KG completion, KG construction, KG-to-text generation, and KG question answering, respectively. Representative works are summarized in Table \ref{tab:summary_of_LLMs_for_kgs}.

\begin{table}[]
    \centering
    \caption{Summary of representative LLM-augmented KG methods.}
    \label{tab:summary_of_LLMs_for_kgs}
    \resizebox{\columnwidth}{!}{%
    \begin{threeparttable}[b]
        \begin{tabular}{@{}l|llll@{}}
            \toprule
            Task                                           & Method                                               & Year & LLM &Technique                            \\ \midrule
            \multirow{8}{*}{LLM-augmented KG embedding}          & Pretrain-KGE \cite{zhang2020pretrain}                & 2020 & \textbf{E} & LLMs as Text Encoders                \\
                                                            & KEPLER \cite{wang-etal-2021-kepler}                         & 2020 & \textbf{E} & LLMs as Text Encoders                \\
                                                            & Nayyeri et al. \cite{nayyeri2022integrating}         & 2022 & \textbf{E} & LLMs as Text Encoders                \\
                                                            & Huang et al. \cite{huang2022endowing}                & 2022 & \textbf{E} & LLMs as Text Encoders                \\
                                                            & CoDEx \cite{alam2022language}                        & 2022 & \textbf{E} & LLMs as Text Encoders                \\ \cmidrule(l){2-5}
                                                            & LMKE \cite{wang2022language}                         & 2022 & \textbf{E} & LLMs for Joint Text and KG Embedding \\
                                                            & kNN-KGE \cite{zhang2022reasoning}                    & 2022 & \textbf{E} & LLMs for Joint Text and KG Embedding \\
                                                            & LambdaKG \cite{lambdakg}                             & 2023 & \textbf{E} + \textbf{D} + \textbf{ED} & LLMs for Joint Text and KG Embedding \\ \midrule
            \multirow{12}{*}{LLM-augmented KG completion}        & KG-BERT~\cite{yao2019kg}                               & 2019 & \textbf{E} & Joint Encoding                       \\
                                                            & MTL-KGC~\cite{MTL-KGC}                               & 2020 & \textbf{E} & Joint Encoding                       \\
                                                            & PKGC~\cite{PKGC}                                     & 2022 & \textbf{E} & Joint Encoding                       \\
                                                            & LASS~\cite{LASS}                                     & 2022 & \textbf{E} & Joint Encoding                       \\ \cmidrule(l){2-5}
                                                            & MEM-KGC~\cite{MEM-KGC}                               & 2021 & \textbf{E} & MLM Encoding                         \\
                                                            & OpenWorld KGC \cite{open-world-KGC}                  & 2023 & \textbf{E} & MLM Encoding                         \\ \cmidrule(l){2-5}
                                                            & StAR~\cite{StAR}                                     & 2021 & \textbf{E} & Separated Encoding                   \\
                                                            & SimKGC~\cite{SimKGC}                                 & 2022 & \textbf{E} & Separated Encoding                   \\
                                                            & LP-BERT \cite{LP-BERT}                               & 2022 & \textbf{E} & Separated Encoding                   \\ \cmidrule(l){2-5}
                                                            & GenKGC~\cite{GenKGC}                                 & 2022 & \textbf{ED} & LLM as decoders                      \\
                                                            & KGT5~\cite{KGT5}                                     & 2022 & \textbf{ED} & LLM as decoders                      \\
                                                            & KG-S2S~\cite{KG-S2S}                                 & 2022 & \textbf{ED} & LLM as decoders                      \\
                                                            & AutoKG~\cite{zhu2023llms} & 2023 & \textbf{D} & LLM as decoders \\
                                                            \midrule
            \multirow{18}{*}{LLM-augmented KG construction}      & ELMO \cite{Elmo} & 2018 & \textbf{E} & Named Entity Recognition \\
            & GenerativeNER \cite{generativeNER} & 2021 & \textbf{ED} & Named Entity Recognition \\
                                                            & LDET \cite{LDET}                                     & 2019 & \textbf{E} & Entity Typing                        \\
                                                            & BOX4Types \cite{BOX4Types}                           & 2021 & \textbf{E} & Entity Typing                        \\
                                                            & ELQ~\cite{ELQ}                                       & 2020 & \textbf{E} & Entity Linking                       \\
                                                            & ReFinED \cite{ReFinED}                                 & 2022 & \textbf{E} & Entity Linking                       \\ \cmidrule(l){2-5}
                                                            & BertCR \cite{CR1}                                    & 2019 & \textbf{E} & CR (Within-document)                 \\
                                                            & Spanbert \cite{SpanBERT}                             & 2020 & \textbf{E} & CR (Within-document)                 \\
                                                            & CDLM \cite{CDLM}                                     & 2021 & \textbf{E} & CR (Cross-document)                  \\
                                                            & CrossCR \cite{crossCR}                               & 2021 & \textbf{E} & CR (Cross-document)                  \\
                                                            & CR-RL \cite{CR-RL}                                   & 2021 & \textbf{E} & CR (Cross-document)                  \\ \cmidrule(l){2-5}
                                                            & SentRE \cite{sent-re1}                               & 2019 & \textbf{E} & RE (Sentence-level)                  \\
                                                            & Curriculum-RE \cite{Curriculum-RE}                   & 2021 & \textbf{E} & RE (Sentence-level)                  \\
                                                            & DREEAM \cite{DREEAM}                                 & 2023 & \textbf{E} & RE (Document-level)                  \\ \cmidrule(l){2-5}
                                                            & Kumar et al. \cite{kumar2020building}                & 2020 & \textbf{E} & End-to-End Construction              \\
                                                            & Guo et al. \cite{guo2021constructing}                & 2021 & \textbf{E} & End-to-End Construction              \\
                                                            & Grapher \cite{melnyk2021grapher}                     & 2021 & \textbf{ED} & End-to-End Construction              \\
                                                            & PiVE \cite{han2023pive}                              & 2023 & \textbf{D} +  \textbf{ED} & End-to-End Construction              \\
            \cmidrule(l){2-5}
                                                            & COMET \cite{bosselut2019comet}                       & 2019 & \textbf{D}  & Distilling KGs from LLMs             \\
                                                            & BertNet \cite{hao2022bertnet}                        & 2022 & \textbf{E} & Distilling KGs from LLMs             \\
                                                            & West et al. \cite{west2022symbolic}                  & 2022 & \textbf{D} & Distilling KGs from LLMs             \\ \midrule
            \multirow{6}{*}{LLM-augmented KG-to-text Generation} & Ribeiro et al \cite{ribeiro-etal-2021-investigating} & 2021 & \textbf{ED} & Leveraging Knowledge from LLMs       \\                                      
            & JointGT \cite{ke-etal-2021-jointgt}                  & 2021 & \textbf{ED} & Leveraging Knowledge from LLMs       \\
                                                            & FSKG2Text \cite{li-etal-2021-shot-knowledge}         & 2021 & \textbf{D} + \textbf{ED}& Leveraging Knowledge from LLMs       \\
                                                            & GAP \cite{colas-etal-2022-gap}                       & 2022 & \textbf{ED} & Leveraging Knowledge from LLMs       \\ \cmidrule(l){2-5}
                                                            & GenWiki \cite{jin-etal-2020-genwiki}                 & 2020 & - & Constructing KG-text aligned Corpus  \\
                                                            & KGPT \cite{chen-etal-2020-kgpt}                      & 2020 & \textbf{ED} & Constructing KG-text aligned Corpus  \\ \midrule
            \multirow{10}{*}{LLM-augmented KGQA}                                                                             & Lukovnikov et al. \cite{lukovnikov2019pretrained}    & 2019 & \textbf{E}  & Entity/Relation Extractor            \\
            & Luo et al. \cite{luo2020bert}                        & 2020 & \textbf{E} & Entity/Relation Extractor            \\
                                                            & QA-GNN \cite{yasunaga-etal-2021-qa}                  & 2021 & \textbf{E} & Entity/Relation Extractor            \\
                                                            & Nan et al. \cite{hu2023empirical}                    & 2023 & \textbf{E} + \textbf{D} + \textbf{ED} & Entity/Relation Extractor            \\ \cmidrule(l){2-5}
                                                            & DEKCOR \cite{xu2021fusing}                           & 2021 & \textbf{E} & Answer Reasoner                      \\
                                                            & DRLK \cite{zhang2022drlk}                            & 2022 & \textbf{E} & Answer Reasoner                      \\
                                                            & OreoLM \cite{hu2022empowering}                       & 2022 & \textbf{E} & Answer Reasoner                      \\
                                                            & GreaseLM \cite{zhang2022greaselm}                    & 2022 & \textbf{E} & Answer Reasoner                      \\
                                                            & ReLMKG \cite{cao2022relmkg}                          & 2022 & \textbf{E} & Answer Reasoner                      \\
                                                            & UniKGQA \cite{jiang2023unikgqa}                      & 2023 & \textbf{E} & Answer Reasoner                      \\ \bottomrule
        \end{tabular}%
        \begin{tablenotes}
            \item \textbf{E}: Encoder-only LLMs, \textbf{D}: Decoder-only LLMs, \textbf{ED}: Encoder-decoder LLMs.
          \end{tablenotes}
         \end{threeparttable}
    }
\end{table}

\subsection{LLM-augmented KG Embedding}
\begin{figure}[]
    \centering
    \includegraphics[width=0.7\columnwidth]{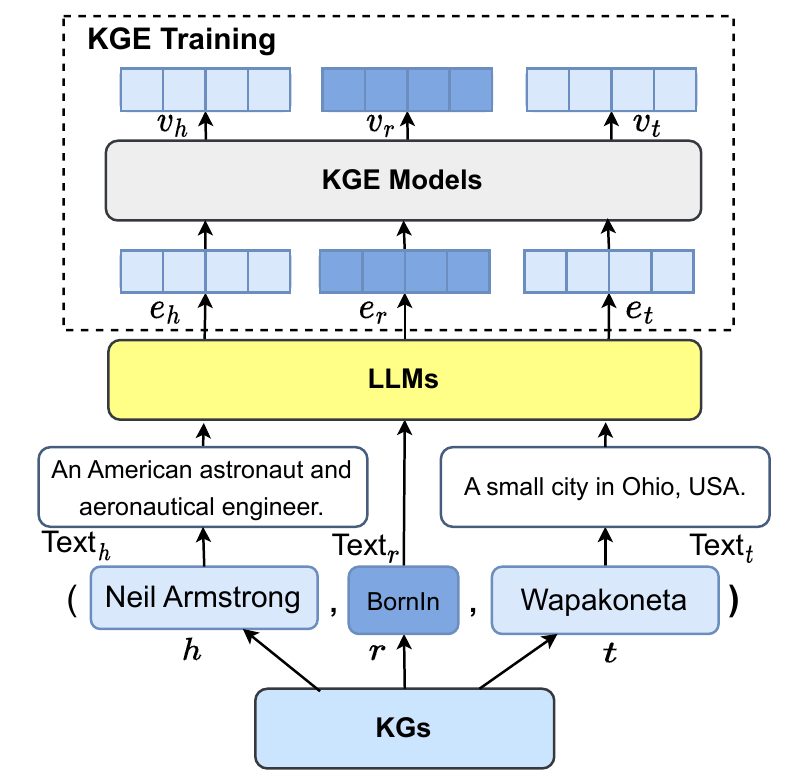}
    \caption{LLMs as text encoder for knowledge graph embedding (KGE).}
    \label{fig:LLM_for_KGE}
\end{figure}

Knowledge graph embedding (KGE) aims to map each entity and relation into a low-dimensional vector (embedding) space. These embeddings contain both semantic and structural information of KGs, which can be utilized for various tasks such as question answering \cite{huang2019knowledge}, reasoning \cite{lin-etal-2019-kagnet}, and recommendation \cite{wang2018dkn}. Conventional knowledge graph embedding methods mainly rely on the structural information of KGs to optimize a scoring function defined on embeddings (e.g., TransE \cite{bordes2013translating}, and DisMult \cite{yang2015embedding}). However, these approaches often fall short in representing unseen entities and long-tailed relations due to their limited structural connectivity \cite{xiong2018one,wang2019logic}. To address this issue, as shown in Fig. \ref{fig:LLM_for_KGE}, recent research adopts LLMs to enrich representations of KGs by encoding the textual descriptions of entities and relations \cite{zhang2020pretrain,wang-etal-2021-kepler}.

\subsubsection{LLMs as Text Encoders}
Pretrain-KGE \cite{zhang2020pretrain} is a representative method that follows the framework shown in Fig. \ref{fig:LLM_for_KGE}. Given a triple $(h,r,t)$ from KGs, it firsts uses a LLM encoder to encode the textual descriptions of entities $h$, $t$, and relations $r$ into representations as
\begin{gather}
    e_h = \text{LLM}(\text{Text}_h),
    e_t = \text{LLM}(\text{Text}_t),
    e_r = \text{LLM}(\text{Text}_r),
\end{gather}
where $e_h,e_r,$ and $e_t$ denotes the initial embeddings of entities $h$, $t$, and relations $r$, respectively. Pretrain-KGE uses the BERT as the LLM encoder in experiments. Then, the initial embeddings are fed into a KGE model to generate the final embeddings $v_h,v_r$, and $v_t$. During the KGE training phase, they optimize the KGE model by following the standard KGE loss function as
\begin{equation}
    \mathcal{L} = [\gamma + f(v_h,v_r,v_t)-f(v'_h,v'_r,v'_t)],
\end{equation}
where $f$ is the KGE scoring function, $\gamma$ is a margin hyperparameter, and $v'_h,v'_r$, and $v'_t$ are the negative samples. In this way, the KGE model could learn adequate structure information, while reserving partial knowledge from LLM enabling better knowledge graph embedding. KEPLER \cite{wang-etal-2021-kepler} offers a unified model for knowledge embedding and pre-trained language representation. This model not only generates effective text-enhanced knowledge embedding using powerful LLMs but also seamlessly integrates factual knowledge into LLMs. Nayyeri et al. \cite{nayyeri2022integrating} use LLMs to generate the world-level, sentence-level, and document-level representations. They are integrated with graph structure embeddings into a unified vector by Dihedron and Quaternion representations of 4D hypercomplex numbers. Huang et al. \cite{huang2022endowing} combine LLMs with other vision and graph encoders to learn multi-modal knowledge graph embedding that enhances the performance of downstream tasks. CoDEx \cite{alam2022language} presents a novel loss function empowered by LLMs that guides the KGE models in measuring the likelihood of triples by considering the textual information. The proposed loss function is agnostic to model structure that can be incorporated with any KGE model.

\subsubsection{LLMs for Joint Text and KG Embedding}
\begin{figure}[]
    \centering
    \includegraphics[width=0.8\columnwidth]{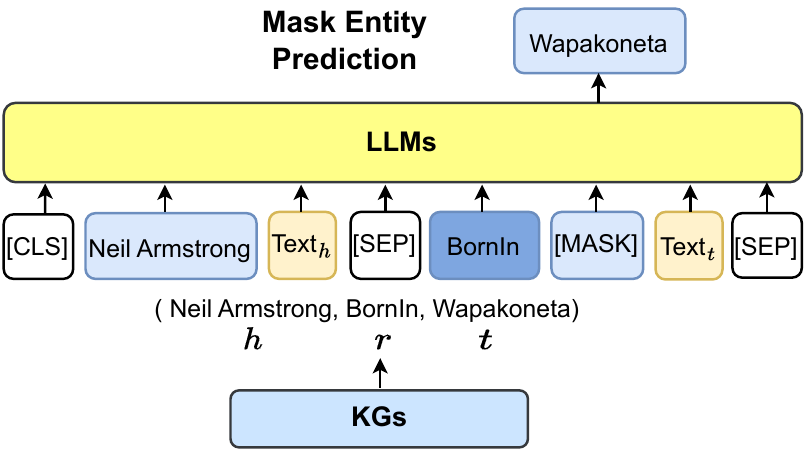}
    \caption{LLMs for joint text and knowledge graph embedding.}
    \label{fig:LLM_for_KGE_2}
\end{figure}

Instead of using KGE model to consider graph structure, another line of methods directly employs LLMs to incorporate both the graph structure and textual information into the embedding space simultaneously. As shown in Fig. \ref{fig:LLM_for_KGE_2}, $k$NN-KGE \cite{zhang2022reasoning} treats the entities and relations as special tokens in the LLM. During training, it transfers each triple $(h,r,t)$ and corresponding text descriptions into a sentence $x$ as
\begin{equation}
    x = \texttt{[CLS]} \ h \ \ \text{Text}_h \texttt{[SEP]} \ r  \ \texttt{[SEP]} \ \texttt{[MASK]} \ \ \text{Text}_t \texttt{[SEP]},
\end{equation}
where the tailed entities are replaced by \texttt{[MASK]}. The sentence is fed into a LLM, which then finetunes the model to predict the masked entity, formulated as
\begin{equation}
    P_{LLM}(t|h,r) = P(\texttt{[MASK]=t}|x, \Theta), \label{eq:LLM_kge_masking}
\end{equation}
where $\Theta$ denotes the parameters of the LLM. The LLM is optimized to maximize the probability of the correct entity $t$. After training, the corresponding token representations in LLMs are used as embeddings for entities and relations. Similarly, LMKE \cite{wang2022language} proposes a contrastive learning method to improve the learning of embeddings generated by LLMs for KGE. Meanwhile, to better capture graph structure, LambdaKG \cite{lambdakg} samples 1-hop neighbor entities and concatenates their tokens with the triple as a sentence feeding into LLMs.

\subsection{LLM-augmented KG Completion}

Knowledge Graph Completion (KGC) refers to the task of inferring missing facts in a given knowledge graph. Similar to KGE, conventional KGC methods mainly focused on the structure of the KG, without considering the extensive textual information. However, the recent integration of LLMs enables KGC methods to encode text or generate facts for better KGC performance. These methods fall into two distinct categories based on their utilization styles: \emph{1) LLM as Encoders (PaE)}, and \emph{2) LLM as Generators (PaG)}.
\subsubsection{LLM as Encoders (PaE).} 
\begin{figure}[]
    \centering
    \includegraphics[width=.7\columnwidth]{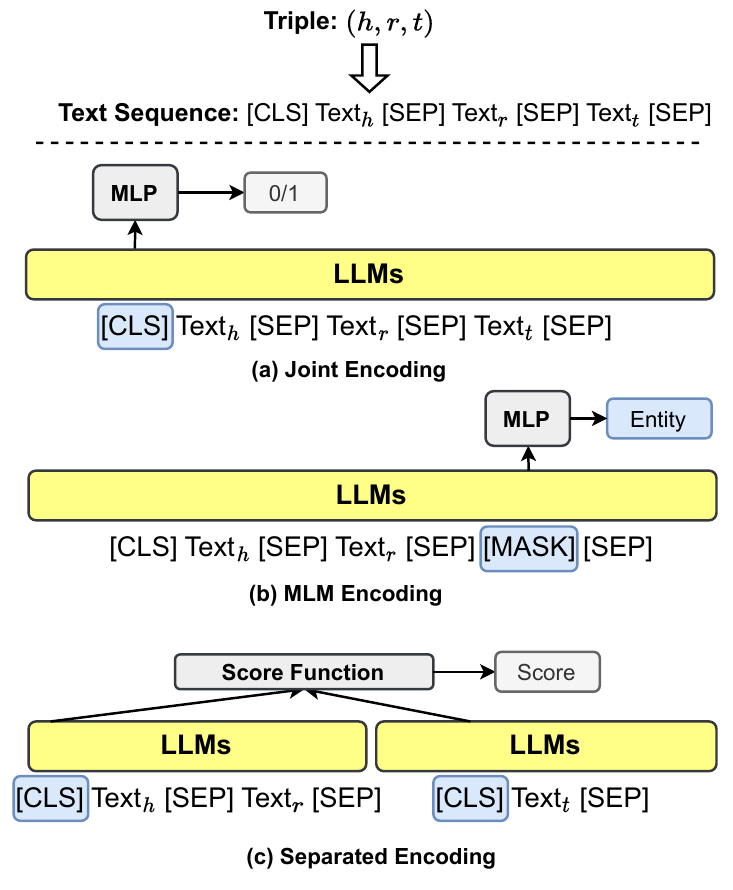}
    \caption{The general framework of adopting LLMs as encoders (PaE) for KG Completion.}
    \label{fig:PaE}
\end{figure}

As shown in Fig. \ref{fig:PaE} (a), (b), and (c), this line of work first uses encoder-only LLMs to encode textual information as well as KG facts. Then, they predict the plausibility of the triples or masked entities by feeding the encoded representation into a prediction head, which could be a simple MLP or conventional KG score function (e.g., TransE \cite{bordes2013translating} and TransR \cite{lin2015learning}).

\textbf{Joint Encoding.}
Since the encoder-only LLMs (e.g., Bert \cite{devlin2018bert}) are well at encoding text sequences, KG-BERT~\cite{yao2019kg} represents a triple $(h, r, t)$ as a text sequence and encodes it with LLM Fig.~\ref{fig:PaE}(a).
\begin{equation}
    x = \texttt{[CLS]} \ \text{Text}_h \ \texttt{[SEP]} \ \text{Text}_r \ \texttt{[SEP]} \ \text{Text}_t \ \texttt{[SEP]},
\end{equation}
The final hidden state of the $\texttt{[CLS]}$ token is fed into a classifier to predict the possibility of the triple, formulated as
\begin{equation}
    s = \sigma(\text{MLP}(e_{\texttt{[CLS]}})),
\end{equation}
where $\sigma(\cdot)$ denotes the sigmoid function and $e_{\texttt{[CLS]}}$ denotes the representation encoded by LLMs.
To improve the efficacy of KG-BERT, MTL-KGC~\cite{MTL-KGC} proposed a Multi-Task Learning for the KGC framework which incorporates additional auxiliary tasks into the model's training, i.e. prediction (RP) and relevance ranking (RR). PKGC~\cite{PKGC} assesses the validity of a triplet $(h, r, t)$ by transforming the triple and its supporting information into natural language sentences with pre-defined templates. These sentences are then processed by LLMs for binary classification. The supporting information of the triplet is derived from the attributes of $h$ and $t$ with a verbalizing function. For instance, if the triple is \textit{(Lebron James, member of sports team, Lakers)}, the information regarding Lebron James is verbalized as "Lebron James: American basketball player". LASS~\cite{LASS} observes that language semantics and graph structures are equally vital to KGC. As a result, LASS is proposed to jointly learn two types of embeddings: semantic embedding and structure embedding. In this method, the full text of a triple is forwarded to the LLM, and the mean pooling of the corresponding LLM outputs for $h$, $r$, and $t$ are separately calculated. These embeddings are then passed to a graph-based method, i.e. TransE, to reconstruct the KG structures.

\textbf{MLM Encoding.} Instead of encoding the full text of a triple, many works introduce the concept of Masked Language Model (MLM) to encode KG text (Fig.~\ref{fig:PaE}(b)). MEM-KGC~\cite{MEM-KGC} uses Masked Entity Model (MEM) classification mechanism to predict the masked entities of the triple. The input text is in the form of
\begin{equation}
    x = \texttt{[CLS]} \ \text{Text}_h \ \texttt{[SEP]} \ \text{Text}_r  \ \texttt{[SEP]} \ \texttt{[MASK]} \ \texttt{[SEP]},\label{eq:mlm_masking}
\end{equation}
Similar to Eq. \ref{eq:LLM_kge_masking}, it tries to maximize the probability that the masked entity is the correct entity $t$.
Additionally, to enable the model to learn unseen entities, MEM-KGC integrates multitask learning for entities and super-class prediction based on the text description of entities:
\begin{equation}
    x = \texttt{[CLS]} \ \texttt{[MASK]} \ \texttt{[SEP]} \ \text{Text}_h \ \texttt{[SEP]}. \label{eq:edp}
\end{equation}
OpenWorld KGC \cite{open-world-KGC} expands the MEM-KGC model to address the challenges of open-world KGC with a pipeline framework, where two sequential MLM-based modules are defined: Entity Description Prediction (EDP), an auxiliary module that predicts a corresponding entity with a given textual description; Incomplete Triple Prediction (ITP), the target module that predicts a plausible entity for a given incomplete triple $(h, r, ?)$. EDP first encodes the triple with Eq.~\ref{eq:edp} and generates the final hidden state, which is then forwarded into ITP as an embedding of the head entity in Eq.~\ref{eq:mlm_masking} to predict target entities.

\textbf{Separated Encoding.} As shown in Fig.~\ref{fig:PaE}(c), these methods involve partitioning a triple $(h, r, t)$ into two distinct parts, i.e. $(h, r)$ and $t$, which can be expressed as
\begin{align}
    x_{(h,r)} & = \texttt{[CLS]} \ \text{Text}_h \ \texttt{[SEP]} \ \text{Text}_r \ \texttt{[SEP]}, \\
    x_{t}     & = \texttt{[CLS]} \ \text{Text}_t \ \texttt{[SEP]}.
\end{align}
Then the two parts are encoded separately by LLMs, and the final hidden states of the $\texttt{[CLS]}$ tokens are used as the representations of $(h, r)$ and $t$, respectively. The representations are then fed into a scoring function to predict the possibility of the triple, formulated as
\begin{equation}
    s = f_{score} (e_{(h,r)}, e_{t}),
\end{equation}
where $f_{score}$ denotes the score function like TransE.

StAR~\cite{StAR} applies Siamese-style textual encoders on their text, encoding them into separate contextualized representations. To avoid the combinatorial explosion of textual encoding approaches, e.g., KG-BERT, StAR employs a scoring module that involves both deterministic classifier and spatial measurement for representation and structure learning respectively, which also enhances structured knowledge by exploring the spatial characteristics. SimKGC~\cite{SimKGC} is another instance of leveraging a Siamese textual encoder to encode textual representations. Following the encoding process, SimKGC applies contrastive learning techniques to these representations. This process involves computing the similarity between the encoded representations of a given triple and its positive and negative samples. In particular, the similarity between the encoded representation of the triple and the positive sample is maximized, while the similarity between the encoded representation of the triple and the negative sample is minimized. This enables SimKGC to learn a representation space that separates plausible and implausible triples. To avoid overfitting textural information, CSPromp-KG~\cite{CSProm-KG} employs parameter-efficient prompt learning for KGC.

LP-BERT \cite{LP-BERT} is a hybrid KGC method that combines both MLM Encoding and Separated Encoding. This approach consists of two stages, namely pre-training and fine-tuning. During pre-training, the method utilizes the standard MLM mechanism to pre-train a LLM with KGC data. During the fine-tuning stage, the LLM encodes both parts and is optimized using a contrastive learning strategy (similar to SimKGC~\cite{SimKGC}).

\subsubsection{LLM as Generators (PaG).}

\begin{figure}[]
    \centering
    \includegraphics[width=.8\columnwidth]{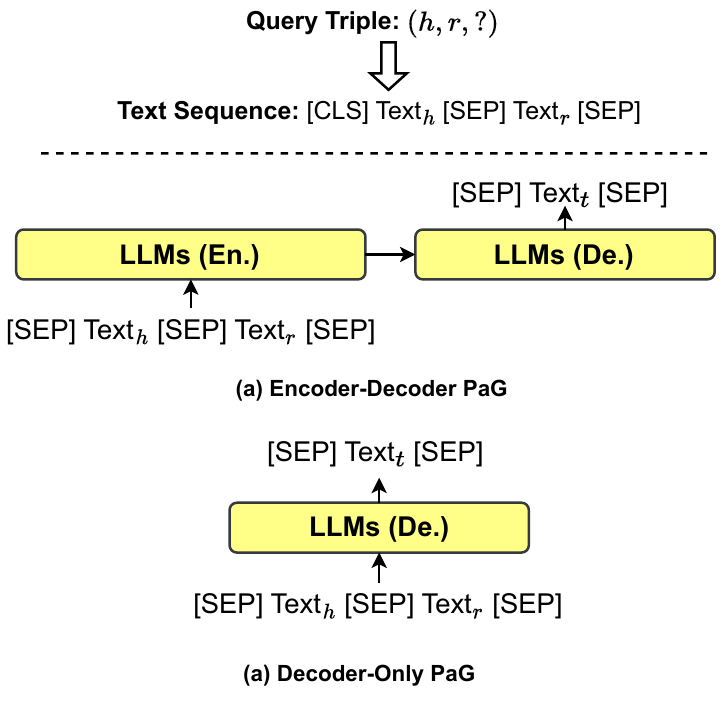}
    \caption{The general framework of adopting LLMs as decoders (PaG) for KG Completion. The En. and De. denote the encoder and decoder, respectively.}
    \label{fig:PaG}
\end{figure}

Recent works use LLMs as sequence-to-sequence generators in KGC. As presented in Fig.~\ref{fig:PaG} (a) and (b), these approaches involve encoder-decoder or decoder-only LLMs. The LLMs receive a sequence text input of the query triple $(h, r, ?)$, and generate the text of tail entity $t$ directly.

\begin{figure}[]
    \centering
    \includegraphics[width=.7\columnwidth]{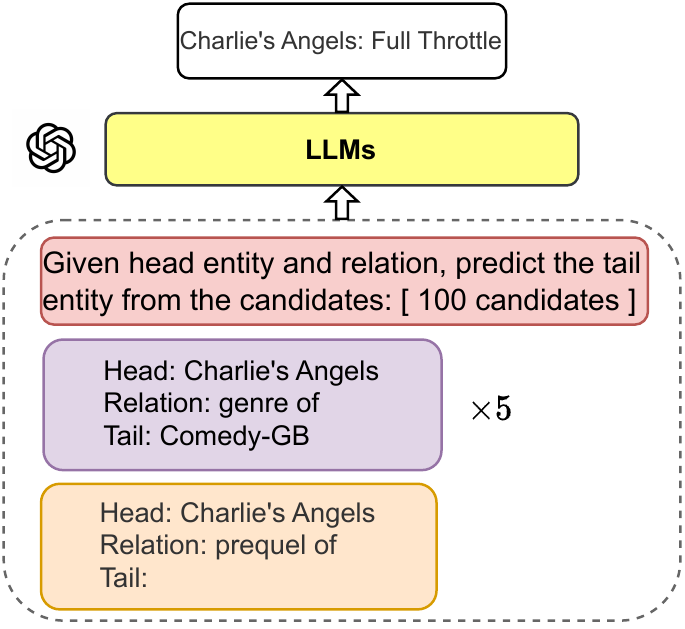}
    \caption{The framework of prompt-based PaG for KG Completion.}
    \label{fig:prompt_kgc}
\end{figure}

GenKGC~\cite{GenKGC} uses the large language model BART~\cite{lewis2020bart} as the backbone model. Inspired by the in-context learning approach used in GPT-3~\cite{brown2020language}, where the model concatenates relevant samples to learn correct output answers, GenKGC proposes a relation-guided demonstration technique that includes triples with the same relation to facilitating the model's learning process. In addition, during generation, an entity-aware hierarchical decoding method is proposed to reduce the time complexity. KGT5~\cite{KGT5} introduces a novel KGC model that fulfils four key requirements of such models: scalability, quality, versatility, and simplicity. To address these objectives, the proposed model employs a straightforward T5 small architecture. The model is distinct from previous KGC methods, in which it is randomly initialized rather than using pre-trained models. KG-S2S~\cite{KG-S2S} is a comprehensive framework that can be applied to various types of KGC tasks, including Static KGC, Temporal KGC, and Few-shot KGC. To achieve this objective, KG-S2S reformulates the standard triple KG fact by introducing an additional element, forming a quadruple $(h, r, t, m)$, where $m$ represents the additional "condition" element. Although different KGC tasks may refer to different conditions, they typically have a similar textual format, which enables unification across different KGC tasks. The KG-S2S approach incorporates various techniques such as entity description, soft prompt, and Seq2Seq Dropout to improve the model's performance. In addition, it utilizes constrained decoding to ensure the generated entities are valid. For closed-source LLMs (e.g., ChatGPT and GPT-4), AutoKG adopts prompt engineering to design customized prompts \cite{zhu2023llms}. As shown in Fig. \ref{fig:prompt_kgc}, these prompts contain the task description, few-shot examples, and test input, which instruct LLMs to predict the tail entity for KG completion.

\textbf{Comparison between PaE and PaG.}
LLMs as Encoders (PaE) applies an additional prediction head on the top of the representation encoded by LLMs. Therefore, the PaE framework is much easier to finetune since we can only optimize the prediction heads and freeze the LLMs. Moreover, the output of the prediction can be easily specified and integrated with existing KGC functions for different KGC tasks. However, during the inference stage, the PaE requires to compute a score for every candidate in KGs, which could be computationally expensive. Besides, they cannot generalize to unseen entities. Furthermore, the PaE requires the representation output of the LLMs, whereas some state-of-the-art LLMs (e.g. GPT-4\footnotemark[1]) are closed sources and do not grant access to the representation output.

LLMs as Generators (PaG), on the other hand, which does not need the prediction head, can be used without finetuning or access to representations. Therefore, the framework of PaG is suitable for all kinds of LLMs. In addition, PaG directly generates the tail entity, making it efficient in inference without ranking all the candidates and easily generalizing to unseen entities. But, the challenge of PaG is that the generated entities could be diverse and not lie in KGs. What is more, the time of a single inference is longer due to the auto-regressive generation. Last, how to design a powerful prompt that feeds KGs into LLMs is still an open question.
Consequently, while PaG has demonstrated promising results for KGC tasks, the trade-off between model complexity and computational efficiency must be carefully considered when selecting an appropriate LLM-based KGC framework.



\subsubsection{Model Analysis}
Justin et al. \cite{KGC_analysis} provide a comprehensive analysis of KGC methods integrated with LLMs. Their research investigates  the quality of LLM embeddings and finds that they are suboptimal for effective entity ranking. In response, they propose several techniques for processing embeddings to improve their suitability for candidate retrieval. The study also compares different model selection dimensions, such as Embedding Extraction, Query Entity Extraction, and Language Model Selection. Lastly, the authors propose a framework that effectively adapts LLM for knowledge graph completion.

\subsection{LLM-augmented KG Construction}

\begin{figure}[]
    \centering
    \includegraphics[width=\columnwidth]{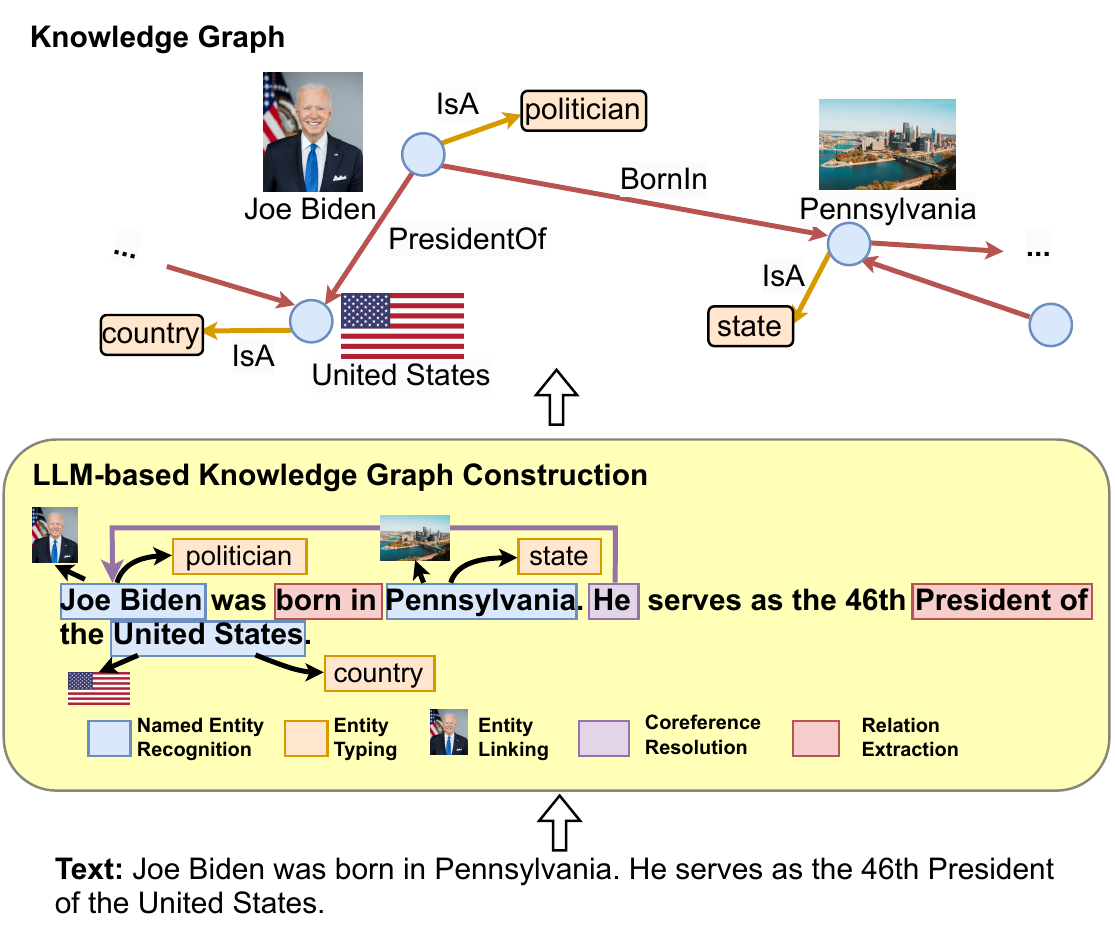}
    \caption{The general framework of LLM-based KG construction.}
    \label{fig: KGConsturction}
\end{figure}

Knowledge graph construction involves creating a structured representation of knowledge within a specific domain. This includes identifying entities and their relationships with each other. The process of knowledge graph construction typically involves multiple stages, including \emph{1) entity discovery}, \emph{2) coreference resolution}, and \emph{3) relation extraction}. Fig~\ref{fig: KGConsturction} presents the general framework of applying LLMs for each stage in KG construction. More recent approaches have explored \emph{4) end-to-end knowledge graph construction}, which involves constructing a complete knowledge graph in one step or directly \emph{5) distilling knowledge graphs from LLMs}.

\subsubsection{Entity Discovery}
Entity discovery in KG construction refers to the process of identifying and extracting entities from unstructured data sources, such as text documents, web pages, or social media posts, and incorporating them to construct knowledge graphs.

\textbf{Named Entity Recognition (NER)} involves identifying and tagging named entities in text data with their positions and classifications. The named entities include people, organizations, locations, and other types of entities. The state-of-the-art NER methods usually employ LLMs to leverage their contextual understanding and linguistic knowledge for accurate entity recognition and classification. There are three NER sub-tasks based on the types of NER spans identified, i.e., flat NER, nested NER, and discontinuous NER. \emph{1) Flat NER is to identify non-overlapping named entities from input text.} It is usually conceptualized as a sequence labelling problem where each token in the text is assigned a unique label based on its position in the sequence~\cite{Elmo,devlin2018bert,Larger-Context-Tagging,Ptuning-v2}. \emph{2) Nested NER considers complex scenarios which allow a token to belong to multiple entities.} The span-based method~\cite{NER-as-DP,DiscontinuousNER,Span-based-NER1,Span-based-NER2,Span-based-NER3} is a popular branch of nested NER which involves enumerating all candidate spans and classifying them into entity types (including a non-entity type). Parsing-based methods~\cite{parserNER1,parserNER2,parserNER3} reveal similarities between nested NER and constituency parsing tasks (predicting nested and non-overlapping spans), and propose to integrate the insights of constituency parsing into nested NER.
\emph{3) Discontinuous NER identifies named entities that may not be contiguous in the text.} To address this challenge, \cite{discontinuousNER1} uses the LLM output to identify entity fragments and determine whether they are overlapped or in succession.


Unlike the task-specific methods, GenerativeNER \cite{generativeNER} uses a sequence-to-sequence LLM with a pointer mechanism to generate an entity sequence, which is capable of solving all three types of NER sub-tasks.

\textbf{Entity Typing (ET)} aims to provide fine-grained and ultra-grained type information for a given entity mentioned in context. These methods usually utilize LLM to encode mentions, context and types. LDET~\cite{LDET} applies pre-trained ELMo embeddings~\cite{Elmo} for word representation and adopts LSTM as its sentence and mention encoders. BOX4Types~\cite{BOX4Types} recognizes the importance of type dependency and uses BERT to represent the hidden vector and each type in a hyperrectangular (box) space. LRN~\cite{LRN} considers extrinsic and intrinsic dependencies between labels. It encodes the context and entity with BERT and employs these output embeddings to conduct deductive and inductive reasoning. MLMET~\cite{MLMET} uses predefined patterns to construct input samples for the BERT MLM and employs [MASK] to predict context-dependent hypernyms of the mention, which can be viewed as type labels. 
PL~\cite{PL} and DFET~\cite{DFET} utilize prompt learning for entity typing. LITE~\cite{LITE} formulates entity typing as textual inference and uses RoBERTa-large-MNLI as the backbone network. 

\textbf{Entity Linking (EL)}, as known as entity disambiguation, involves linking entity mentions appearing in the text to their corresponding entities in a knowledge graph.  \cite{EL1} proposed BERT-based end-to-end EL systems that jointly discover and link entities. ELQ~\cite{ELQ} employs a fast bi-encoder architecture to jointly perform mention detection and linking in one pass for downstream question answering systems. 
Unlike previous models that frame EL as matching in vector space, GENRE~\cite{GENRE} formulates it as a sequence-to-sequence problem, autoregressively generating a version of the input markup-annotated with the unique identifiers of an entity expressed in natural language. GENRE is extended to its multilingual version mGENRE~\cite{mGENRE}. 
Considering the efficiency challenges of generative EL approaches, \cite{EL2} parallelizes autoregressive linking across all potential mentions and relies on a shallow and efficient decoder. 
ReFinED~\cite{ReFinED} proposes an efficient zero-shot-capable EL approach by taking advantage of fine-grained entity types and entity descriptions which are processed by a LLM-based encoder. 

\subsubsection{Coreference Resolution (CR)}
Coreference resolution is to find all expressions (i.e., mentions) that refer to the same entity or event in a text.

\textbf{Within-document CR} refers to the CR sub-task where all these mentions are in a single document. Mandar et al. \cite{CR1} initialize LLM-based coreferences resolution by replacing the previous LSTM encoder~\cite{LSTM-CR} with BERT. This work is followed by the introduction of SpanBERT~\cite{SpanBERT} which is pre-trained on BERT architecture with a span-based masked language model (MLM). Inspired by these works, Tuan Manh et al. \cite{CR2} present a strong baseline by incorporating the SpanBERT encoder into a non-LLM approach e2e-coref~\cite{LSTM-CR}. CorefBERT leverages Mention Reference Prediction (MRP) task which masks one or several mentions and requires the model to predict the masked mention’s corresponding referents. CorefQA~\cite{CorefQA} formulates coreference resolution as a question answering task, where contextual queries are generated for each candidate mention and the coreferent spans are extracted from the document using the queries.
Tuan Manh et al. \cite{CR3} introduce a gating mechanism  and a noisy training method to extract information from event mentions using the SpanBERT encoder.

In order to reduce the large memory footprint faced by large LLM-based NER models, Yuval et al. \cite{efficientCR1} and Raghuveer el al. \cite{efficientCR2} proposed start-to-end and approximation models, respectively, both utilizing bilinear functions to calculate mention and antecedent scores with reduced reliance on span-level representations.

\textbf{Cross-document CR} refers to the sub-task where the mentions refer to the same entity or event might be across multiple documents. CDML \cite{CDLM} proposes a cross document language modeling method which pre-trains a Longformer~\cite{Longformer} encoder on concatenated related documents and employs an MLP for binary classification to determine whether a pair of mentions is coreferent or not. CrossCR \cite{crossCR} utilizes an end-to-end model for cross-document coreference resolution which pre-trained the mention scorer on gold mention spans and uses a pairwise scorer to compare mentions with all spans across all documents. CR-RL \cite{CR-RL} proposes an actor-critic deep reinforcement learning-based coreference resolver for cross-document CR.


\subsubsection{Relation Extraction (RE)}
Relation extraction involves identifying semantic relationships between entities mentioned in natural language text. There are two types of relation extraction methods, i.e. sentence-level RE and document-level RE, according to the scope of the text analyzed.

\textbf{Sentence-level RE} focuses on identifying relations between entities within a single sentence. Peng et al. \cite{sent-re1} and TRE~\cite{TRE} introduce LLM to improve the performance of relation extraction models. BERT-MTB~\cite{BERT-MTB} learns relation representations based on BERT by performing the matching-the-blanks task and incorporating designed objectives for relation extraction. Curriculum-RE~\cite{Curriculum-RE} utilizes curriculum learning to improve relation extraction models by gradually increasing the difficulty of the data during training.
RECENT~\cite{RECENT} introduces SpanBERT and exploits entity type restriction to reduce the noisy candidate relation types. Jiewen~\cite{sent-re3} extends RECENT by combining both the entity information and the label information into sentence-level embeddings, which enables the embedding to be entity-label aware.

\textbf{Document-level RE (DocRE)} aims to extract relations between entities across multiple sentences within a document. Hong et al. \cite{docRE1} propose a strong baseline for DocRE by replacing the BiLSTM backbone with LLMs. HIN~\cite{HIN} use LLM to encode and aggregate entity representation at different levels, including entity, sentence, and document levels.  GLRE~\cite{GLRE} is a global-to-local network, which uses LLM to encode the document information in terms of entity global and local representations as well as context relation representations. SIRE~\cite{SIRE} uses two LLM-based encoders to extract intra-sentence and inter-sentence relations. LSR~\cite{LSR} and GAIN~\cite{GAIN} propose graph-based approaches which induce graph structures on top of LLM to better extract relations. DocuNet~\cite{DocuNet} formulates DocRE as a semantic segmentation task and introduces a U-Net~\cite{U-net} on the LLM encoder to capture local and global dependencies between entities. ATLOP \cite{ATLOP} focuses on the multi-label problems in DocRE, which could be handled with two techniques, i.e., adaptive thresholding for classifier and localized context pooling for LLM. DREEAM~\cite{DREEAM} further extends and improves ATLOP by incorporating evidence information.

\noindent\textbf{End-to-End KG Construction.}
Currently, researchers are exploring the use of LLMs for end-to-end KG construction. Kumar et al. \cite{kumar2020building} propose a unified approach to build KGs from raw text, which contains two LLMs powered components. They first finetune a LLM on named entity recognition tasks to make it capable of recognizing entities in raw text. Then, they propose another ``2-model BERT'' for solving the relation extraction task, which contains two BERT-based classifiers. The first classifier learns the relation class whereas the second binary classifier learns the direction of the relations between the two entities. The predicted triples and relations are then used to construct the KG. Guo et al. \cite{guo2021constructing} propose an end-to-end knowledge extraction model based on BERT, which can be applied to construct KGs from Classical Chinese text.
Grapher \cite{melnyk2021grapher} presents a novel end-to-end multi-stage system. It first utilizes LLMs to generate KG entities, followed by a simple relation construction head, enabling efficient KG construction from the textual description. PiVE \cite{han2023pive} proposes a prompting with an iterative verification framework that utilizes a smaller LLM like T5 to correct the errors in KGs generated by a larger LLM (e.g., ChatGPT). To further explore advanced LLMs, AutoKG design several prompts for different KG construction tasks (e.g., entity typing, entity linking, and relation extraction). Then, it adopts the prompt to perform KG construction using ChatGPT and GPT-4.


\begin{figure}[]
    \centering
    \includegraphics[width=1\columnwidth]{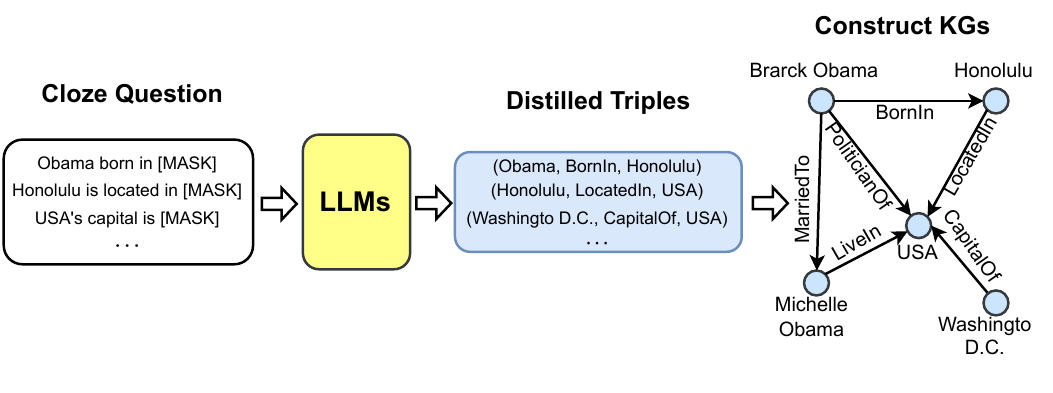}
    \caption{The general framework of distilling KGs from LLMs.}
    \label{fig:distill_kgs}
\end{figure}

\subsubsection{Distilling Knowledge Graphs from LLMs }
LLMs have been shown to implicitly encode massive knowledge \cite{petroni2019language}. As shown in Fig. \ref{fig:distill_kgs}, some research aims to distill knowledge from LLMs to construct KGs.  COMET \cite{bosselut2019comet} proposes a commonsense transformer model that constructs commonsense KGs by using existing tuples as a seed set of knowledge on which to train. Using this seed set, a LLM learns to adapt its learned representations to knowledge generation, and produces novel tuples that are high quality. Experimental results reveal that implicit knowledge from LLMs is transferred to generate explicit knowledge in commonsense KGs. BertNet \cite{hao2022bertnet} proposes a novel framework for automatic KG construction empowered by LLMs. It requires only the minimal definition of relations as inputs and automatically generates diverse prompts, and performs an efficient knowledge search within a given LLM for consistent outputs. The constructed KGs show competitive quality, diversity, and novelty with a richer set of new and complex relations, which cannot be extracted by previous methods. West et al. \cite{west2022symbolic} propose a symbolic knowledge distillation framework that distills symbolic knowledge from LLMs. They first finetune a small student LLM by distilling commonsense facts from a large LLM like GPT-3. Then, the student LLM is utilized to generate commonsense KGs.

\subsection{LLM-augmented KG-to-text Generation}
The goal of Knowledge-graph-to-text (KG-to-text) generation is to generate high-quality texts that accurately and consistently describe the input knowledge graph information~\cite{gardent-etal-2017-webnlg}. KG-to-text generation connects knowledge graphs and texts, significantly improving the applicability of KG in more realistic NLG scenarios, including storytelling~\cite{DBLP:conf/aaai/GuanWH19} and knowledge-grounded dialogue~\cite{DBLP:conf/ijcai/ZhouYHZXZ18}. However, it is challenging and costly to collect large amounts of graph-text parallel data, resulting in insufficient training and poor generation quality. Thus, many research efforts resort to either: \emph{1) leverage knowledge from LLMs} or \emph{2) construct large-scale weakly-supervised KG-text corpus} to solve this issue.

\subsubsection{Leveraging Knowledge from LLMs}
\begin{figure}[]
    \centering
    \includegraphics[width=1\columnwidth]{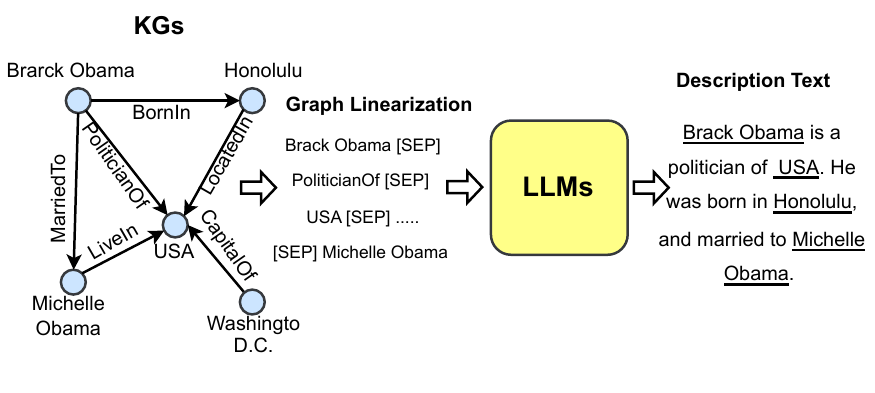}
    \caption{The general framework of KG-to-text generation.}
    \label{fig:kg_to_text}
\end{figure}

As pioneering research efforts in using LLMs for KG-to-Text generation, Ribeiro et al.~\cite{ribeiro-etal-2021-investigating} and Kale and Rastogi~\cite{kale-rastogi-2020-text} directly fine-tune various LLMs, including BART and T5, with the goal of transferring LLMs knowledge for this task. As shown in Fig. \ref{fig:kg_to_text}, both works simply represent the input graph as a linear traversal and find that such a naive approach successfully outperforms many existing state-of-the-art KG-to-text generation systems. Interestingly, Ribeiro et al.~\cite{ribeiro-etal-2021-investigating} also find that continue pre-training could further improve model performance. However, these methods are unable to \emph{explicitly} incorporate rich graph semantics in KGs. To enhance LLMs with KG structure information, JointGT~\cite{ke-etal-2021-jointgt} proposes to inject KG structure-preserving representations into the Seq2Seq large language models. Given input sub-KGs and corresponding text, JointGT first represents the KG entities and their relations as a sequence of tokens, then concatenate them with the textual tokens which are fed into LLM. After the standard self-attention module, JointGT then uses a pooling layer to obtain the contextual semantic representations of knowledge entities and relations. Finally, these pooled KG representations are then aggregated in another structure-aware self-attention layer. JointGT also deploys additional pre-training objectives, including KG and text reconstruction tasks given masked inputs, to improve the alignment between text and graph information.  Li et al.~\cite{li-etal-2021-shot-knowledge} focus on the few-shot scenario. It first employs a novel breadth-first search (BFS) strategy to better traverse the input KG structure and feed the enhanced linearized graph representations into LLMs for high-quality generated outputs, then aligns the GCN-based and LLM-based KG entity representation. Colas et al.~\cite{colas-etal-2022-gap} first transform the graph into its appropriate representation before linearizing the graph. Next, each KG node is encoded via a global attention mechanism, followed by a graph-aware attention module, ultimately being decoded into a sequence of tokens. Different from these works, KG-BART~\cite{DBLP:conf/aaai/LiuW0PY21} keeps the structure of KGs and leverages the graph attention to aggregate the rich concept semantics in the sub-KG, which enhances the model generalization on unseen concept sets.

\subsubsection{Constructing large weakly KG-text aligned Corpus}
Although LLMs have achieved remarkable empirical success, their unsupervised pre-training objectives are not necessarily aligned well with the task of KG-to-text generation, motivating researchers to develop large-scale KG-text aligned corpus. Jin et al.~\cite{jin-etal-2020-genwiki} propose a 1.3M unsupervised KG-to-graph training data from Wikipedia. Specifically, they first detect the entities appearing in the text via hyperlinks and named entity detectors, and then only add text that shares a common set of entities with the corresponding knowledge graph, similar to the idea of distance supervision in the relation extraction task~\cite{mintz-etal-2009-distant}.
They also provide a 1,000+ human annotated KG-to-Text test data to verify the effectiveness of the pre-trained KG-to-Text models. Similarly, Chen et al.~\cite{chen-etal-2020-kgpt} also propose a KG-grounded text corpus collected from the English Wikidump. To ensure the connection between KG and text, they only extract sentences with at least two Wikipedia anchor links. Then, they use the entities from those links to query their surrounding neighbors in WikiData and calculate the lexical overlapping between these neighbors and the original sentences. Finally, only highly overlapped pairs are selected. The authors explore both graph-based and sequence-based encoders and identify their advantages in various different tasks and settings.

\subsection{LLM-augmented KG Question Answering}

\begin{figure}[]
    \centering
    \includegraphics[width=.8\columnwidth]{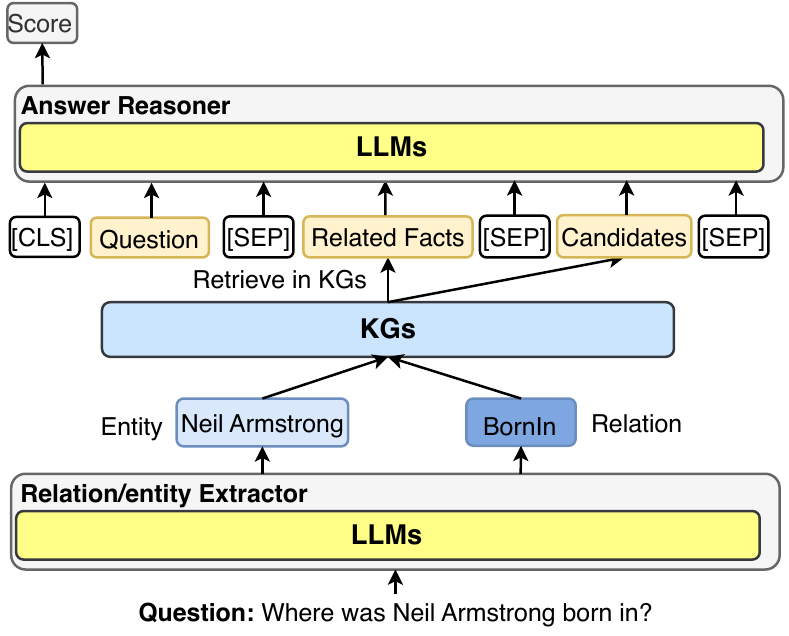}
    \caption{The general framework of applying LLMs for knowledge graph question answering (KGQA).}
    \label{fig:LLM_for_KGQA}
\end{figure}

Knowledge graph question answering (KGQA) aims to find answers to natural language questions based on the structured facts stored in knowledge graphs \cite{saxena2020improving,feng-etal-2020-scalable}. The inevitable challenge in KGQA is to retrieve related facts and extend the reasoning advantage of KGs to QA. Therefore, recent studies adopt LLMs to bridge the gap between natural language questions and structured knowledge graphs \cite{xu2021fusing,yan2021large,hu2023empirical}. The general framework of applying LLMs for KGQA is illustrated in Fig. \ref{fig:LLM_for_KGQA}, where LLMs can be used as 1) entity/relation extractors, and 2) answer reasoners.

\subsubsection{LLMs as Entity/relation Extractors}
Entity/relation extractors are designed to identify entities and relationships mentioned in natural language questions and retrieve related facts in KGs. Given the proficiency in language comprehension, LLMs can be effectively utilized for this purpose. Lukovnikov et al. \cite{lukovnikov2019pretrained} are the first to utilize LLMs as classifiers for relation prediction, resulting in a notable improvement in performance compared to shallow neural networks. Nan et al. \cite{hu2023empirical} introduce two LLM-based KGQA frameworks that adopt LLMs to detect mentioned entities and relations. Then, they query the answer in KGs using the extracted entity-relation pairs. QA-GNN \cite{yasunaga-etal-2021-qa} uses LLMs to encode the question and candidate answer pairs, which are adopted to estimate the importance of relative KG entities. The entities are retrieved to form a subgraph, where an answer reasoning is conducted by a graph neural network.
Luo et al. \cite{luo2020bert} use LLMs to calculate the similarities between relations and questions to retrieve related facts, formulated as
\begin{equation}
    s(r,q) = \text{LLM}(r)^\top \text{LLM}(q),\label{eq:relation_sim}
\end{equation}
where $q$ denotes the question, $r$ denotes the relation, and $\text{LLM}(\cdot)$ would generate representation for $q$ and $r$, respectively. Furthermore, Zhang et al. \cite{zhang2022subgraph} propose a LLM-based path retriever to retrieve question-related relations hop-by-hop and construct several paths. The probability of each path can be calculated as
\begin{equation}
    P(p|q) = \prod_{t=1}^{|p|} s(r_t,q),\label{eq:path_prob}
\end{equation}
where $p$ denotes the path, and $r_t$ denotes the relation at the $t$-th hop of $p$. The retrieved relations and paths can be used as context knowledge to improve the performance of answer reasoners as
\begin{equation}
    P(a|q) = \sum_{p\in \mathcal{P}} P(a|p)P(p|q),
\end{equation}
where $\mathcal{P}$ denotes retrieved paths and $a$ denotes the answer.

\subsubsection{LLMs as Answer Reasoners}
Answer reasoners are designed to reason over the retrieved facts and generate answers. LLMs can be used as answer reasoners to generate answers directly. For example, as shown in Fig. 3 \ref{fig:LLM_for_KGQA}, DEKCOR \cite{xu2021fusing} concatenates the retrieved facts with questions and candidate answers as
\begin{equation}
    x = \texttt{[CLS]} \  q \  \texttt{[SEP]} \  \text{Related Facts} \  \texttt{[SEP]} \  a \  \texttt{[SEP]},
\end{equation}
where $a$ denotes candidate answers. Then, it feeds them into LLMs to predict answer scores. After utilizing LLMs to generate the representation of $x$ as QA context, DRLK \cite{zhang2022drlk} proposes a Dynamic Hierarchical Reasoner to capture the interactions between QA context and answers for answer prediction. Yan et al. \cite{yan2021large} propose a LLM-based KGQA framework consisting of two stages: (1) retrieve related facts from KGs and (2) generate answers based on the retrieved facts. The first stage is similar to the entity/relation extractors. Given a candidate answer entity $a$, it extracts a series of paths $p_1,\ldots,p_n$ from KGs. But the second stage is a LLM-based answer reasoner. It first verbalizes the paths by using the entity names and relation names in KGs. Then, it concatenates the question $q$ and all paths $p_1,\ldots,p_n$ to make an input sample as
\begin{equation}
    x = \texttt{[CLS]} \  q \  \texttt{[SEP]} \  p_1 \  \texttt{[SEP]} \  \cdots \  \texttt{[SEP]} \  p_n \  \texttt{[SEP]}.\label{eq:answer_reasoner}
\end{equation}
These paths are regarded as the related facts for the candidate answer $a$. Finally, it uses LLMs to predict whether the hypothesis: ``$a$ is the answer of $q$'' is supported by those facts, which is formulated as
\begin{align}
    e_\texttt{[CLS]} & = \text{LLM}(x),                                                        \\
    s                & = \sigma(\text{MLP}(e_\texttt{[CLS]})),\label{eq:answer_reasoner_score}
\end{align}
where it encodes $x$ using a LLM and feeds representation corresponding to $\texttt{[CLS]}$ token for binary classification, and $\sigma(\cdot)$ denotes the sigmoid function.

To better guide LLMs reason through KGs, OreoLM \cite{hu2022empowering} proposes a Knowledge Interaction Layer (KIL) which is inserted amid LLM layers. KIL interacts with a KG reasoning module, where it discovers different reasoning paths, and then the reasoning module can reason over the paths to generate answers. GreaseLM \cite{zhang2022greaselm} fuses the representations from LLMs and graph neural networks to effectively reason over KG facts and language context. UniKGQA \cite{jiang2023unikgqa} unifies the facts retrieval and reasoning into a unified framework. UniKGQA consists of two modules. The first module is a semantic matching module that uses a LLM to match questions with their corresponding relations semantically. The second module is a matching information propagation module, which propagates the matching information along directed edges on KGs for answer reasoning. Similarly, ReLMKG \cite{cao2022relmkg} performs joint reasoning on a large language model and the associated knowledge graph. The question and verbalized paths are encoded by the language model, and different layers of the language model produce outputs that guide a graph neural network to perform message passing. This process utilizes the explicit knowledge contained in the structured knowledge graph for reasoning purposes. StructGPT \cite{jiang2023structgpt} adopts a customized interface to allow large language models (e.g., ChatGPT) directly reasoning on KGs to perform multi-step question answering.

\begin{table}[]
    \centering
    \caption{Summary of methods that synergize KGs and LLMs.}
    \label{tab:unify}
    \begin{tabular}{@{}c|cc@{}}
    \toprule
    Task & Method                              & Year \\ \midrule
    \multirow{4}{*}{Synergized Knowledge representation} & JointGT \cite{ke-etal-2021-jointgt}   & 2021 \\
                                              & KEPLER \cite{wang-etal-2021-kepler}          & 2021 \\
                                              & DRAGON \cite{yasunaga2022deep}        & 2022 \\
                                              & HKLM \cite{zhu2023pre}                & 2023 \\\midrule
    \multirow{5}{*}{Synergized Reasoning}  & LARK \cite{choudhary2023complex}      & 2023 \\
                                              & Siyuan et al. \cite{wang2023unifying} & 2023 \\
         & KSL \cite{feng2023knowledge}        & 2023 \\
         & StructGPT \cite{jiang2023structgpt} & 2023 \\
         & Think-on-graph \cite{sun2023think}  & 2023 \\ \bottomrule
    \end{tabular}%
\end{table}

\section{Synergized LLMs + KGs}
\label{sec:unification_LLM_kg}
The synergy of LLMs and KGs has attracted increasing attention these years, which marries the merits of LLMs and KGs to mutually enhance performance in various downstream applications. For example, LLMs can be used to understand natural language, while KGs are treated as a knowledge base, which provides factual knowledge. The unification of LLMs and KGs could result in a powerful model for knowledge representation and reasoning.

In this section, we will discuss the state-of-the-art \textit{Synergized LLMs + KGs} from two perspectives: \emph{1) Synergized Knowledge Representation}, and \emph{2)} Synergized Reasoning. Representative works are summarized in Table \ref{tab:unify}.

\begin{figure}[]
    \centering
    \includegraphics[width=0.7\columnwidth]{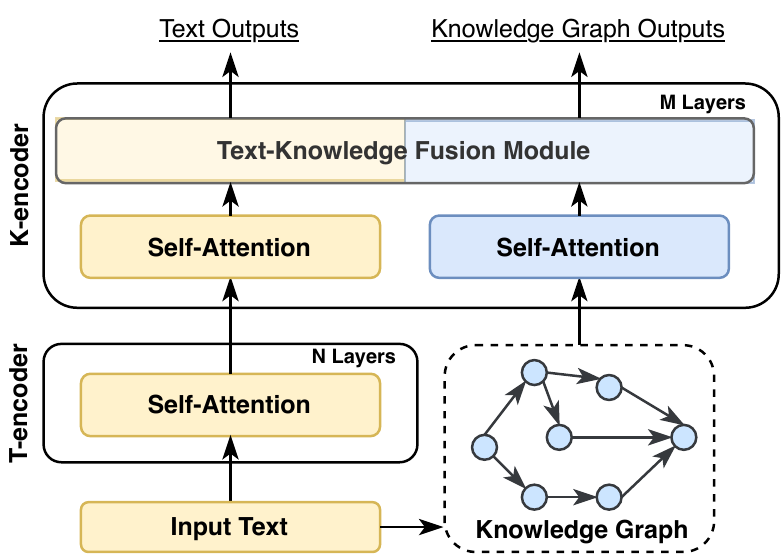}
    \caption{Synergized knowledge representation by additional KG fusion modules.}
    \label{fig:KG_module}
\end{figure}

\subsection{Synergized Knowledge Representation}\label{sec:synergized_model}
Text corpus and knowledge graphs both contain enormous knowledge. However, the knowledge in text corpus is usually implicit and unstructured, while the knowledge in KGs is explicit and structured. Synergized Knowledge Representation aims to design a synergized model that can effectively represent knowledge from both LLMs and KGs. The synergized model can provide a better understanding of the knowledge from both sources, making it valuable for many downstream tasks. 

To jointly represent the knowledge, researchers propose the synergized models by introducing additional KG fusion modules, which are jointly trained with LLMs. As shown in Fig.~\ref{fig:KG_module}, ERNIE~\cite{zhang-etal-2019-ernie} proposes a textual-knowledge dual encoder architecture where a \emph{T-encoder} first encodes the input sentences, then a \emph{K-encoder} processes knowledge graphs which are fused them with the textual representation from the \emph{T-encoder}. BERT-MK~\cite{he-etal-2020-bert} employs a similar dual-encoder architecture but it introduces additional information of neighboring entities in the knowledge encoder component during the pre-training of LLMs. However, some of the neighboring entities in KGs may not be relevant to the input text, resulting in extra redundancy and noise. CokeBERT~\cite{SU2021127} focuses on this issue and proposes a GNN-based module to filter out irrelevant KG entities using the input text. JAKET~\cite{DBLP:conf/aaai/Yu0Y022} proposes to fuse the entity information in the middle of the large language model.

KEPLER \cite{wang-etal-2021-kepler} presents a unified model for knowledge embedding and pre-trained language representation. In KEPLER, they encode textual entity descriptions with a LLM as their embeddings, and then jointly optimize the knowledge embedding and language modeling objectives. JointGT~\cite{ke-etal-2021-jointgt} proposes a graph-text joint representation learning model, which proposes three pre-training tasks to align representations of graph and text. DRAGON \cite{yasunaga2022deep} presents a self-supervised method to pre-train a joint language-knowledge foundation model from text and KG. It takes text segments and relevant KG subgraphs as input and bidirectionally fuses information from both modalities. Then, DRAGON utilizes two self-supervised reasoning tasks, i.e., masked language modeling and KG link prediction to optimize the model parameters. HKLM \cite{zhu2023pre} introduces a unified LLM which incorporates KGs to learn representations of domain-specific knowledge.


\subsection{Synergized Reasoning}
To better utilize the knowledge from text corpus and knowledge graph reasoning, Synergized Reasoning aims to design a synergized model that can effectively conduct reasoning with both LLMs and KGs. 

\begin{figure}[]
    \centering
    \includegraphics[width=\columnwidth]{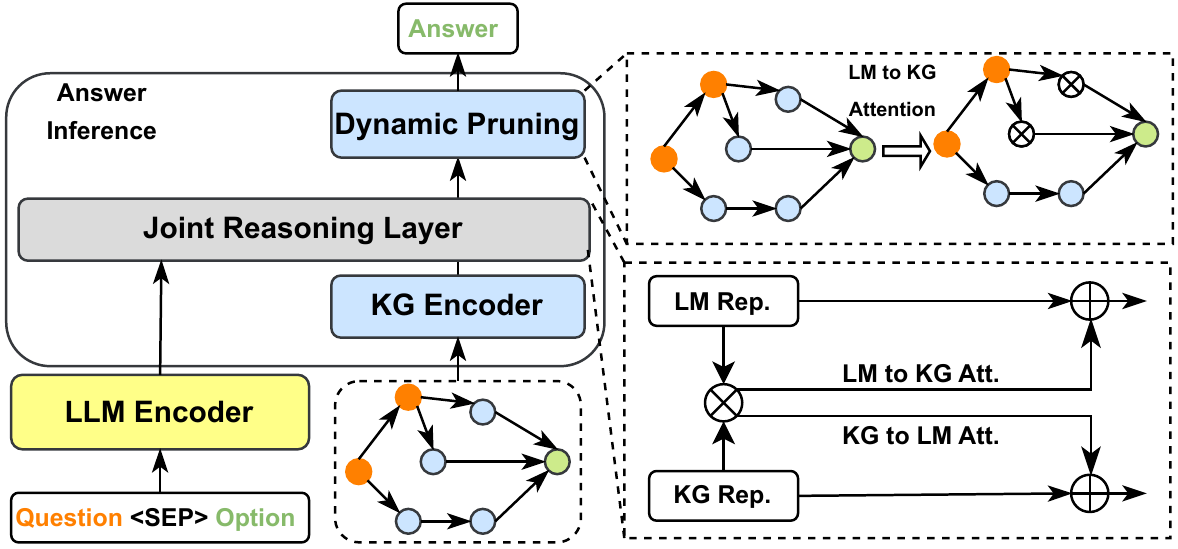}
    \caption{The framework of LLM-KG Fusion Reasoning.}
    \label{fig:jointlk}
\end{figure}

\noindent\textbf{LLM-KG Fusion Reasoning.}
LLM-KG Fusion Reasoning leverages two separated LLM and KG encoders to process the text and relevant KG inputs~\cite{DBLP:conf/aaai/WangKMYTACFMMW19}. These two encoders are equally important and jointly fusing the knowledge from two sources for reasoning. To improve the interaction between text and knowledge, KagNet~\cite{lin-etal-2019-kagnet} proposes to first encode the input KG, and then augment the input textual representation. In contrast, MHGRN~\cite{feng-etal-2020-scalable} uses the final LLM outputs of the input text to guide the reasoning process on the KGs. Yet, both of them only design a single-direction interaction between the text and KGs. To tackle this issue, QA-GNN~\cite{yasunaga-etal-2021-qa} proposes to use a GNN-based model to jointly reason over input context and KG information via message passing. Specifically, QA-GNN represents the input textual information as a special node via a pooling operation and connects this node with other entities in KG. However, the textual inputs are only pooled into a single dense vector, limiting the information fusion performance. JointLK~\cite{sun-etal-2022-jointlk} then proposes a framework with fine-grained interaction between any tokens in the textual inputs and any KG entities through LM-to-KG and KG-to-LM bi-directional attention mechanism. As shown in Fig.~\ref{fig:jointlk}, pairwise dot-product scores are calculated over all textual tokens and KG entities, the bi-directional attentive scores are computed separately. In addition, at each jointLK layer, the KGs are also dynamically pruned based on the attention score to allow later layers to focus on more important sub-KG structures. Despite being effective, in JointLK, the fusion process between the input text and KG still uses the final LLM outputs as the input text representations. GreaseLM~\cite{zhang2022greaselm} designs deep and rich interaction between the input text tokens and KG entities at each layer of the LLMs. The architecture and fusion approach is mostly similar to ERNIE~\cite{zhang-etal-2019-ernie} discussed in Section~\ref{sec:synergized_model}, except that GreaseLM does not use the text-only \emph{T-encoder} to handle input text.

\begin{figure}[]
    \centering
    \includegraphics[width=0.7\columnwidth]{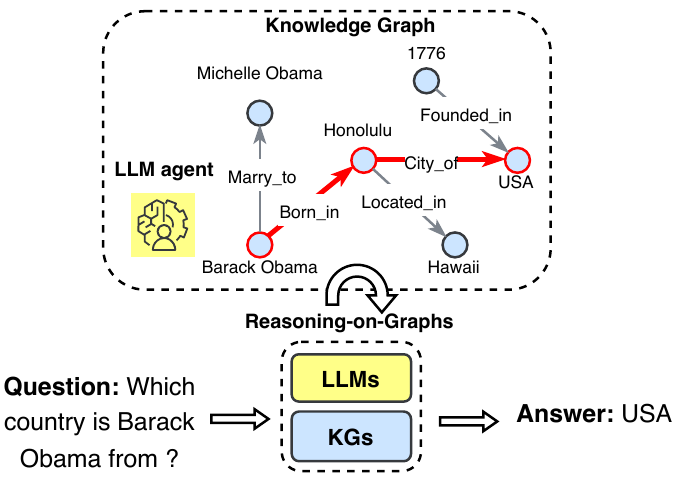}
    \caption{Using LLMs as agents for reasoning on KGs.}
    \label{fig:llm_agent}
\end{figure}

\noindent\textbf{LLMs as Agents Reasoning.}
Instead using two encoders to fuse the knowledge, LLMs can also be treated as agents to interact with the KGs to conduct reasoning \cite{liu2023agentbench}, as illustrated in Fig. \ref{fig:llm_agent}. KD-CoT \cite{wang2023knowledge} iteratively retrieves facts from KGs and produces faithful reasoning traces, which guide LLMs to generate answers.
KSL \cite{feng2023knowledge} teaches LLMs to search on KGs to retrieve relevant facts and then generate answers. StructGPT \cite{jiang2023structgpt} designs several API interfaces to allow LLMs to access the structural data and perform reasoning by traversing on KGs. Think-on-graph \cite{sun2023think} provides a flexible plug-and-play framework where LLM agents iteratively execute beam searches on KGs to discover the reasoning paths and generate answers. To enhance the agent abilities, AgentTuning \cite{zeng2023agenttuning} presents several instruction-tuning datasets to guide LLM agents to perform reasoning on KGs.

\noindent\textbf{Comparison and Discussion.}
LLM-KG Fusion Reasoning combines the LLM encoder and KG encoder to represent knowledge in a unified manner. It then employs a synergized reasoning module to jointly reason the results. This framework allows for different encoders and reasoning modules, which are trained end-to-end to effectively utilize the knowledge and reasoning capabilities of LLMs and KGs. However, these additional modules may introduce extra parameters and computational costs while lacking interpretability. LLMs as Agents for KG reasoning provides a flexible framework for reasoning on KGs without additional training cost, which can be generalized to different LLMs and KGs. Meanwhile, the reasoning process is interpretable, which can be used to explain the results. Nevertheless, defining the actions and policies for LLM agents is also challenging. The synergy of LLMs and KGs is still an ongoing research topic, with the potential to have more powerful frameworks in the future.

\section{Future Directions and Milestones}\label{sec:challenge}
In this section, we discuss the future directions and several milestones in the research area of unifying KGs and LLMs.

\subsection{KGs for Hallucination Detection in LLMs}
The hallucination problem in LLMs, which generates factually incorrect content, significantly hinders the reliability of LLMs. As discussed in Section \ref{sec:KG_for_LLM}, existing studies try to utilize KGs to obtain more reliable LLMs through pre-training or KG-enhanced inference. Despite the efforts, the issue of hallucination may continue to persist in the realm of LLMs for the foreseeable future. Consequently, in order to gain the public's trust and border applications, it is imperative to detect and assess instances of hallucination within LLMs and other forms of AI-generated content (AIGC). Existing methods strive to detect hallucination by training a neural classifier on a small set of documents \cite{kryscinski2019evaluating}, which are neither robust nor powerful to handle ever-growing LLMs. Recently, researchers try to use KGs as an external source to validate LLMs \cite{ji2022rho}. Further studies combine LLMs and KGs to achieve a generalized fact-checking model that can detect hallucinations across domains \cite{feng2023factkb}. Therefore, it opens a new door to utilizing KGs for hallucination detection.

\subsection{KGs for Editing Knowledge in LLMs}
Although LLMs are capable of storing massive real-world knowledge, they cannot quickly update their internal knowledge updated as real-world situations change. There are some research efforts proposed for editing knowledge in LLMs \cite{yao2023editing} without re-training the whole LLMs. Yet, such solutions still suffer from poor performance or computational overhead \cite{li2023unveiling}. Existing studies \cite{cohen2023evaluating} also reveal that edit a single fact would cause a ripple effect for other related knowledge. Therefore, it is necessary to develop a more efficient and effective method to edit knowledge in LLMs. Recently, researchers try to leverage KGs to edit knowledge in LLMs efficiently.

\subsection{KGs for Black-box LLMs Knowledge Injection}
Although pre-training and knowledge editing could update LLMs to catch up with the latest knowledge, they still need to access the internal structures and parameters of LLMs. However, many state-of-the-art large LLMs (e.g., ChatGPT) only provide APIs for users and developers to access, making themselves black-box to the public. Consequently, it is impossible to follow conventional KG injection approaches described \cite{DBLP:conf/aaai/WangKMYTACFMMW19,lin-etal-2019-kagnet} that change LLM structure by adding additional knowledge fusion modules. Converting various types of knowledge into different text prompts seems to be a feasible solution. However, it is unclear whether these prompts can generalize well to new LLMs. Moreover, the prompt-based approach is limited to the length of input tokens of LLMs. Therefore, how to enable effective knowledge injection for black-box LLMs is still an open question for us to explore \cite{diao2022black,sun2022black}.

\subsection{Multi-Modal LLMs for KGs}
Current knowledge graphs typically rely on textual and graph structure to handle KG-related applications. However, real-world knowledge graphs are often constructed by data from diverse modalities \cite{challenge-mm-1, challenge-mm-2, challenge-mm-3}. Therefore, effectively leveraging representations from multiple modalities would be a significant challenge for future research in KGs \cite{chen2022lako}. One potential solution is to develop methods that can accurately encode and align entities across different modalities. Recently, with the development of multi-modal LLMs \cite{girdhar2023imagebind,zhu2023minigpt}, leveraging LLMs for modality alignment holds promise in this regard. But, bridging the gap between multi-modal LLMs and KG structure remains a crucial challenge in this field, demanding further investigation and advancements.

\subsection{LLMs for Understanding KG Structure}
Conventional LLMs trained on plain text data are not designed to understand structured data like knowledge graphs. Thus, LLMs might not fully grasp or understand the information conveyed by the KG structure. A straightforward way is to linearize the structured data into a sentence that LLMs can understand. However, the scale of the KGs makes it impossible to linearize the whole KGs as input. Moreover, the linearization process may lose some underlying information in KGs. Therefore, it is necessary to develop LLMs that can directly understand the KG structure and reason over it \cite{jiang2023structgpt}.

\begin{figure}
    \centering
    \includegraphics[width=0.6\columnwidth]{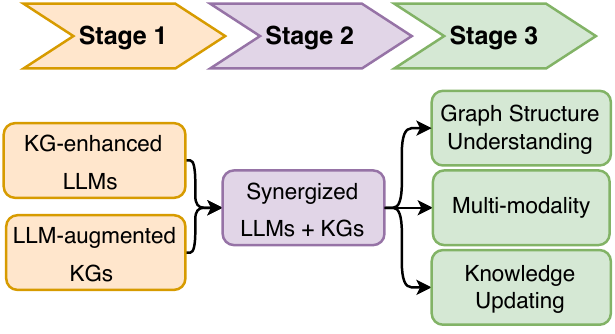}
    \caption{The milestones of unifying KGs and LLMs.}
    \label{fig:milestone}
\end{figure}

\subsection{Synergized LLMs and KGs for Birectional Reasoning}
KGs and LLMs are two complementary technologies that can synergize each other. However, the synergy of LLMs and KGs is less explored by existing researchers. A desired synergy of LLMs and KGs would involve leveraging the strengths of both technologies to overcome their individual limitations. LLMs, such as ChatGPT, excel in generating human-like text and understanding natural language, while KGs are structured databases that capture and represent knowledge in a structured manner. By combining their capabilities, we can create a powerful system that benefits from the contextual understanding of LLMs and the structured knowledge representation of KGs. To better unify LLMs and KGs, many advanced techniques need to be incorporated, such as multi-modal learning \cite{zhang2020emotion}, graph neural network \cite{zhang2022trustworthy}, and continuous learning \cite{wu2022pretrained}. Last, the synergy of LLMs and KGs can be applied to many real-world applications, such as search engines \cite{thoppilan2022lamda}, recommender systems \cite{liu2023chatgpt,sun2020multi}, and drug discovery. 

With a given application problem, we can apply a KG to perform a knowledge-driven search for potential goals and unseen data, and simultaneously start with LLMs to perform a data/text-driven inference to see what new data/goal items can be derived. When the knowledge-based search is combined with data/text-driven inference, they can mutually validate each other, resulting in efficient and effective solutions powered by dual-driving wheels.
Therefore, we can anticipate increasing attention to unlock the potential of integrating KGs and LLMs for diverse downstream applications with both generative and reasoning capabilities in the near future.


\section{Conclusion}\label{sec:conclusion}
Unifying large language models (LLMs) and knowledge graphs (KGs) is an active research direction that has attracted increasing attention from both academia and industry. In this article, we provide a thorough overview of the recent research in this field. We first introduce different manners that integrate KGs to enhance LLMs. Then, we introduce existing methods that apply LLMs for KGs and establish taxonomy based on varieties of KG tasks. Finally, we discuss the challenges and future directions in this field. 
We envision that there will be multiple stages (milestones) in the roadmap of unifying KGs and LLMs, as shown in Fig. \ref{fig:milestone}. In particular, we will anticipate increasing research on three stages: \emph{Stage 1}:  KG-enhanced LLMs, LLM-augmented KGs, \emph{Stage 2:} Synergized LLMs + KGs, and \emph{Stage 3:} Graph Structure Understanding, Multi-modality, Knowledge Updating. We hope that this article will provide a guideline to advance future research.
\vspace{-4mm}
\section*{Acknowledgments}
This research was supported by the Australian Research Council (ARC) under grants FT210100097 and DP240101547 and the National Natural Science Foundation of China (NSFC) under grant 62120106008.

\balance

\bibliographystyle{IEEEtran}
\bibliography{IEEEabrv,sections/main.bib}

\begin{thebibliography}{100}
\providecommand{\url}[1]{#1}
\csname url@samestyle\endcsname
\providecommand{\newblock}{\relax}
\providecommand{\bibinfo}[2]{#2}
\providecommand{\BIBentrySTDinterwordspacing}{\spaceskip=0pt\relax}
\providecommand{\BIBentryALTinterwordstretchfactor}{4}
\providecommand{\BIBentryALTinterwordspacing}{\spaceskip=\fontdimen2\font plus
\BIBentryALTinterwordstretchfactor\fontdimen3\font minus \fontdimen4\font\relax}
\providecommand{\BIBforeignlanguage}[2]{{%
\expandafter\ifx\csname l@#1\endcsname\relax
\typeout{** WARNING: IEEEtran.bst: No hyphenation pattern has been}%
\typeout{** loaded for the language `#1'. Using the pattern for}%
\typeout{** the default language instead.}%
\else
\language=\csname l@#1\endcsname
\fi
#2}}
\providecommand{\BIBdecl}{\relax}
\BIBdecl

\bibitem{devlin2018bert}
J.~Devlin, M.-W. Chang, K.~Lee, and K.~Toutanova, ``Bert: Pre-training of deep bidirectional transformers for language understanding,'' \emph{arXiv preprint arXiv:1810.04805}, 2018.

\bibitem{liu2019roberta}
Y.~Liu, M.~Ott, N.~Goyal, J.~Du, M.~Joshi, D.~Chen, O.~Levy, M.~Lewis, L.~Zettlemoyer, and V.~Stoyanov, ``Roberta: A robustly optimized bert pretraining approach,'' \emph{arXiv preprint arXiv:1907.11692}, 2019.

\bibitem{raffel2020exploring}
C.~Raffel, N.~Shazeer, A.~Roberts, K.~Lee, S.~Narang, M.~Matena, Y.~Zhou, W.~Li, and P.~J. Liu, ``Exploring the limits of transfer learning with a unified text-to-text transformer,'' \emph{The Journal of Machine Learning Research}, vol.~21, no.~1, pp. 5485--5551, 2020.

\bibitem{su2019generalizing}
D.~Su, Y.~Xu, G.~I. Winata, P.~Xu, H.~Kim, Z.~Liu, and P.~Fung, ``Generalizing question answering system with pre-trained language model fine-tuning,'' in \emph{Proceedings of the 2nd Workshop on Machine Reading for Question Answering}, 2019, pp. 203--211.

\bibitem{lewis2020bart}
M.~Lewis, Y.~Liu, N.~Goyal, M.~Ghazvininejad, A.~Mohamed, O.~Levy, V.~Stoyanov, and L.~Zettlemoyer, ``Bart: Denoising sequence-to-sequence pre-training for natural language generation, translation, and comprehension,'' in \emph{ACL}, 2020, pp. 7871--7880.

\bibitem{li2021pretrained}
J.~Li, T.~Tang, W.~X. Zhao, and J.-R. Wen, ``Pretrained language models for text generation: A survey,'' \emph{arXiv preprint arXiv:2105.10311}, 2021.

\bibitem{weiemergent}
J.~Wei, Y.~Tay, R.~Bommasani, C.~Raffel, B.~Zoph, S.~Borgeaud, D.~Yogatama, M.~Bosma, D.~Zhou, D.~Metzler \emph{et~al.}, ``Emergent abilities of large language models,'' \emph{Transactions on Machine Learning Research}.

\bibitem{malinka2023educational}
K.~Malinka, M.~Pere{\v{s}}{\'\i}ni, A.~Firc, O.~Huj{\v{n}}{\'a}k, and F.~Janu{\v{s}}, ``On the educational impact of chatgpt: Is artificial intelligence ready to obtain a university degree?'' \emph{arXiv preprint arXiv:2303.11146}, 2023.

\bibitem{li2022cctest}
Z.~Li, C.~Wang, Z.~Liu, H.~Wang, S.~Wang, and C.~Gao, ``Cctest: Testing and repairing code completion systems,'' \emph{ICSE}, 2023.

\bibitem{liu2023chatgpt}
J.~Liu, C.~Liu, R.~Lv, K.~Zhou, and Y.~Zhang, ``Is chatgpt a good recommender? a preliminary study,'' \emph{arXiv preprint arXiv:2304.10149}, 2023.

\bibitem{zhao2023survey}
W.~X. Zhao, K.~Zhou, J.~Li, T.~Tang, X.~Wang, Y.~Hou, Y.~Min, B.~Zhang, J.~Zhang, Z.~Dong \emph{et~al.}, ``A survey of large language models,'' \emph{arXiv preprint arXiv:2303.18223}, 2023.

\bibitem{qiu2020pre}
X.~Qiu, T.~Sun, Y.~Xu, Y.~Shao, N.~Dai, and X.~Huang, ``Pre-trained models for natural language processing: A survey,'' \emph{Science China Technological Sciences}, vol.~63, no.~10, pp. 1872--1897, 2020.

\bibitem{yang2023harnessing}
J.~Yang, H.~Jin, R.~Tang, X.~Han, Q.~Feng, H.~Jiang, B.~Yin, and X.~Hu, ``Harnessing the power of llms in practice: A survey on chatgpt and beyond,'' \emph{arXiv preprint arXiv:2304.13712}, 2023.

\bibitem{petroni2019language}
F.~Petroni, T.~Rockt{\"a}schel, S.~Riedel, P.~Lewis, A.~Bakhtin, Y.~Wu, and A.~Miller, ``Language models as knowledge bases?'' in \emph{EMNLP-IJCNLP}, 2019, pp. 2463--2473.

\bibitem{ji2023survey}
Z.~Ji, N.~Lee, R.~Frieske, T.~Yu, D.~Su, Y.~Xu, E.~Ishii, Y.~J. Bang, A.~Madotto, and P.~Fung, ``Survey of hallucination in natural language generation,'' \emph{ACM Computing Surveys}, vol.~55, no.~12, pp. 1--38, 2023.

\bibitem{zhang2022survey}
H.~Zhang, H.~Song, S.~Li, M.~Zhou, and D.~Song, ``A survey of controllable text generation using transformer-based pre-trained language models,'' \emph{arXiv preprint arXiv:2201.05337}, 2022.

\bibitem{danilevsky2020survey}
M.~Danilevsky, K.~Qian, R.~Aharonov, Y.~Katsis, B.~Kawas, and P.~Sen, ``A survey of the state of explainable ai for natural language processing,'' \emph{arXiv preprint arXiv:2010.00711}, 2020.

\bibitem{wang2023robustness}
J.~Wang, X.~Hu, W.~Hou, H.~Chen, R.~Zheng, Y.~Wang, L.~Yang, H.~Huang, W.~Ye, X.~Geng \emph{et~al.}, ``On the robustness of chatgpt: An adversarial and out-of-distribution perspective,'' \emph{arXiv preprint arXiv:2302.12095}, 2023.

\bibitem{ji2021survey}
S.~Ji, S.~Pan, E.~Cambria, P.~Marttinen, and S.~Y. Philip, ``A survey on knowledge graphs: Representation, acquisition, and applications,'' \emph{IEEE TNNLS}, vol.~33, no.~2, pp. 494--514, 2021.

\bibitem{vrandevcic2014wikidata}
D.~Vrande{\v{c}}i{\'c} and M.~Kr{\"o}tzsch, ``Wikidata: a free collaborative knowledgebase,'' \emph{Communications of the ACM}, vol.~57, no.~10, pp. 78--85, 2014.

\bibitem{hu2018state}
S.~Hu, L.~Zou, and X.~Zhang, ``A state-transition framework to answer complex questions over knowledge base,'' in \emph{EMNLP}, 2018, pp. 2098--2108.

\bibitem{zhang2021neural}
J.~Zhang, B.~Chen, L.~Zhang, X.~Ke, and H.~Ding, ``Neural, symbolic and neural-symbolic reasoning on knowledge graphs,'' \emph{AI Open}, vol.~2, pp. 14--35, 2021.

\bibitem{abu2021domain}
B.~Abu-Salih, ``Domain-specific knowledge graphs: A survey,'' \emph{Journal of Network and Computer Applications}, vol. 185, p. 103076, 2021.

\bibitem{Mitchell2015NeverEndingL}
T.~Mitchell, W.~Cohen, E.~Hruschka, P.~Talukdar, B.~Yang, J.~Betteridge, A.~Carlson, B.~Dalvi, M.~Gardner, B.~Kisiel, K.~Jayant, L.~Ni, M.~Kathryn, M.~Thahir, N.~Ndapandula, P.~Emmanouil, R.~Alan, S.~Mehdi, S.~Burr, W.~Derry, G.~Abhinav, C.~Xi, S.~Abulhair, and W.~Joel, ``Never-ending learning,'' \emph{Communications of the ACM}, vol.~61, no.~5, pp. 103--115, 2018.

\bibitem{zhong2023comprehensive}
L.~Zhong, J.~Wu, Q.~Li, H.~Peng, and X.~Wu, ``A comprehensive survey on automatic knowledge graph construction,'' \emph{arXiv preprint arXiv:2302.05019}, 2023.

\bibitem{yao2019kg}
L.~Yao, C.~Mao, and Y.~Luo, ``Kg-bert: Bert for knowledge graph completion,'' \emph{arXiv preprint arXiv:1909.03193}, 2019.

\bibitem{luo2023normalizing}
L.~Luo, Y.-F. Li, G.~Haffari, and S.~Pan, ``Normalizing flow-based neural process for few-shot knowledge graph completion,'' \emph{SIGIR}, 2023.

\bibitem{bang2023multitask}
Y.~Bang, S.~Cahyawijaya, N.~Lee, W.~Dai, D.~Su, B.~Wilie, H.~Lovenia, Z.~Ji, T.~Yu, W.~Chung \emph{et~al.}, ``A multitask, multilingual, multimodal evaluation of chatgpt on reasoning, hallucination, and interactivity,'' \emph{arXiv preprint arXiv:2302.04023}, 2023.

\bibitem{wang2022self}
X.~Wang, J.~Wei, D.~Schuurmans, Q.~Le, E.~Chi, and D.~Zhou, ``Self-consistency improves chain of thought reasoning in language models,'' \emph{arXiv preprint arXiv:2203.11171}, 2022.

\bibitem{golovneva2022roscoe}
O.~Golovneva, M.~Chen, S.~Poff, M.~Corredor, L.~Zettlemoyer, M.~Fazel-Zarandi, and A.~Celikyilmaz, ``Roscoe: A suite of metrics for scoring step-by-step reasoning,'' \emph{ICLR}, 2023.

\bibitem{suchanek2007yago}
F.~M. Suchanek, G.~Kasneci, and G.~Weikum, ``Yago: a core of semantic knowledge,'' in \emph{WWW}, 2007, pp. 697--706.

\bibitem{carlson2010toward}
A.~Carlson, J.~Betteridge, B.~Kisiel, B.~Settles, E.~Hruschka, and T.~Mitchell, ``Toward an architecture for never-ending language learning,'' in \emph{Proceedings of the AAAI conference on artificial intelligence}, vol.~24, no.~1, 2010, pp. 1306--1313.

\bibitem{bordes2013translating}
A.~Bordes, N.~Usunier, A.~Garcia-Duran, J.~Weston, and O.~Yakhnenko, ``Translating embeddings for modeling multi-relational data,'' \emph{NeurIPS}, vol.~26, 2013.

\bibitem{wan2021reasoning}
G.~Wan, S.~Pan, C.~Gong, C.~Zhou, and G.~Haffari, ``Reasoning like human: Hierarchical reinforcement learning for knowledge graph reasoning,'' in \emph{AAAI}, 2021, pp. 1926--1932.

\bibitem{zhang-etal-2019-ernie}
Z.~Zhang, X.~Han, Z.~Liu, X.~Jiang, M.~Sun, and Q.~Liu, ``{ERNIE}: Enhanced language representation with informative entities,'' in \emph{ACL}, 2019, pp. 1441--1451.

\bibitem{DBLP:conf/aaai/LiuZ0WJD020}
W.~Liu, P.~Zhou, Z.~Zhao, Z.~Wang, Q.~Ju, H.~Deng, and P.~Wang, ``{K-BERT:} enabling language representation with knowledge graph,'' in \emph{AAAI}, 2020, pp. 2901--2908.

\bibitem{DBLP:conf/aaai/LiuW0PY21}
Y.~Liu, Y.~Wan, L.~He, H.~Peng, and P.~S. Yu, ``{KG-BART:} knowledge graph-augmented {BART} for generative commonsense reasoning,'' in \emph{AAAI}, 2021, pp. 6418--6425.

\bibitem{lin-etal-2019-kagnet}
B.~Y. Lin, X.~Chen, J.~Chen, and X.~Ren, ``{K}ag{N}et: Knowledge-aware graph networks for commonsense reasoning,'' in \emph{EMNLP-IJCNLP}, 2019, pp. 2829--2839.

\bibitem{dai2021knowledge}
D.~Dai, L.~Dong, Y.~Hao, Z.~Sui, B.~Chang, and F.~Wei, ``Knowledge neurons in pretrained transformers,'' \emph{arXiv preprint arXiv:2104.08696}, 2021.

\bibitem{wang-etal-2021-kepler}
X.~Wang, T.~Gao, Z.~Zhu, Z.~Zhang, Z.~Liu, J.~Li, and J.~Tang, ``{KEPLER}: A unified model for knowledge embedding and pre-trained language representation,'' \emph{Transactions of the Association for Computational Linguistics}, vol.~9, pp. 176--194, 2021.

\bibitem{melnyk2021grapher}
I.~Melnyk, P.~Dognin, and P.~Das, ``Grapher: Multi-stage knowledge graph construction using pretrained language models,'' in \emph{NeurIPS 2021 Workshop on Deep Generative Models and Downstream Applications}, 2021.

\bibitem{ke-etal-2021-jointgt}
P.~Ke, H.~Ji, Y.~Ran, X.~Cui, L.~Wang, L.~Song, X.~Zhu, and M.~Huang, ``{J}oint{GT}: Graph-text joint representation learning for text generation from knowledge graphs,'' in \emph{ACL Finding}, 2021, pp. 2526--2538.

\bibitem{jiang2023unikgqa}
J.~Jiang, K.~Zhou, W.~X. Zhao, and J.-R. Wen, ``Unikgqa: Unified retrieval and reasoning for solving multi-hop question answering over knowledge graph,'' \emph{ICLR 2023}, 2023.

\bibitem{yasunaga2022deep}
M.~Yasunaga, A.~Bosselut, H.~Ren, X.~Zhang, C.~D. Manning, P.~S. Liang, and J.~Leskovec, ``Deep bidirectional language-knowledge graph pretraining,'' \emph{NeurIPS}, vol.~35, pp. 37\,309--37\,323, 2022.

\bibitem{choudhary2023complex}
N.~Choudhary and C.~K. Reddy, ``Complex logical reasoning over knowledge graphs using large language models,'' \emph{arXiv preprint arXiv:2305.01157}, 2023.

\bibitem{wang2023unifying}
S.~Wang, Z.~Wei, J.~Xu, and Z.~Fan, ``Unifying structure reasoning and language model pre-training for complex reasoning,'' \emph{arXiv preprint arXiv:2301.08913}, 2023.

\bibitem{zhen2022survey}
C.~Zhen, Y.~Shang, X.~Liu, Y.~Li, Y.~Chen, and D.~Zhang, ``A survey on knowledge-enhanced pre-trained language models,'' \emph{arXiv preprint arXiv:2212.13428}, 2022.

\bibitem{wei2021knowledge}
X.~Wei, S.~Wang, D.~Zhang, P.~Bhatia, and A.~Arnold, ``Knowledge enhanced pretrained language models: A compreshensive survey,'' \emph{arXiv preprint arXiv:2110.08455}, 2021.

\bibitem{yin2022survey}
D.~Yin, L.~Dong, H.~Cheng, X.~Liu, K.-W. Chang, F.~Wei, and J.~Gao, ``A survey of knowledge-intensive nlp with pre-trained language models,'' \emph{arXiv preprint arXiv:2202.08772}, 2022.

\bibitem{vaswani2017attention}
A.~Vaswani, N.~Shazeer, N.~Parmar, J.~Uszkoreit, L.~Jones, A.~N. Gomez, {\L}.~Kaiser, and I.~Polosukhin, ``Attention is all you need,'' \emph{NeurIPS}, vol.~30, 2017.

\bibitem{lanalbert}
Z.~Lan, M.~Chen, S.~Goodman, K.~Gimpel, P.~Sharma, and R.~Soricut, ``Albert: A lite bert for self-supervised learning of language representations,'' in \emph{ICLR}, 2019.

\bibitem{clark2020electra}
K.~Clark, M.-T. Luong, Q.~V. Le, and C.~D. Manning, ``Electra: Pre-training text encoders as discriminators rather than generators,'' \emph{arXiv preprint arXiv:2003.10555}, 2020.

\bibitem{hakala2019biomedical}
K.~Hakala and S.~Pyysalo, ``Biomedical named entity recognition with multilingual bert,'' in \emph{Proceedings of the 5th workshop on BioNLP open shared tasks}, 2019, pp. 56--61.

\bibitem{tay2022ul2}
Y.~Tay, M.~Dehghani, V.~Q. Tran, X.~Garcia, J.~Wei, X.~Wang, H.~W. Chung, D.~Bahri, T.~Schuster, S.~Zheng \emph{et~al.}, ``Ul2: Unifying language learning paradigms,'' in \emph{ICLR}, 2022.

\bibitem{sanhmultitask}
V.~Sanh, A.~Webson, C.~Raffel, S.~Bach, L.~Sutawika, Z.~Alyafeai, A.~Chaffin, A.~Stiegler, A.~Raja, M.~Dey \emph{et~al.}, ``Multitask prompted training enables zero-shot task generalization,'' in \emph{ICLR}, 2022.

\bibitem{zoph2022st}
B.~Zoph, I.~Bello, S.~Kumar, N.~Du, Y.~Huang, J.~Dean, N.~Shazeer, and W.~Fedus, ``St-moe: Designing stable and transferable sparse expert models,'' \emph{URL https://arxiv. org/abs/2202.08906}, 2022.

\bibitem{zeng2023glm-130b}
A.~Zeng, X.~Liu, Z.~Du, Z.~Wang, H.~Lai, M.~Ding, Z.~Yang, Y.~Xu, W.~Zheng, X.~Xia, W.~L. Tam, Z.~Ma, Y.~Xue, J.~Zhai, W.~Chen, Z.~Liu, P.~Zhang, Y.~Dong, and J.~Tang, ``{GLM}-130b: An open bilingual pre-trained model,'' in \emph{ICLR}, 2023.

\bibitem{xue2021mt5}
L.~Xue, N.~Constant, A.~Roberts, M.~Kale, R.~Al-Rfou, A.~Siddhant, A.~Barua, and C.~Raffel, ``mt5: A massively multilingual pre-trained text-to-text transformer,'' in \emph{NAACL}, 2021, pp. 483--498.

\bibitem{brown2020language}
T.~Brown, B.~Mann, N.~Ryder, M.~Subbiah, J.~D. Kaplan, P.~Dhariwal, A.~Neelakantan, P.~Shyam, G.~Sastry, A.~Askell \emph{et~al.}, ``Language models are few-shot learners,'' \emph{Advances in neural information processing systems}, vol.~33, pp. 1877--1901, 2020.

\bibitem{ouyang2022training}
L.~Ouyang, J.~Wu, X.~Jiang, D.~Almeida, C.~Wainwright, P.~Mishkin, C.~Zhang, S.~Agarwal, K.~Slama, A.~Ray \emph{et~al.}, ``Training language models to follow instructions with human feedback,'' \emph{NeurIPS}, vol.~35, pp. 27\,730--27\,744, 2022.

\bibitem{touvron2023llama}
H.~Touvron, T.~Lavril, G.~Izacard, X.~Martinet, M.-A. Lachaux, T.~Lacroix, B.~Rozi{\`e}re, N.~Goyal, E.~Hambro, F.~Azhar \emph{et~al.}, ``Llama: Open and efficient foundation language models,'' \emph{arXiv preprint arXiv:2302.13971}, 2023.

\bibitem{Saravia_Prompt_Engineering_Guide_2022}
E.~Saravia, ``{Prompt Engineering Guide},'' \url{https://github.com/dair-ai/Prompt-Engineering-Guide}, 2022, accessed: 2022-12.

\bibitem{weichain}
J.~Wei, X.~Wang, D.~Schuurmans, M.~Bosma, F.~Xia, E.~H. Chi, Q.~V. Le, D.~Zhou \emph{et~al.}, ``Chain-of-thought prompting elicits reasoning in large language models,'' in \emph{NeurIPS}.

\bibitem{li2023graph}
S.~Li, Y.~Gao, H.~Jiang, Q.~Yin, Z.~Li, X.~Yan, C.~Zhang, and B.~Yin, ``Graph reasoning for question answering with triplet retrieval,'' in \emph{ACL}, 2023.

\bibitem{wen2023mindmap}
Y.~Wen, Z.~Wang, and J.~Sun, ``Mindmap: Knowledge graph prompting sparks graph of thoughts in large language models,'' \emph{arXiv preprint arXiv:2308.09729}, 2023.

\bibitem{bollacker2008freebase}
K.~Bollacker, C.~Evans, P.~Paritosh, T.~Sturge, and J.~Taylor, ``Freebase: A collaboratively created graph database for structuring human knowledge,'' in \emph{SIGMOD}, 2008, pp. 1247--1250.

\bibitem{auer2007dbpedia}
S.~Auer, C.~Bizer, G.~Kobilarov, J.~Lehmann, R.~Cyganiak, and Z.~Ives, ``Dbpedia: A nucleus for a web of open data,'' in \emph{The Semantic Web: 6th International Semantic Web Conference}.\hskip 1em plus 0.5em minus 0.4em\relax Springer, 2007, pp. 722--735.

\bibitem{xu2017cn}
B.~Xu, Y.~Xu, J.~Liang, C.~Xie, B.~Liang, W.~Cui, and Y.~Xiao, ``Cn-dbpedia: A never-ending chinese knowledge extraction system,'' in \emph{30th International Conference on Industrial Engineering and Other Applications of Applied Intelligent Systems}.\hskip 1em plus 0.5em minus 0.4em\relax Springer, 2017, pp. 428--438.

\bibitem{hai2022vikidia}
P.~Hai-Nyzhnyk, ``Vikidia as a universal multilingual online encyclopedia for children,'' \emph{The Encyclopedia Herald of Ukraine}, vol.~14, 2022.

\bibitem{ilievski2021cskg}
F.~Ilievski, P.~Szekely, and B.~Zhang, ``Cskg: The commonsense knowledge graph,'' \emph{Extended Semantic Web Conference (ESWC)}, 2021.

\bibitem{speer2017conceptnet}
R.~Speer, J.~Chin, and C.~Havasi, ``Conceptnet 5.5: An open multilingual graph of general knowledge,'' in \emph{Proceedings of the AAAI conference on artificial intelligence}, vol.~31, no.~1, 2017.

\bibitem{ji2020language}
H.~Ji, P.~Ke, S.~Huang, F.~Wei, X.~Zhu, and M.~Huang, ``Language generation with multi-hop reasoning on commonsense knowledge graph,'' in \emph{EMNLP}, 2020, pp. 725--736.

\bibitem{hwang2021comet}
J.~D. Hwang, C.~Bhagavatula, R.~Le~Bras, J.~Da, K.~Sakaguchi, A.~Bosselut, and Y.~Choi, ``(comet-) atomic 2020: On symbolic and neural commonsense knowledge graphs,'' in \emph{AAAI}, vol.~35, no.~7, 2021, pp. 6384--6392.

\bibitem{zhang2020aser}
H.~Zhang, X.~Liu, H.~Pan, Y.~Song, and C.~W.-K. Leung, ``Aser: A large-scale eventuality knowledge graph,'' in \emph{Proceedings of the web conference 2020}, 2020, pp. 201--211.

\bibitem{zhang2021transomcs}
H.~Zhang, D.~Khashabi, Y.~Song, and D.~Roth, ``Transomcs: from linguistic graphs to commonsense knowledge,'' in \emph{IJCAI}, 2021, pp. 4004--4010.

\bibitem{ijcai2020-guided}
Z.~Li, X.~Ding, T.~Liu, J.~E. Hu, and B.~Van~Durme, ``Guided generation of cause and effect,'' in \emph{IJCAI}, 2020.

\bibitem{bodenreider2004unified}
O.~Bodenreider, ``The unified medical language system (umls): integrating biomedical terminology,'' \emph{Nucleic acids research}, vol.~32, no. suppl\_1, pp. D267--D270, 2004.

\bibitem{liu2019anticipating}
Y.~Liu, Q.~Zeng, J.~Ordieres~Mer{\'e}, and H.~Yang, ``Anticipating stock market of the renowned companies: a knowledge graph approach,'' \emph{Complexity}, vol. 2019, 2019.

\bibitem{zhu2017intelligent}
Y.~Zhu, W.~Zhou, Y.~Xu, J.~Liu, Y.~Tan \emph{et~al.}, ``Intelligent learning for knowledge graph towards geological data,'' \emph{Scientific Programming}, vol. 2017, 2017.

\bibitem{choi2019inference}
W.~Choi and H.~Lee, ``Inference of biomedical relations among chemicals, genes, diseases, and symptoms using knowledge representation learning,'' \emph{IEEE Access}, vol.~7, pp. 179\,373--179\,384, 2019.

\bibitem{farazi2020knowledge}
F.~Farazi, M.~Salamanca, S.~Mosbach, J.~Akroyd, A.~Eibeck, L.~K. Aditya, A.~Chadzynski, K.~Pan, X.~Zhou, S.~Zhang \emph{et~al.}, ``Knowledge graph approach to combustion chemistry and interoperability,'' \emph{ACS omega}, vol.~5, no.~29, pp. 18\,342--18\,348, 2020.

\bibitem{wu-genealogy-23}
X.~Wu, T.~Jiang, Y.~Zhu, and C.~Bu, ``Knowledge graph for china's genealogy,'' \emph{IEEE TKDE}, vol.~35, no.~1, pp. 634--646, 2023.

\bibitem{zhu2022multi}
X.~Zhu, Z.~Li, X.~Wang, X.~Jiang, P.~Sun, X.~Wang, Y.~Xiao, and N.~J. Yuan, ``Multi-modal knowledge graph construction and application: A survey,'' \emph{IEEE TKDE}, 2022.

\bibitem{ferrada2017imgpedia}
S.~Ferrada, B.~Bustos, and A.~Hogan, ``Imgpedia: a linked dataset with content-based analysis of wikimedia images,'' in \emph{The Semantic Web--ISWC 2017}.\hskip 1em plus 0.5em minus 0.4em\relax Springer, 2017, pp. 84--93.

\bibitem{liu2019mmkg}
Y.~Liu, H.~Li, A.~Garcia-Duran, M.~Niepert, D.~Onoro-Rubio, and D.~S. Rosenblum, ``Mmkg: multi-modal knowledge graphs,'' in \emph{The Semantic Web: 16th International Conference, ESWC 2019, Portoro{\v{z}}, Slovenia, June 2--6, 2019, Proceedings 16}.\hskip 1em plus 0.5em minus 0.4em\relax Springer, 2019, pp. 459--474.

\bibitem{wang2020richpedia}
M.~Wang, H.~Wang, G.~Qi, and Q.~Zheng, ``Richpedia: a large-scale, comprehensive multi-modal knowledge graph,'' \emph{Big Data Research}, vol.~22, p. 100159, 2020.

\bibitem{shi2019knowledge}
B.~Shi, L.~Ji, P.~Lu, Z.~Niu, and N.~Duan, ``Knowledge aware semantic concept expansion for image-text matching.'' in \emph{IJCAI}, vol.~1, 2019, p.~2.

\bibitem{shah2019kvqa}
S.~Shah, A.~Mishra, N.~Yadati, and P.~P. Talukdar, ``Kvqa: Knowledge-aware visual question answering,'' in \emph{AAAI}, vol.~33, no.~01, 2019, pp. 8876--8884.

\bibitem{sun2020multi}
R.~Sun, X.~Cao, Y.~Zhao, J.~Wan, K.~Zhou, F.~Zhang, Z.~Wang, and K.~Zheng, ``Multi-modal knowledge graphs for recommender systems,'' in \emph{CIKM}, 2020, pp. 1405--1414.

\bibitem{ICDE2023_OpenBG}
S.~Deng, C.~Wang, Z.~Li, N.~Zhang, Z.~Dai, H.~Chen, F.~Xiong, M.~Yan, Q.~Chen, M.~Chen, J.~Chen, J.~Z. Pan, B.~Hooi, and H.~Chen, ``Construction and applications of billion-scale pre-trained multimodal business knowledge graph,'' in \emph{{ICDE}}, 2023.

\bibitem{rosset2020knowledge}
C.~Rosset, C.~Xiong, M.~Phan, X.~Song, P.~Bennett, and S.~Tiwary, ``Knowledge-aware language model pretraining,'' \emph{arXiv preprint arXiv:2007.00655}, 2020.

\bibitem{NEURIPS2020_6b493230}
P.~Lewis, E.~Perez, A.~Piktus, F.~Petroni, V.~Karpukhin, N.~Goyal, H.~K\"{u}ttler, M.~Lewis, W.-t. Yih, T.~Rockt\"{a}schel, S.~Riedel, and D.~Kiela, ``Retrieval-augmented generation for knowledge-intensive nlp tasks,'' in \emph{NeurIPS}, vol.~33, 2020, pp. 9459--9474.

\bibitem{zhu2023llms}
Y.~Zhu, X.~Wang, J.~Chen, S.~Qiao, Y.~Ou, Y.~Yao, S.~Deng, H.~Chen, and N.~Zhang, ``Llms for knowledge graph construction and reasoning: Recent capabilities and future opportunities,'' \emph{arXiv preprint arXiv:2305.13168}, 2023.

\bibitem{zhang2020pretrain}
Z.~Zhang, X.~Liu, Y.~Zhang, Q.~Su, X.~Sun, and B.~He, ``Pretrain-kge: learning knowledge representation from pretrained language models,'' in \emph{EMNLP Finding}, 2020, pp. 259--266.

\bibitem{kumar2020building}
A.~Kumar, A.~Pandey, R.~Gadia, and M.~Mishra, ``Building knowledge graph using pre-trained language model for learning entity-aware relationships,'' in \emph{2020 IEEE International Conference on Computing, Power and Communication Technologies (GUCON)}.\hskip 1em plus 0.5em minus 0.4em\relax IEEE, 2020, pp. 310--315.

\bibitem{GenKGC}
X.~Xie, N.~Zhang, Z.~Li, S.~Deng, H.~Chen, F.~Xiong, M.~Chen, and H.~Chen, ``From discrimination to generation: Knowledge graph completion with generative transformer,'' in \emph{WWW}, 2022, pp. 162--165.

\bibitem{chen2023incorporating}
Z.~Chen, C.~Xu, F.~Su, Z.~Huang, and Y.~Dou, ``Incorporating structured sentences with time-enhanced bert for fully-inductive temporal relation prediction,'' \emph{SIGIR}, 2023.

\bibitem{zhu2023minigpt}
D.~Zhu, J.~Chen, X.~Shen, X.~Li, and M.~Elhoseiny, ``Minigpt-4: Enhancing vision-language understanding with advanced large language models,'' \emph{arXiv preprint arXiv:2304.10592}, 2023.

\bibitem{challenge-mm-3}
M.~Warren, D.~A. Shamma, and P.~J. Hayes, ``Knowledge engineering with image data in real-world settings,'' in \emph{AAAI}, ser. {CEUR} Workshop Proceedings, vol. 2846, 2021.

\bibitem{thoppilan2022lamda}
R.~Thoppilan, D.~De~Freitas, J.~Hall, N.~Shazeer, A.~Kulshreshtha, H.-T. Cheng, A.~Jin, T.~Bos, L.~Baker, Y.~Du \emph{et~al.}, ``Lamda: Language models for dialog applications,'' \emph{arXiv preprint arXiv:2201.08239}, 2022.

\bibitem{sun2021ernie}
Y.~Sun, S.~Wang, S.~Feng, S.~Ding, C.~Pang, J.~Shang, J.~Liu, X.~Chen, Y.~Zhao, Y.~Lu \emph{et~al.}, ``Ernie 3.0: Large-scale knowledge enhanced pre-training for language understanding and generation,'' \emph{arXiv preprint arXiv:2107.02137}, 2021.

\bibitem{shen-etal-2020-exploiting}
T.~Shen, Y.~Mao, P.~He, G.~Long, A.~Trischler, and W.~Chen, ``Exploiting structured knowledge in text via graph-guided representation learning,'' in \emph{EMNLP}, 2020, pp. 8980--8994.

\bibitem{zhang2020bert}
D.~Zhang, Z.~Yuan, Y.~Liu, F.~Zhuang, H.~Chen, and H.~Xiong, ``E-bert: A phrase and product knowledge enhanced language model for e-commerce,'' \emph{arXiv preprint arXiv:2009.02835}, 2020.

\bibitem{li-etal-2022-pre-training}
S.~Li, X.~Li, L.~Shang, C.~Sun, B.~Liu, Z.~Ji, X.~Jiang, and Q.~Liu, ``Pre-training language models with deterministic factual knowledge,'' in \emph{EMNLP}, 2022, pp. 11\,118--11\,131.

\bibitem{kang2022kala}
M.~Kang, J.~Baek, and S.~J. Hwang, ``Kala: Knowledge-augmented language model adaptation,'' in \emph{NAACL}, 2022, pp. 5144--5167.

\bibitem{Xiong2020Pretrained}
W.~Xiong, J.~Du, W.~Y. Wang, and V.~Stoyanov, ``Pretrained encyclopedia: Weakly supervised knowledge-pretrained language model,'' in \emph{ICLR}, 2020.

\bibitem{sun-etal-2020-colake}
T.~Sun, Y.~Shao, X.~Qiu, Q.~Guo, Y.~Hu, X.~Huang, and Z.~Zhang, ``{C}o{LAKE}: Contextualized language and knowledge embedding,'' in \emph{Proceedings of the 28th International Conference on Computational Linguistics}, 2020, pp. 3660--3670.

\bibitem{DBLP:conf/aaai/Zhang0HQTH022}
T.~Zhang, C.~Wang, N.~Hu, M.~Qiu, C.~Tang, X.~He, and J.~Huang, ``{DKPLM:} decomposable knowledge-enhanced pre-trained language model for natural language understanding,'' in \emph{AAAI}, 2022, pp. 11\,703--11\,711.

\bibitem{wang2022knowledge}
J.~Wang, W.~Huang, M.~Qiu, Q.~Shi, H.~Wang, X.~Li, and M.~Gao, ``Knowledge prompting in pre-trained language model for natural language understanding,'' in \emph{Proceedings of the 2022 Conference on Empirical Methods in Natural Language Processing}, 2022, pp. 3164--3177.

\bibitem{ye2022ontology}
H.~Ye, N.~Zhang, S.~Deng, X.~Chen, H.~Chen, F.~Xiong, X.~Chen, and H.~Chen, ``Ontology-enhanced prompt-tuning for few-shot learning,'' in \emph{Proceedings of the ACM Web Conference 2022}, 2022, pp. 778--787.

\bibitem{luo2023chatkbqa}
H.~Luo, Z.~Tang, S.~Peng, Y.~Guo, W.~Zhang, C.~Ma, G.~Dong, M.~Song, W.~Lin \emph{et~al.}, ``Chatkbqa: A generate-then-retrieve framework for knowledge base question answering with fine-tuned large language models,'' \emph{arXiv preprint arXiv:2310.08975}, 2023.

\bibitem{luo_rog}
L.~Luo, Y.-F. Li, G.~Haffari, and S.~Pan, ``Reasoning on graphs: Faithful and interpretable large language model reasoning,'' \emph{arXiv preprint arxiv:2310.01061}, 2023.

\bibitem{logan-etal-2019-baracks}
R.~Logan, N.~F. Liu, M.~E. Peters, M.~Gardner, and S.~Singh, ``{B}arack{'}s wife hillary: Using knowledge graphs for fact-aware language modeling,'' in \emph{ACL}, 2019, pp. 5962--5971.

\bibitem{10.5555/3524938.3525306}
K.~Guu, K.~Lee, Z.~Tung, P.~Pasupat, and M.-W. Chang, ``Realm: Retrieval-augmented language model pre-training,'' in \emph{ICML}, 2020.

\bibitem{wu-etal-2022-efficient}
Y.~Wu, Y.~Zhao, B.~Hu, P.~Minervini, P.~Stenetorp, and S.~Riedel, ``An efficient memory-augmented transformer for knowledge-intensive {NLP} tasks,'' in \emph{EMNLP}, 2022, pp. 5184--5196.

\bibitem{luo2023chatrule}
L.~Luo, J.~Ju, B.~Xiong, Y.-F. Li, G.~Haffari, and S.~Pan, ``Chatrule: Mining logical rules with large language models for knowledge graph reasoning,'' \emph{arXiv preprint arXiv:2309.01538}, 2023.

\bibitem{wang2023boosting}
J.~Wang, Q.~Sun, N.~Chen, X.~Li, and M.~Gao, ``Boosting language models reasoning with chain-of-knowledge prompting,'' \emph{arXiv preprint arXiv:2306.06427}, 2023.

\bibitem{jiang2020can}
Z.~Jiang, F.~F. Xu, J.~Araki, and G.~Neubig, ``How can we know what language models know?'' \emph{Transactions of the Association for Computational Linguistics}, vol.~8, pp. 423--438, 2020.

\bibitem{shin2020autoprompt}
T.~Shin, Y.~Razeghi, R.~L. Logan~IV, E.~Wallace, and S.~Singh, ``Autoprompt: Eliciting knowledge from language models with automatically generated prompts,'' \emph{arXiv preprint arXiv:2010.15980}, 2020.

\bibitem{meng2021rewire}
Z.~Meng, F.~Liu, E.~Shareghi, Y.~Su, C.~Collins, and N.~Collier, ``Rewire-then-probe: A contrastive recipe for probing biomedical knowledge of pre-trained language models,'' \emph{arXiv preprint arXiv:2110.08173}, 2021.

\bibitem{luo2023systematic}
L.~Luo, T.-T. Vu, D.~Phung, and G.~Haffari, ``Systematic assessment of factual knowledge in large language models,'' in \emph{EMNLP}, 2023.

\bibitem{swamy2021interpreting}
V.~Swamy, A.~Romanou, and M.~Jaggi, ``Interpreting language models through knowledge graph extraction,'' \emph{arXiv preprint arXiv:2111.08546}, 2021.

\bibitem{li2022pre}
S.~Li, X.~Li, L.~Shang, Z.~Dong, C.~Sun, B.~Liu, Z.~Ji, X.~Jiang, and Q.~Liu, ``How pre-trained language models capture factual knowledge? a causal-inspired analysis,'' \emph{arXiv preprint arXiv:2203.16747}, 2022.

\bibitem{tian-etal-2020-skep}
H.~Tian, C.~Gao, X.~Xiao, H.~Liu, B.~He, H.~Wu, H.~Wang, and F.~Wu, ``{SKEP}: Sentiment knowledge enhanced pre-training for sentiment analysis,'' in \emph{ACL}, 2020, pp. 4067--4076.

\bibitem{yu-etal-2022-dict}
W.~Yu, C.~Zhu, Y.~Fang, D.~Yu, S.~Wang, Y.~Xu, M.~Zeng, and M.~Jiang, ``Dict-{BERT}: Enhancing language model pre-training with dictionary,'' in \emph{ACL}, 2022, pp. 1907--1918.

\bibitem{mccoy-etal-2019-right}
T.~McCoy, E.~Pavlick, and T.~Linzen, ``Right for the wrong reasons: Diagnosing syntactic heuristics in natural language inference,'' in \emph{ACL}, 2019, pp. 3428--3448.

\bibitem{wilmot-keller-2021-memory}
D.~Wilmot and F.~Keller, ``Memory and knowledge augmented language models for inferring salience in long-form stories,'' in \emph{EMNLP}, 2021, pp. 851--865.

\bibitem{adolphs2021query}
L.~Adolphs, S.~Dhuliawala, and T.~Hofmann, ``How to query language models?'' \emph{arXiv preprint arXiv:2108.01928}, 2021.

\bibitem{sung2021can}
M.~Sung, J.~Lee, S.~Yi, M.~Jeon, S.~Kim, and J.~Kang, ``Can language models be biomedical knowledge bases?'' in \emph{EMNLP}, 2021, pp. 4723--4734.

\bibitem{mallen2022not}
A.~Mallen, A.~Asai, V.~Zhong, R.~Das, H.~Hajishirzi, and D.~Khashabi, ``When not to trust language models: Investigating effectiveness and limitations of parametric and non-parametric memories,'' \emph{arXiv preprint arXiv:2212.10511}, 2022.

\bibitem{yasunaga-etal-2021-qa}
M.~Yasunaga, H.~Ren, A.~Bosselut, P.~Liang, and J.~Leskovec, ``{QA}-{GNN}: Reasoning with language models and knowledge graphs for question answering,'' in \emph{NAACL}, 2021, pp. 535--546.

\bibitem{nayyeri2022integrating}
M.~Nayyeri, Z.~Wang, M.~Akter, M.~M. Alam, M.~R. A.~H. Rony, J.~Lehmann, S.~Staab \emph{et~al.}, ``Integrating knowledge graph embedding and pretrained language models in hypercomplex spaces,'' \emph{arXiv preprint arXiv:2208.02743}, 2022.

\bibitem{huang2022endowing}
N.~Huang, Y.~R. Deshpande, Y.~Liu, H.~Alberts, K.~Cho, C.~Vania, and I.~Calixto, ``Endowing language models with multimodal knowledge graph representations,'' \emph{arXiv preprint arXiv:2206.13163}, 2022.

\bibitem{alam2022language}
M.~M. Alam, M.~R. A.~H. Rony, M.~Nayyeri, K.~Mohiuddin, M.~M. Akter, S.~Vahdati, and J.~Lehmann, ``Language model guided knowledge graph embeddings,'' \emph{IEEE Access}, vol.~10, pp. 76\,008--76\,020, 2022.

\bibitem{wang2022language}
X.~Wang, Q.~He, J.~Liang, and Y.~Xiao, ``Language models as knowledge embeddings,'' \emph{arXiv preprint arXiv:2206.12617}, 2022.

\bibitem{zhang2022reasoning}
N.~Zhang, X.~Xie, X.~Chen, S.~Deng, C.~Tan, F.~Huang, X.~Cheng, and H.~Chen, ``Reasoning through memorization: Nearest neighbor knowledge graph embeddings,'' \emph{arXiv preprint arXiv:2201.05575}, 2022.

\bibitem{lambdakg}
X.~Xie, Z.~Li, X.~Wang, Y.~Zhu, N.~Zhang, J.~Zhang, S.~Cheng, B.~Tian, S.~Deng, F.~Xiong, and H.~Chen, ``Lambdakg: A library for pre-trained language model-based knowledge graph embeddings,'' 2022.

\bibitem{MTL-KGC}
B.~Kim, T.~Hong, Y.~Ko, and J.~Seo, ``Multi-task learning for knowledge graph completion with pre-trained language models,'' in \emph{COLING}, 2020, pp. 1737--1743.

\bibitem{PKGC}
X.~Lv, Y.~Lin, Y.~Cao, L.~Hou, J.~Li, Z.~Liu, P.~Li, and J.~Zhou, ``Do pre-trained models benefit knowledge graph completion? {A} reliable evaluation and a reasonable approach,'' in \emph{ACL}, 2022, pp. 3570--3581.

\bibitem{LASS}
J.~Shen, C.~Wang, L.~Gong, and D.~Song, ``Joint language semantic and structure embedding for knowledge graph completion,'' in \emph{COLING}, 2022, pp. 1965--1978.

\bibitem{MEM-KGC}
B.~Choi, D.~Jang, and Y.~Ko, ``{MEM-KGC:} masked entity model for knowledge graph completion with pre-trained language model,'' \emph{{IEEE} Access}, vol.~9, pp. 132\,025--132\,032, 2021.

\bibitem{open-world-KGC}
B.~Choi and Y.~Ko, ``Knowledge graph extension with a pre-trained language model via unified learning method,'' \emph{Knowl. Based Syst.}, vol. 262, p. 110245, 2023.

\bibitem{StAR}
B.~Wang, T.~Shen, G.~Long, T.~Zhou, Y.~Wang, and Y.~Chang, ``Structure-augmented text representation learning for efficient knowledge graph completion,'' in \emph{WWW}, 2021, pp. 1737--1748.

\bibitem{SimKGC}
L.~Wang, W.~Zhao, Z.~Wei, and J.~Liu, ``Simkgc: Simple contrastive knowledge graph completion with pre-trained language models,'' in \emph{ACL}, 2022, pp. 4281--4294.

\bibitem{LP-BERT}
D.~Li, M.~Yi, and Y.~He, ``Lp-bert: Multi-task pre-training knowledge graph bert for link prediction,'' \emph{arXiv preprint arXiv:2201.04843}, 2022.

\bibitem{KGT5}
A.~Saxena, A.~Kochsiek, and R.~Gemulla, ``Sequence-to-sequence knowledge graph completion and question answering,'' in \emph{ACL}, 2022, pp. 2814--2828.

\bibitem{KG-S2S}
C.~Chen, Y.~Wang, B.~Li, and K.~Lam, ``Knowledge is flat: {A} seq2seq generative framework for various knowledge graph completion,'' in \emph{COLING}, 2022, pp. 4005--4017.

\bibitem{Elmo}
M.~E. Peters, M.~Neumann, M.~Iyyer, M.~Gardner, C.~Clark, K.~Lee, and L.~Zettlemoyer, ``Deep contextualized word representations,'' in \emph{NAACL}, 2018, pp. 2227--2237.

\bibitem{generativeNER}
H.~Yan, T.~Gui, J.~Dai, Q.~Guo, Z.~Zhang, and X.~Qiu, ``A unified generative framework for various {NER} subtasks,'' in \emph{ACL}, 2021, pp. 5808--5822.

\bibitem{LDET}
Y.~Onoe and G.~Durrett, ``Learning to denoise distantly-labeled data for entity typing,'' in \emph{NAACL}, 2019, pp. 2407--2417.

\bibitem{BOX4Types}
Y.~Onoe, M.~Boratko, A.~McCallum, and G.~Durrett, ``Modeling fine-grained entity types with box embeddings,'' in \emph{ACL}, 2021, pp. 2051--2064.

\bibitem{ELQ}
B.~Z. Li, S.~Min, S.~Iyer, Y.~Mehdad, and W.~Yih, ``Efficient one-pass end-to-end entity linking for questions,'' in \emph{EMNLP}, 2020, pp. 6433--6441.

\bibitem{ReFinED}
T.~Ayoola, S.~Tyagi, J.~Fisher, C.~Christodoulopoulos, and A.~Pierleoni, ``Refined: An efficient zero-shot-capable approach to end-to-end entity linking,'' in \emph{NAACL}, 2022, pp. 209--220.

\bibitem{CR1}
M.~Joshi, O.~Levy, L.~Zettlemoyer, and D.~S. Weld, ``{BERT} for coreference resolution: Baselines and analysis,'' in \emph{EMNLP}, 2019, pp. 5802--5807.

\bibitem{SpanBERT}
M.~Joshi, D.~Chen, Y.~Liu, D.~S. Weld, L.~Zettlemoyer, and O.~Levy, ``Spanbert: Improving pre-training by representing and predicting spans,'' \emph{Trans. Assoc. Comput. Linguistics}, vol.~8, pp. 64--77, 2020.

\bibitem{CDLM}
A.~Caciularu, A.~Cohan, I.~Beltagy, M.~E. Peters, A.~Cattan, and I.~Dagan, ``{CDLM:} cross-document language modeling,'' in \emph{EMNLP}, 2021, pp. 2648--2662.

\bibitem{crossCR}
A.~Cattan, A.~Eirew, G.~Stanovsky, M.~Joshi, and I.~Dagan, ``Cross-document coreference resolution over predicted mentions,'' in \emph{ACL}, 2021, pp. 5100--5107.

\bibitem{CR-RL}
Y.~Wang, Y.~Shen, and H.~Jin, ``An end-to-end actor-critic-based neural coreference resolution system,'' in \emph{{IEEE} International Conference on Acoustics, Speech and Signal Processing, {ICASSP} 2021, Toronto, ON, Canada, June 6-11, 2021}, 2021, pp. 7848--7852.

\bibitem{sent-re1}
P.~Shi and J.~Lin, ``Simple {BERT} models for relation extraction and semantic role labeling,'' \emph{CoRR}, vol. abs/1904.05255, 2019.

\bibitem{Curriculum-RE}
S.~Park and H.~Kim, ``Improving sentence-level relation extraction through curriculum learning,'' \emph{CoRR}, vol. abs/2107.09332, 2021.

\bibitem{DREEAM}
Y.~Ma, A.~Wang, and N.~Okazaki, ``{DREEAM:} guiding attention with evidence for improving document-level relation extraction,'' in \emph{EACL}, 2023, pp. 1963--1975.

\bibitem{guo2021constructing}
Q.~Guo, Y.~Sun, G.~Liu, Z.~Wang, Z.~Ji, Y.~Shen, and X.~Wang, ``Constructing chinese historical literature knowledge graph based on bert,'' in \emph{Web Information Systems and Applications: 18th International Conference, WISA 2021, Kaifeng, China, September 24--26, 2021, Proceedings 18}.\hskip 1em plus 0.5em minus 0.4em\relax Springer, 2021, pp. 323--334.

\bibitem{han2023pive}
J.~Han, N.~Collier, W.~Buntine, and E.~Shareghi, ``Pive: Prompting with iterative verification improving graph-based generative capability of llms,'' \emph{arXiv preprint arXiv:2305.12392}, 2023.

\bibitem{bosselut2019comet}
A.~Bosselut, H.~Rashkin, M.~Sap, C.~Malaviya, A.~Celikyilmaz, and Y.~Choi, ``Comet: Commonsense transformers for knowledge graph construction,'' in \emph{ACL}, 2019.

\bibitem{hao2022bertnet}
S.~Hao, B.~Tan, K.~Tang, H.~Zhang, E.~P. Xing, and Z.~Hu, ``Bertnet: Harvesting knowledge graphs from pretrained language models,'' \emph{arXiv preprint arXiv:2206.14268}, 2022.

\bibitem{west2022symbolic}
P.~West, C.~Bhagavatula, J.~Hessel, J.~Hwang, L.~Jiang, R.~Le~Bras, X.~Lu, S.~Welleck, and Y.~Choi, ``Symbolic knowledge distillation: from general language models to commonsense models,'' in \emph{NAACL}, 2022, pp. 4602--4625.

\bibitem{ribeiro-etal-2021-investigating}
L.~F.~R. Ribeiro, M.~Schmitt, H.~Sch{\"u}tze, and I.~Gurevych, ``Investigating pretrained language models for graph-to-text generation,'' in \emph{Proceedings of the 3rd Workshop on Natural Language Processing for Conversational AI}, 2021, pp. 211--227.

\bibitem{li-etal-2021-shot-knowledge}
J.~Li, T.~Tang, W.~X. Zhao, Z.~Wei, N.~J. Yuan, and J.-R. Wen, ``Few-shot knowledge graph-to-text generation with pretrained language models,'' in \emph{ACL}, 2021, pp. 1558--1568.

\bibitem{colas-etal-2022-gap}
A.~Colas, M.~Alvandipour, and D.~Z. Wang, ``{GAP}: A graph-aware language model framework for knowledge graph-to-text generation,'' in \emph{Proceedings of the 29th International Conference on Computational Linguistics}, 2022, pp. 5755--5769.

\bibitem{jin-etal-2020-genwiki}
Z.~Jin, Q.~Guo, X.~Qiu, and Z.~Zhang, ``{G}en{W}iki: A dataset of 1.3 million content-sharing text and graphs for unsupervised graph-to-text generation,'' in \emph{Proceedings of the 28th International Conference on Computational Linguistics}, 2020, pp. 2398--2409.

\bibitem{chen-etal-2020-kgpt}
W.~Chen, Y.~Su, X.~Yan, and W.~Y. Wang, ``{KGPT}: Knowledge-grounded pre-training for data-to-text generation,'' in \emph{EMNLP}, 2020, pp. 8635--8648.

\bibitem{lukovnikov2019pretrained}
D.~Lukovnikov, A.~Fischer, and J.~Lehmann, ``Pretrained transformers for simple question answering over knowledge graphs,'' in \emph{The Semantic Web--ISWC 2019: 18th International Semantic Web Conference, Auckland, New Zealand, October 26--30, 2019, Proceedings, Part I 18}.\hskip 1em plus 0.5em minus 0.4em\relax Springer, 2019, pp. 470--486.

\bibitem{luo2020bert}
D.~Luo, J.~Su, and S.~Yu, ``A bert-based approach with relation-aware attention for knowledge base question answering,'' in \emph{IJCNN}.\hskip 1em plus 0.5em minus 0.4em\relax IEEE, 2020, pp. 1--8.

\bibitem{hu2023empirical}
N.~Hu, Y.~Wu, G.~Qi, D.~Min, J.~Chen, J.~Z. Pan, and Z.~Ali, ``An empirical study of pre-trained language models in simple knowledge graph question answering,'' \emph{arXiv preprint arXiv:2303.10368}, 2023.

\bibitem{xu2021fusing}
Y.~Xu, C.~Zhu, R.~Xu, Y.~Liu, M.~Zeng, and X.~Huang, ``Fusing context into knowledge graph for commonsense question answering,'' in \emph{ACL}, 2021, pp. 1201--1207.

\bibitem{zhang2022drlk}
M.~Zhang, R.~Dai, M.~Dong, and T.~He, ``Drlk: Dynamic hierarchical reasoning with language model and knowledge graph for question answering,'' in \emph{EMNLP}, 2022, pp. 5123--5133.

\bibitem{hu2022empowering}
Z.~Hu, Y.~Xu, W.~Yu, S.~Wang, Z.~Yang, C.~Zhu, K.-W. Chang, and Y.~Sun, ``Empowering language models with knowledge graph reasoning for open-domain question answering,'' in \emph{EMNLP}, 2022, pp. 9562--9581.

\bibitem{zhang2022greaselm}
X.~Zhang, A.~Bosselut, M.~Yasunaga, H.~Ren, P.~Liang, C.~D. Manning, and J.~Leskovec, ``Greaselm: Graph reasoning enhanced language models,'' in \emph{ICLR}, 2022.

\bibitem{cao2022relmkg}
X.~Cao and Y.~Liu, ``Relmkg: reasoning with pre-trained language models and knowledge graphs for complex question answering,'' \emph{Applied Intelligence}, pp. 1--15, 2022.

\bibitem{huang2019knowledge}
X.~Huang, J.~Zhang, D.~Li, and P.~Li, ``Knowledge graph embedding based question answering,'' in \emph{WSDM}, 2019, pp. 105--113.

\bibitem{wang2018dkn}
H.~Wang, F.~Zhang, X.~Xie, and M.~Guo, ``Dkn: Deep knowledge-aware network for news recommendation,'' in \emph{WWW}, 2018, pp. 1835--1844.

\bibitem{yang2015embedding}
B.~Yang, S.~W.-t. Yih, X.~He, J.~Gao, and L.~Deng, ``Embedding entities and relations for learning and inference in knowledge bases,'' in \emph{ICLR}, 2015.

\bibitem{xiong2018one}
W.~Xiong, M.~Yu, S.~Chang, X.~Guo, and W.~Y. Wang, ``One-shot relational learning for knowledge graphs,'' in \emph{EMNLP}, 2018, pp. 1980--1990.

\bibitem{wang2019logic}
P.~Wang, J.~Han, C.~Li, and R.~Pan, ``Logic attention based neighborhood aggregation for inductive knowledge graph embedding,'' in \emph{AAAI}, vol.~33, no.~01, 2019, pp. 7152--7159.

\bibitem{lin2015learning}
Y.~Lin, Z.~Liu, M.~Sun, Y.~Liu, and X.~Zhu, ``Learning entity and relation embeddings for knowledge graph completion,'' in \emph{Proceedings of the AAAI conference on artificial intelligence}, vol.~29, no.~1, 2015.

\bibitem{CSProm-KG}
C.~Chen, Y.~Wang, A.~Sun, B.~Li, and L.~Kwok-Yan, ``Dipping plms sauce: Bridging structure and text for effective knowledge graph completion via conditional soft prompting,'' in \emph{ACL}, 2023.

\bibitem{KGC_analysis}
J.~Lovelace and C.~P. Ros{\'{e}}, ``A framework for adapting pre-trained language models to knowledge graph completion,'' in \emph{Proceedings of the 2022 Conference on Empirical Methods in Natural Language Processing, {EMNLP} 2022, Abu Dhabi, United Arab Emirates, December 7-11, 2022}, 2022, pp. 5937--5955.

\bibitem{Larger-Context-Tagging}
J.~Fu, L.~Feng, Q.~Zhang, X.~Huang, and P.~Liu, ``Larger-context tagging: When and why does it work?'' in \emph{Proceedings of the 2021 Conference of the North American Chapter of the Association for Computational Linguistics: Human Language Technologies, {NAACL-HLT} 2021, Online, June 6-11, 2021}, 2021, pp. 1463--1475.

\bibitem{Ptuning-v2}
X.~Liu, K.~Ji, Y.~Fu, Z.~Du, Z.~Yang, and J.~Tang, ``P-tuning v2: Prompt tuning can be comparable to fine-tuning universally across scales and tasks,'' \emph{CoRR}, vol. abs/2110.07602, 2021.

\bibitem{NER-as-DP}
J.~Yu, B.~Bohnet, and M.~Poesio, ``Named entity recognition as dependency parsing,'' in \emph{ACL}, 2020, pp. 6470--6476.

\bibitem{DiscontinuousNER}
F.~Li, Z.~Lin, M.~Zhang, and D.~Ji, ``A span-based model for joint overlapped and discontinuous named entity recognition,'' in \emph{ACL}, 2021, pp. 4814--4828.

\bibitem{Span-based-NER1}
C.~Tan, W.~Qiu, M.~Chen, R.~Wang, and F.~Huang, ``Boundary enhanced neural span classification for nested named entity recognition,'' in \emph{The Thirty-Fourth {AAAI} Conference on Artificial Intelligence, {AAAI} 2020, The Thirty-Second Innovative Applications of Artificial Intelligence Conference, {IAAI} 2020, The Tenth {AAAI} Symposium on Educational Advances in Artificial Intelligence, {EAAI} 2020, New York, NY, USA, February 7-12, 2020}, 2020, pp. 9016--9023.

\bibitem{Span-based-NER2}
Y.~Xu, H.~Huang, C.~Feng, and Y.~Hu, ``A supervised multi-head self-attention network for nested named entity recognition,'' in \emph{Thirty-Fifth {AAAI} Conference on Artificial Intelligence, {AAAI} 2021, Thirty-Third Conference on Innovative Applications of Artificial Intelligence, {IAAI} 2021, The Eleventh Symposium on Educational Advances in Artificial Intelligence, {EAAI} 2021, Virtual Event, February 2-9, 2021}, 2021, pp. 14\,185--14\,193.

\bibitem{Span-based-NER3}
J.~Yu, B.~Ji, S.~Li, J.~Ma, H.~Liu, and H.~Xu, ``{S-NER:} {A} concise and efficient span-based model for named entity recognition,'' \emph{Sensors}, vol.~22, no.~8, p. 2852, 2022.

\bibitem{parserNER1}
Y.~Fu, C.~Tan, M.~Chen, S.~Huang, and F.~Huang, ``Nested named entity recognition with partially-observed treecrfs,'' in \emph{AAAI}, 2021, pp. 12\,839--12\,847.

\bibitem{parserNER2}
C.~Lou, S.~Yang, and K.~Tu, ``Nested named entity recognition as latent lexicalized constituency parsing,'' in \emph{Proceedings of the 60th Annual Meeting of the Association for Computational Linguistics (Volume 1: Long Papers), {ACL} 2022, Dublin, Ireland, May 22-27, 2022}, 2022, pp. 6183--6198.

\bibitem{parserNER3}
S.~Yang and K.~Tu, ``Bottom-up constituency parsing and nested named entity recognition with pointer networks,'' in \emph{Proceedings of the 60th Annual Meeting of the Association for Computational Linguistics (Volume 1: Long Papers), {ACL} 2022, Dublin, Ireland, May 22-27, 2022}, 2022, pp. 2403--2416.

\bibitem{discontinuousNER1}
F.~Li, Z.~Lin, M.~Zhang, and D.~Ji, ``A span-based model for joint overlapped and discontinuous named entity recognition,'' in \emph{Proceedings of the 59th Annual Meeting of the Association for Computational Linguistics and the 11th International Joint Conference on Natural Language Processing, {ACL/IJCNLP} 2021, (Volume 1: Long Papers), Virtual Event, August 1-6, 2021}, 2021, pp. 4814--4828.

\bibitem{LRN}
Q.~Liu, H.~Lin, X.~Xiao, X.~Han, L.~Sun, and H.~Wu, ``Fine-grained entity typing via label reasoning,'' in \emph{Proceedings of the 2021 Conference on Empirical Methods in Natural Language Processing, {EMNLP} 2021, Virtual Event / Punta Cana, Dominican Republic, 7-11 November, 2021}, 2021, pp. 4611--4622.

\bibitem{MLMET}
H.~Dai, Y.~Song, and H.~Wang, ``Ultra-fine entity typing with weak supervision from a masked language model,'' in \emph{Proceedings of the 59th Annual Meeting of the Association for Computational Linguistics and the 11th International Joint Conference on Natural Language Processing, {ACL/IJCNLP} 2021, (Volume 1: Long Papers), Virtual Event, August 1-6, 2021}, 2021, pp. 1790--1799.

\bibitem{PL}
N.~Ding, Y.~Chen, X.~Han, G.~Xu, X.~Wang, P.~Xie, H.~Zheng, Z.~Liu, J.~Li, and H.~Kim, ``Prompt-learning for fine-grained entity typing,'' in \emph{Findings of the Association for Computational Linguistics: {EMNLP} 2022, Abu Dhabi, United Arab Emirates, December 7-11, 2022}, 2022, pp. 6888--6901.

\bibitem{DFET}
W.~Pan, W.~Wei, and F.~Zhu, ``Automatic noisy label correction for fine-grained entity typing,'' in \emph{Proceedings of the Thirty-First International Joint Conference on Artificial Intelligence, {IJCAI} 2022, Vienna, Austria, 23-29 July 2022}, 2022, pp. 4317--4323.

\bibitem{LITE}
B.~Li, W.~Yin, and M.~Chen, ``Ultra-fine entity typing with indirect supervision from natural language inference,'' \emph{Trans. Assoc. Comput. Linguistics}, vol.~10, pp. 607--622, 2022.

\bibitem{EL1}
S.~Broscheit, ``Investigating entity knowledge in {BERT} with simple neural end-to-end entity linking,'' \emph{CoRR}, vol. abs/2003.05473, 2020.

\bibitem{GENRE}
N.~D. Cao, G.~Izacard, S.~Riedel, and F.~Petroni, ``Autoregressive entity retrieval,'' in \emph{9th ICLR, {ICLR} 2021, Virtual Event, Austria, May 3-7, 2021}, 2021.

\bibitem{mGENRE}
N.~D. Cao, L.~Wu, K.~Popat, M.~Artetxe, N.~Goyal, M.~Plekhanov, L.~Zettlemoyer, N.~Cancedda, S.~Riedel, and F.~Petroni, ``Multilingual autoregressive entity linking,'' \emph{Trans. Assoc. Comput. Linguistics}, vol.~10, pp. 274--290, 2022.

\bibitem{EL2}
N.~D. Cao, W.~Aziz, and I.~Titov, ``Highly parallel autoregressive entity linking with discriminative correction,'' in \emph{Proceedings of the 2021 Conference on Empirical Methods in Natural Language Processing, {EMNLP} 2021, Virtual Event / Punta Cana, Dominican Republic, 7-11 November, 2021}, 2021, pp. 7662--7669.

\bibitem{LSTM-CR}
K.~Lee, L.~He, and L.~Zettlemoyer, ``Higher-order coreference resolution with coarse-to-fine inference,'' in \emph{NAACL}, 2018, pp. 687--692.

\bibitem{CR2}
T.~M. Lai, T.~Bui, and D.~S. Kim, ``End-to-end neural coreference resolution revisited: {A} simple yet effective baseline,'' in \emph{{IEEE} International Conference on Acoustics, Speech and Signal Processing, {ICASSP} 2022, Virtual and Singapore, 23-27 May 2022}, 2022, pp. 8147--8151.

\bibitem{CorefQA}
W.~Wu, F.~Wang, A.~Yuan, F.~Wu, and J.~Li, ``Corefqa: Coreference resolution as query-based span prediction,'' in \emph{Proceedings of the 58th Annual Meeting of the Association for Computational Linguistics, {ACL} 2020, Online, July 5-10, 2020}, 2020, pp. 6953--6963.

\bibitem{CR3}
T.~M. Lai, H.~Ji, T.~Bui, Q.~H. Tran, F.~Dernoncourt, and W.~Chang, ``A context-dependent gated module for incorporating symbolic semantics into event coreference resolution,'' in \emph{Proceedings of the 2021 Conference of the North American Chapter of the Association for Computational Linguistics: Human Language Technologies, {NAACL-HLT} 2021, Online, June 6-11, 2021}, 2021, pp. 3491--3499.

\bibitem{efficientCR1}
Y.~Kirstain, O.~Ram, and O.~Levy, ``Coreference resolution without span representations,'' in \emph{Proceedings of the 59th Annual Meeting of the Association for Computational Linguistics and the 11th International Joint Conference on Natural Language Processing, {ACL/IJCNLP} 2021, (Volume 2: Short Papers), Virtual Event, August 1-6, 2021}, 2021, pp. 14--19.

\bibitem{efficientCR2}
R.~Thirukovalluru, N.~Monath, K.~Shridhar, M.~Zaheer, M.~Sachan, and A.~McCallum, ``Scaling within document coreference to long texts,'' in \emph{Findings of the Association for Computational Linguistics: {ACL/IJCNLP} 2021, Online Event, August 1-6, 2021}, ser. Findings of {ACL}, vol. {ACL/IJCNLP} 2021, 2021, pp. 3921--3931.

\bibitem{Longformer}
I.~Beltagy, M.~E. Peters, and A.~Cohan, ``Longformer: The long-document transformer,'' \emph{CoRR}, vol. abs/2004.05150, 2020.

\bibitem{TRE}
C.~Alt, M.~H{\"{u}}bner, and L.~Hennig, ``Improving relation extraction by pre-trained language representations,'' in \emph{1st Conference on Automated Knowledge Base Construction, {AKBC} 2019, Amherst, MA, USA, May 20-22, 2019}, 2019.

\bibitem{BERT-MTB}
L.~B. Soares, N.~FitzGerald, J.~Ling, and T.~Kwiatkowski, ``Matching the blanks: Distributional similarity for relation learning,'' in \emph{ACL}, 2019, pp. 2895--2905.

\bibitem{RECENT}
S.~Lyu and H.~Chen, ``Relation classification with entity type restriction,'' in \emph{Findings of the Association for Computational Linguistics: {ACL/IJCNLP} 2021, Online Event, August 1-6, 2021}, ser. Findings of {ACL}, vol. {ACL/IJCNLP} 2021, 2021, pp. 390--395.

\bibitem{sent-re3}
J.~Zheng and Z.~Chen, ``Sentence-level relation extraction via contrastive learning with descriptive relation prompts,'' \emph{CoRR}, vol. abs/2304.04935, 2023.

\bibitem{docRE1}
H.~Wang, C.~Focke, R.~Sylvester, N.~Mishra, and W.~Y. Wang, ``Fine-tune bert for docred with two-step process,'' \emph{CoRR}, vol. abs/1909.11898, 2019.

\bibitem{HIN}
H.~Tang, Y.~Cao, Z.~Zhang, J.~Cao, F.~Fang, S.~Wang, and P.~Yin, ``{HIN:} hierarchical inference network for document-level relation extraction,'' in \emph{PAKDD}, ser. Lecture Notes in Computer Science, vol. 12084, 2020, pp. 197--209.

\bibitem{GLRE}
D.~Wang, W.~Hu, E.~Cao, and W.~Sun, ``Global-to-local neural networks for document-level relation extraction,'' in \emph{Proceedings of the 2020 Conference on Empirical Methods in Natural Language Processing, {EMNLP} 2020, Online, November 16-20, 2020}, 2020, pp. 3711--3721.

\bibitem{SIRE}
S.~Zeng, Y.~Wu, and B.~Chang, ``{SIRE:} separate intra- and inter-sentential reasoning for document-level relation extraction,'' in \emph{Findings of the Association for Computational Linguistics: {ACL/IJCNLP} 2021, Online Event, August 1-6, 2021}, ser. Findings of {ACL}, vol. {ACL/IJCNLP} 2021, 2021, pp. 524--534.

\bibitem{LSR}
G.~Nan, Z.~Guo, I.~Sekulic, and W.~Lu, ``Reasoning with latent structure refinement for document-level relation extraction,'' in \emph{ACL}, 2020, pp. 1546--1557.

\bibitem{GAIN}
S.~Zeng, R.~Xu, B.~Chang, and L.~Li, ``Double graph based reasoning for document-level relation extraction,'' in \emph{Proceedings of the 2020 Conference on Empirical Methods in Natural Language Processing, {EMNLP} 2020, Online, November 16-20, 2020}, 2020, pp. 1630--1640.

\bibitem{DocuNet}
N.~Zhang, X.~Chen, X.~Xie, S.~Deng, C.~Tan, M.~Chen, F.~Huang, L.~Si, and H.~Chen, ``Document-level relation extraction as semantic segmentation,'' in \emph{IJCAI}, 2021, pp. 3999--4006.

\bibitem{U-net}
O.~Ronneberger, P.~Fischer, and T.~Brox, ``U-net: Convolutional networks for biomedical image segmentation,'' in \emph{Medical Image Computing and Computer-Assisted Intervention - {MICCAI} 2015 - 18th International Conference Munich, Germany, October 5 - 9, 2015, Proceedings, Part {III}}, ser. Lecture Notes in Computer Science, vol. 9351, 2015, pp. 234--241.

\bibitem{ATLOP}
W.~Zhou, K.~Huang, T.~Ma, and J.~Huang, ``Document-level relation extraction with adaptive thresholding and localized context pooling,'' in \emph{AAAI}, 2021, pp. 14\,612--14\,620.

\bibitem{gardent-etal-2017-webnlg}
C.~Gardent, A.~Shimorina, S.~Narayan, and L.~Perez-Beltrachini, ``The {W}eb{NLG} challenge: Generating text from {RDF} data,'' in \emph{Proceedings of the 10th International Conference on Natural Language Generation}, 2017, pp. 124--133.

\bibitem{DBLP:conf/aaai/GuanWH19}
J.~Guan, Y.~Wang, and M.~Huang, ``Story ending generation with incremental encoding and commonsense knowledge,'' in \emph{AAAI}, 2019, pp. 6473--6480.

\bibitem{DBLP:conf/ijcai/ZhouYHZXZ18}
H.~Zhou, T.~Young, M.~Huang, H.~Zhao, J.~Xu, and X.~Zhu, ``Commonsense knowledge aware conversation generation with graph attention,'' in \emph{IJCAI}, 2018, pp. 4623--4629.

\bibitem{kale-rastogi-2020-text}
M.~Kale and A.~Rastogi, ``Text-to-text pre-training for data-to-text tasks,'' in \emph{Proceedings of the 13th International Conference on Natural Language Generation}, 2020, pp. 97--102.

\bibitem{mintz-etal-2009-distant}
M.~Mintz, S.~Bills, R.~Snow, and D.~Jurafsky, ``Distant supervision for relation extraction without labeled data,'' in \emph{ACL}, 2009, pp. 1003--1011.

\bibitem{saxena2020improving}
A.~Saxena, A.~Tripathi, and P.~Talukdar, ``Improving multi-hop question answering over knowledge graphs using knowledge base embeddings,'' in \emph{ACL}, 2020, pp. 4498--4507.

\bibitem{feng-etal-2020-scalable}
Y.~Feng, X.~Chen, B.~Y. Lin, P.~Wang, J.~Yan, and X.~Ren, ``Scalable multi-hop relational reasoning for knowledge-aware question answering,'' in \emph{EMNLP}, 2020, pp. 1295--1309.

\bibitem{yan2021large}
Y.~Yan, R.~Li, S.~Wang, H.~Zhang, Z.~Daoguang, F.~Zhang, W.~Wu, and W.~Xu, ``Large-scale relation learning for question answering over knowledge bases with pre-trained language models,'' in \emph{EMNLP}, 2021, pp. 3653--3660.

\bibitem{zhang2022subgraph}
J.~Zhang, X.~Zhang, J.~Yu, J.~Tang, J.~Tang, C.~Li, and H.~Chen, ``Subgraph retrieval enhanced model for multi-hop knowledge base question answering,'' in \emph{ACL (Volume 1: Long Papers)}, 2022, pp. 5773--5784.

\bibitem{jiang2023structgpt}
J.~Jiang, K.~Zhou, Z.~Dong, K.~Ye, W.~X. Zhao, and J.-R. Wen, ``Structgpt: A general framework for large language model to reason over structured data,'' \emph{arXiv preprint arXiv:2305.09645}, 2023.

\bibitem{zhu2023pre}
H.~Zhu, H.~Peng, Z.~Lyu, L.~Hou, J.~Li, and J.~Xiao, ``Pre-training language model incorporating domain-specific heterogeneous knowledge into a unified representation,'' \emph{Expert Systems with Applications}, vol. 215, p. 119369, 2023.

\bibitem{feng2023knowledge}
C.~Feng, X.~Zhang, and Z.~Fei, ``Knowledge solver: Teaching llms to search for domain knowledge from knowledge graphs,'' \emph{arXiv preprint arXiv:2309.03118}, 2023.

\bibitem{sun2023think}
J.~Sun, C.~Xu, L.~Tang, S.~Wang, C.~Lin, Y.~Gong, H.-Y. Shum, and J.~Guo, ``Think-on-graph: Deep and responsible reasoning of large language model with knowledge graph,'' \emph{arXiv preprint arXiv:2307.07697}, 2023.

\bibitem{he-etal-2020-bert}
B.~He, D.~Zhou, J.~Xiao, X.~Jiang, Q.~Liu, N.~J. Yuan, and T.~Xu, ``{BERT}-{MK}: Integrating graph contextualized knowledge into pre-trained language models,'' in \emph{EMNLP}, 2020, pp. 2281--2290.

\bibitem{SU2021127}
Y.~Su, X.~Han, Z.~Zhang, Y.~Lin, P.~Li, Z.~Liu, J.~Zhou, and M.~Sun, ``Cokebert: Contextual knowledge selection and embedding towards enhanced pre-trained language models,'' \emph{AI Open}, vol.~2, pp. 127--134, 2021.

\bibitem{DBLP:conf/aaai/Yu0Y022}
D.~Yu, C.~Zhu, Y.~Yang, and M.~Zeng, ``{JAKET:} joint pre-training of knowledge graph and language understanding,'' in \emph{AAAI}, 2022, pp. 11\,630--11\,638.

\bibitem{DBLP:conf/aaai/WangKMYTACFMMW19}
X.~Wang, P.~Kapanipathi, R.~Musa, M.~Yu, K.~Talamadupula, I.~Abdelaziz, M.~Chang, A.~Fokoue, B.~Makni, N.~Mattei, and M.~Witbrock, ``Improving natural language inference using external knowledge in the science questions domain,'' in \emph{AAAI}, 2019, pp. 7208--7215.

\bibitem{sun-etal-2022-jointlk}
Y.~Sun, Q.~Shi, L.~Qi, and Y.~Zhang, ``{J}oint{LK}: Joint reasoning with language models and knowledge graphs for commonsense question answering,'' in \emph{NAACL}, 2022, pp. 5049--5060.

\bibitem{liu2023agentbench}
X.~Liu, H.~Yu, H.~Zhang, Y.~Xu, X.~Lei, H.~Lai, Y.~Gu, H.~Ding, K.~Men, K.~Yang \emph{et~al.}, ``Agentbench: Evaluating llms as agents,'' \emph{arXiv preprint arXiv:2308.03688}, 2023.

\bibitem{wang2023knowledge}
Y.~Wang, N.~Lipka, R.~A. Rossi, A.~Siu, R.~Zhang, and T.~Derr, ``Knowledge graph prompting for multi-document question answering,'' \emph{arXiv preprint arXiv:2308.11730}, 2023.

\bibitem{zeng2023agenttuning}
A.~Zeng, M.~Liu, R.~Lu, B.~Wang, X.~Liu, Y.~Dong, and J.~Tang, ``Agenttuning: Enabling generalized agent abilities for llms,'' 2023.

\bibitem{kryscinski2019evaluating}
W.~Kry{\'s}ci{\'n}ski, B.~McCann, C.~Xiong, and R.~Socher, ``Evaluating the factual consistency of abstractive text summarization,'' \emph{arXiv preprint arXiv:1910.12840}, 2019.

\bibitem{ji2022rho}
Z.~Ji, Z.~Liu, N.~Lee, T.~Yu, B.~Wilie, M.~Zeng, and P.~Fung, ``Rho ($\backslash \rho $): Reducing hallucination in open-domain dialogues with knowledge grounding,'' \emph{arXiv preprint arXiv:2212.01588}, 2022.

\bibitem{feng2023factkb}
S.~Feng, V.~Balachandran, Y.~Bai, and Y.~Tsvetkov, ``Factkb: Generalizable factuality evaluation using language models enhanced with factual knowledge,'' \emph{arXiv preprint arXiv:2305.08281}, 2023.

\bibitem{yao2023editing}
Y.~Yao, P.~Wang, B.~Tian, S.~Cheng, Z.~Li, S.~Deng, H.~Chen, and N.~Zhang, ``Editing large language models: Problems, methods, and opportunities,'' \emph{arXiv preprint arXiv:2305.13172}, 2023.

\bibitem{li2023unveiling}
Z.~Li, N.~Zhang, Y.~Yao, M.~Wang, X.~Chen, and H.~Chen, ``Unveiling the pitfalls of knowledge editing for large language models,'' \emph{arXiv preprint arXiv:2310.02129}, 2023.

\bibitem{cohen2023evaluating}
R.~Cohen, E.~Biran, O.~Yoran, A.~Globerson, and M.~Geva, ``Evaluating the ripple effects of knowledge editing in language models,'' \emph{arXiv preprint arXiv:2307.12976}, 2023.

\bibitem{diao2022black}
S.~Diao, Z.~Huang, R.~Xu, X.~Li, Y.~Lin, X.~Zhou, and T.~Zhang, ``Black-box prompt learning for pre-trained language models,'' \emph{arXiv preprint arXiv:2201.08531}, 2022.

\bibitem{sun2022black}
T.~Sun, Y.~Shao, H.~Qian, X.~Huang, and X.~Qiu, ``Black-box tuning for language-model-as-a-service,'' in \emph{International Conference on Machine Learning}.\hskip 1em plus 0.5em minus 0.4em\relax PMLR, 2022, pp. 20\,841--20\,855.

\bibitem{challenge-mm-1}
X.~Chen, A.~Shrivastava, and A.~Gupta, ``{NEIL:} extracting visual knowledge from web data,'' in \emph{{IEEE} International Conference on Computer Vision, {ICCV} 2013, Sydney, Australia, December 1-8, 2013}, 2013, pp. 1409--1416.

\bibitem{challenge-mm-2}
M.~Warren and P.~J. Hayes, ``Bounding ambiguity: Experiences with an image annotation system,'' in \emph{Proceedings of the 1st Workshop on Subjectivity, Ambiguity and Disagreement in Crowdsourcing}, ser. {CEUR} Workshop Proceedings, vol. 2276, 2018, pp. 41--54.

\bibitem{chen2022lako}
Z.~Chen, Y.~Huang, J.~Chen, Y.~Geng, Y.~Fang, J.~Z. Pan, N.~Zhang, and W.~Zhang, ``Lako: Knowledge-driven visual estion answering via late knowledge-to-text injection,'' 2022.

\bibitem{girdhar2023imagebind}
R.~Girdhar, A.~El-Nouby, Z.~Liu, M.~Singh, K.~V. Alwala, A.~Joulin, and I.~Misra, ``Imagebind: One embedding space to bind them all,'' in \emph{ICCV}, 2023, pp. 15\,180--15\,190.

\bibitem{zhang2020emotion}
J.~Zhang, Z.~Yin, P.~Chen, and S.~Nichele, ``Emotion recognition using multi-modal data and machine learning techniques: A tutorial and review,'' \emph{Information Fusion}, vol.~59, pp. 103--126, 2020.

\bibitem{zhang2022trustworthy}
H.~Zhang, B.~Wu, X.~Yuan, S.~Pan, H.~Tong, and J.~Pei, ``Trustworthy graph neural networks: Aspects, methods and trends,'' \emph{arXiv:2205.07424}, 2022.

\bibitem{wu2022pretrained}
T.~Wu, M.~Caccia, Z.~Li, Y.-F. Li, G.~Qi, and G.~Haffari, ``Pretrained language model in continual learning: A comparative study,'' in \emph{ICLR}, 2022.

\bibitem{li2022systematic}
X.~L. Li, A.~Kuncoro, J.~Hoffmann, C.~de~Masson~d’Autume, P.~Blunsom, and A.~Nematzadeh, ``A systematic investigation of commonsense knowledge in large language models,'' in \emph{Proceedings of the 2022 Conference on Empirical Methods in Natural Language Processing}, 2022, pp. 11\,838--11\,855.

\bibitem{zheng2023large}
Y.~Zheng, H.~Y. Koh, J.~Ju, A.~T. Nguyen, L.~T. May, G.~I. Webb, and S.~Pan, ``Large language models for scientific synthesis, inference and explanation,'' \emph{arXiv preprint arXiv:2310.07984}, 2023.

\bibitem{min2023recent}
B.~Min, H.~Ross, E.~Sulem, A.~P.~B. Veyseh, T.~H. Nguyen, O.~Sainz, E.~Agirre, I.~Heintz, and D.~Roth, ``Recent advances in natural language processing via large pre-trained language models: A survey,'' \emph{ACM Computing Surveys}, vol.~56, no.~2, pp. 1--40, 2023.

\bibitem{wei2021finetuned}
J.~Wei, M.~Bosma, V.~Zhao, K.~Guu, A.~W. Yu, B.~Lester, N.~Du, A.~M. Dai, and Q.~V. Le, ``Finetuned language models are zero-shot learners,'' in \emph{International Conference on Learning Representations}, 2021.

\bibitem{zhang2023hallucination}
Y.~Zhang, Y.~Li, L.~Cui, D.~Cai, L.~Liu, T.~Fu, X.~Huang, E.~Zhao, Y.~Zhang, Y.~Chen, L.~Wang, A.~T. Luu, W.~Bi, F.~Shi, and S.~Shi, ``Siren's song in the ai ocean: A survey on hallucination in large language models,'' \emph{arXiv preprint arXiv:2309.01219}, 2023.

\end{thebibliography}


\begin{thebibliography}{100}
\providecommand{\url}[1]{#1}
\csname url@samestyle\endcsname
\providecommand{\newblock}{\relax}
\providecommand{\bibinfo}[2]{#2}
\providecommand{\BIBentrySTDinterwordspacing}{\spaceskip=0pt\relax}
\providecommand{\BIBentryALTinterwordstretchfactor}{4}
\providecommand{\BIBentryALTinterwordspacing}{\spaceskip=\fontdimen2\font plus
\BIBentryALTinterwordstretchfactor\fontdimen3\font minus \fontdimen4\font\relax}
\providecommand{\BIBforeignlanguage}[2]{{%
\expandafter\ifx\csname l@#1\endcsname\relax
\typeout{** WARNING: IEEEtran.bst: No hyphenation pattern has been}%
\typeout{** loaded for the language `#1'. Using the pattern for}%
\typeout{** the default language instead.}%
\else
\language=\csname l@#1\endcsname
\fi
#2}}
\providecommand{\BIBdecl}{\relax}
\BIBdecl

\bibitem{zhao2023survey}
W.~X. Zhao, K.~Zhou, J.~Li, T.~Tang, X.~Wang, Y.~Hou, Y.~Min, B.~Zhang, J.~Zhang, Z.~Dong \emph{et~al.}, ``A survey of large language models,'' \emph{arXiv preprint arXiv:2303.18223}, 2023.

\bibitem{qiu2020pre}
X.~Qiu, T.~Sun, Y.~Xu, Y.~Shao, N.~Dai, and X.~Huang, ``Pre-trained models for natural language processing: A survey,'' \emph{Science China Technological Sciences}, vol.~63, no.~10, pp. 1872--1897, 2020.

\bibitem{yang2023harnessing}
J.~Yang, H.~Jin, R.~Tang, X.~Han, Q.~Feng, H.~Jiang, B.~Yin, and X.~Hu, ``Harnessing the power of llms in practice: A survey on chatgpt and beyond,'' \emph{arXiv preprint arXiv:2304.13712}, 2023.

\bibitem{petroni2019language}
F.~Petroni, T.~Rockt{\"a}schel, S.~Riedel, P.~Lewis, A.~Bakhtin, Y.~Wu, and A.~Miller, ``Language models as knowledge bases?'' in \emph{EMNLP-IJCNLP}, 2019, pp. 2463--2473.

\bibitem{ji2023survey}
Z.~Ji, N.~Lee, R.~Frieske, T.~Yu, D.~Su, Y.~Xu, E.~Ishii, Y.~J. Bang, A.~Madotto, and P.~Fung, ``Survey of hallucination in natural language generation,'' \emph{ACM Computing Surveys}, vol.~55, no.~12, pp. 1--38, 2023.

\bibitem{zhang2022survey}
H.~Zhang, H.~Song, S.~Li, M.~Zhou, and D.~Song, ``A survey of controllable text generation using transformer-based pre-trained language models,'' \emph{arXiv preprint arXiv:2201.05337}, 2022.

\bibitem{danilevsky2020survey}
M.~Danilevsky, K.~Qian, R.~Aharonov, Y.~Katsis, B.~Kawas, and P.~Sen, ``A survey of the state of explainable ai for natural language processing,'' \emph{arXiv preprint arXiv:2010.00711}, 2020.

\bibitem{wang2023robustness}
J.~Wang, X.~Hu, W.~Hou, H.~Chen, R.~Zheng, Y.~Wang, L.~Yang, H.~Huang, W.~Ye, X.~Geng \emph{et~al.}, ``On the robustness of chatgpt: An adversarial and out-of-distribution perspective,'' \emph{arXiv preprint arXiv:2302.12095}, 2023.

\bibitem{ji2021survey}
S.~Ji, S.~Pan, E.~Cambria, P.~Marttinen, and S.~Y. Philip, ``A survey on knowledge graphs: Representation, acquisition, and applications,'' \emph{IEEE TNNLS}, vol.~33, no.~2, pp. 494--514, 2021.

\bibitem{vrandevcic2014wikidata}
D.~Vrande{\v{c}}i{\'c} and M.~Kr{\"o}tzsch, ``Wikidata: a free collaborative knowledgebase,'' \emph{Communications of the ACM}, vol.~57, no.~10, pp. 78--85, 2014.

\bibitem{hu2018state}
S.~Hu, L.~Zou, and X.~Zhang, ``A state-transition framework to answer complex questions over knowledge base,'' in \emph{EMNLP}, 2018, pp. 2098--2108.

\bibitem{zhang2021neural}
J.~Zhang, B.~Chen, L.~Zhang, X.~Ke, and H.~Ding, ``Neural, symbolic and neural-symbolic reasoning on knowledge graphs,'' \emph{AI Open}, vol.~2, pp. 14--35, 2021.

\bibitem{abu2021domain}
B.~Abu-Salih, ``Domain-specific knowledge graphs: A survey,'' \emph{Journal of Network and Computer Applications}, vol. 185, p. 103076, 2021.

\bibitem{Mitchell2015NeverEndingL}
T.~Mitchell, W.~Cohen, E.~Hruschka, P.~Talukdar, B.~Yang, J.~Betteridge, A.~Carlson, B.~Dalvi, M.~Gardner, B.~Kisiel, K.~Jayant, L.~Ni, M.~Kathryn, M.~Thahir, N.~Ndapandula, P.~Emmanouil, R.~Alan, S.~Mehdi, S.~Burr, W.~Derry, G.~Abhinav, C.~Xi, S.~Abulhair, and W.~Joel, ``Never-ending learning,'' \emph{Communications of the ACM}, vol.~61, no.~5, pp. 103--115, 2018.

\bibitem{zhong2023comprehensive}
L.~Zhong, J.~Wu, Q.~Li, H.~Peng, and X.~Wu, ``A comprehensive survey on automatic knowledge graph construction,'' \emph{arXiv preprint arXiv:2302.05019}, 2023.

\bibitem{yao2019kg}
L.~Yao, C.~Mao, and Y.~Luo, ``Kg-bert: Bert for knowledge graph completion,'' \emph{arXiv preprint arXiv:1909.03193}, 2019.

\bibitem{luo2023normalizing}
L.~Luo, Y.-F. Li, G.~Haffari, and S.~Pan, ``Normalizing flow-based neural process for few-shot knowledge graph completion,'' \emph{SIGIR}, 2023.

\bibitem{li2022systematic}
X.~L. Li, A.~Kuncoro, J.~Hoffmann, C.~de~Masson~d’Autume, P.~Blunsom, and A.~Nematzadeh, ``A systematic investigation of commonsense knowledge in large language models,'' in \emph{Proceedings of the 2022 Conference on Empirical Methods in Natural Language Processing}, 2022, pp. 11\,838--11\,855.

\bibitem{zheng2023large}
Y.~Zheng, H.~Y. Koh, J.~Ju, A.~T. Nguyen, L.~T. May, G.~I. Webb, and S.~Pan, ``Large language models for scientific synthesis, inference and explanation,'' \emph{arXiv preprint arXiv:2310.07984}, 2023.

\bibitem{min2023recent}
B.~Min, H.~Ross, E.~Sulem, A.~P.~B. Veyseh, T.~H. Nguyen, O.~Sainz, E.~Agirre, I.~Heintz, and D.~Roth, ``Recent advances in natural language processing via large pre-trained language models: A survey,'' \emph{ACM Computing Surveys}, vol.~56, no.~2, pp. 1--40, 2023.

\bibitem{su2019generalizing}
D.~Su, Y.~Xu, G.~I. Winata, P.~Xu, H.~Kim, Z.~Liu, and P.~Fung, ``Generalizing question answering system with pre-trained language model fine-tuning,'' in \emph{Proceedings of the 2nd Workshop on Machine Reading for Question Answering}, 2019, pp. 203--211.

\bibitem{lewis2020bart}
M.~Lewis, Y.~Liu, N.~Goyal, M.~Ghazvininejad, A.~Mohamed, O.~Levy, V.~Stoyanov, and L.~Zettlemoyer, ``Bart: Denoising sequence-to-sequence pre-training for natural language generation, translation, and comprehension,'' in \emph{ACL}, 2020, pp. 7871--7880.

\bibitem{li2021pretrained}
J.~Li, T.~Tang, W.~X. Zhao, and J.-R. Wen, ``Pretrained language models for text generation: A survey,'' \emph{arXiv preprint arXiv:2105.10311}, 2021.

\bibitem{wei2021finetuned}
J.~Wei, M.~Bosma, V.~Zhao, K.~Guu, A.~W. Yu, B.~Lester, N.~Du, A.~M. Dai, and Q.~V. Le, ``Finetuned language models are zero-shot learners,'' in \emph{International Conference on Learning Representations}, 2021.

\bibitem{brown2020language}
T.~Brown, B.~Mann, N.~Ryder, M.~Subbiah, J.~D. Kaplan, P.~Dhariwal, A.~Neelakantan, P.~Shyam, G.~Sastry, A.~Askell \emph{et~al.}, ``Language models are few-shot learners,'' \emph{Advances in neural information processing systems}, vol.~33, pp. 1877--1901, 2020.

\bibitem{raffel2020exploring}
C.~Raffel, N.~Shazeer, A.~Roberts, K.~Lee, S.~Narang, M.~Matena, Y.~Zhou, W.~Li, and P.~J. Liu, ``Exploring the limits of transfer learning with a unified text-to-text transformer,'' \emph{The Journal of Machine Learning Research}, vol.~21, no.~1, pp. 5485--5551, 2020.

\bibitem{zhang2023hallucination}
Y.~Zhang, Y.~Li, L.~Cui, D.~Cai, L.~Liu, T.~Fu, X.~Huang, E.~Zhao, Y.~Zhang, Y.~Chen, L.~Wang, A.~T. Luu, W.~Bi, F.~Shi, and S.~Shi, ``Siren's song in the ai ocean: A survey on hallucination in large language models,'' \emph{arXiv preprint arXiv:2309.01219}, 2023.

\bibitem{bordes2013translating}
A.~Bordes, N.~Usunier, A.~Garcia-Duran, J.~Weston, and O.~Yakhnenko, ``Translating embeddings for modeling multi-relational data,'' \emph{NeurIPS}, vol.~26, 2013.

\bibitem{jiang2023unikgqa}
J.~Jiang, K.~Zhou, W.~X. Zhao, and J.-R. Wen, ``Unikgqa: Unified retrieval and reasoning for solving multi-hop question answering over knowledge graph,'' \emph{ICLR 2023}, 2023.

\bibitem{zhang2020pretrain}
Z.~Zhang, X.~Liu, Y.~Zhang, Q.~Su, X.~Sun, and B.~He, ``Pretrain-kge: learning knowledge representation from pretrained language models,'' in \emph{EMNLP Finding}, 2020, pp. 259--266.

\bibitem{wang2021kepler}
X.~Wang, T.~Gao, Z.~Zhu, Z.~Zhang, Z.~Liu, J.~Li, and J.~Tang, ``Kepler: A unified model for knowledge embedding and pre-trained language representation,'' \emph{Transactions of the Association for Computational Linguistics}, vol.~9, pp. 176--194, 2021.

\bibitem{nayyeri2022integrating}
M.~Nayyeri, Z.~Wang, M.~Akter, M.~M. Alam, M.~R. A.~H. Rony, J.~Lehmann, S.~Staab \emph{et~al.}, ``Integrating knowledge graph embedding and pretrained language models in hypercomplex spaces,'' \emph{arXiv preprint arXiv:2208.02743}, 2022.

\bibitem{huang2022endowing}
N.~Huang, Y.~R. Deshpande, Y.~Liu, H.~Alberts, K.~Cho, C.~Vania, and I.~Calixto, ``Endowing language models with multimodal knowledge graph representations,'' \emph{arXiv preprint arXiv:2206.13163}, 2022.

\bibitem{alam2022language}
M.~M. Alam, M.~R. A.~H. Rony, M.~Nayyeri, K.~Mohiuddin, M.~M. Akter, S.~Vahdati, and J.~Lehmann, ``Language model guided knowledge graph embeddings,'' \emph{IEEE Access}, vol.~10, pp. 76\,008--76\,020, 2022.

\bibitem{wang2022language}
X.~Wang, Q.~He, J.~Liang, and Y.~Xiao, ``Language models as knowledge embeddings,'' \emph{arXiv preprint arXiv:2206.12617}, 2022.

\bibitem{zhang2022reasoning}
N.~Zhang, X.~Xie, X.~Chen, S.~Deng, C.~Tan, F.~Huang, X.~Cheng, and H.~Chen, ``Reasoning through memorization: Nearest neighbor knowledge graph embeddings,'' \emph{arXiv preprint arXiv:2201.05575}, 2022.

\bibitem{lambdakg}
X.~Xie, Z.~Li, X.~Wang, Y.~Zhu, N.~Zhang, J.~Zhang, S.~Cheng, B.~Tian, S.~Deng, F.~Xiong, and H.~Chen, ``Lambdakg: A library for pre-trained language model-based knowledge graph embeddings,'' 2022.

\bibitem{MTL-KGC}
B.~Kim, T.~Hong, Y.~Ko, and J.~Seo, ``Multi-task learning for knowledge graph completion with pre-trained language models,'' in \emph{COLING}, 2020, pp. 1737--1743.

\bibitem{PKGC}
X.~Lv, Y.~Lin, Y.~Cao, L.~Hou, J.~Li, Z.~Liu, P.~Li, and J.~Zhou, ``Do pre-trained models benefit knowledge graph completion? {A} reliable evaluation and a reasonable approach,'' in \emph{ACL}, 2022, pp. 3570--3581.

\bibitem{LASS}
J.~Shen, C.~Wang, L.~Gong, and D.~Song, ``Joint language semantic and structure embedding for knowledge graph completion,'' in \emph{COLING}, 2022, pp. 1965--1978.

\bibitem{MEM-KGC}
B.~Choi, D.~Jang, and Y.~Ko, ``{MEM-KGC:} masked entity model for knowledge graph completion with pre-trained language model,'' \emph{{IEEE} Access}, vol.~9, pp. 132\,025--132\,032, 2021.

\bibitem{open-world-KGC}
B.~Choi and Y.~Ko, ``Knowledge graph extension with a pre-trained language model via unified learning method,'' \emph{Knowl. Based Syst.}, vol. 262, p. 110245, 2023.

\bibitem{StAR}
B.~Wang, T.~Shen, G.~Long, T.~Zhou, Y.~Wang, and Y.~Chang, ``Structure-augmented text representation learning for efficient knowledge graph completion,'' in \emph{WWW}, 2021, pp. 1737--1748.

\bibitem{SimKGC}
L.~Wang, W.~Zhao, Z.~Wei, and J.~Liu, ``Simkgc: Simple contrastive knowledge graph completion with pre-trained language models,'' in \emph{ACL}, 2022, pp. 4281--4294.

\bibitem{LP-BERT}
D.~Li, M.~Yi, and Y.~He, ``Lp-bert: Multi-task pre-training knowledge graph bert for link prediction,'' \emph{arXiv preprint arXiv:2201.04843}, 2022.

\bibitem{GenKGC}
X.~Xie, N.~Zhang, Z.~Li, S.~Deng, H.~Chen, F.~Xiong, M.~Chen, and H.~Chen, ``From discrimination to generation: Knowledge graph completion with generative transformer,'' in \emph{WWW}, 2022, pp. 162--165.

\bibitem{KGT5}
A.~Saxena, A.~Kochsiek, and R.~Gemulla, ``Sequence-to-sequence knowledge graph completion and question answering,'' in \emph{ACL}, 2022, pp. 2814--2828.

\bibitem{KG-S2S}
C.~Chen, Y.~Wang, B.~Li, and K.~Lam, ``Knowledge is flat: {A} seq2seq generative framework for various knowledge graph completion,'' in \emph{COLING}, 2022, pp. 4005--4017.

\bibitem{zhu2023llms}
Y.~Zhu, X.~Wang, J.~Chen, S.~Qiao, Y.~Ou, Y.~Yao, S.~Deng, H.~Chen, and N.~Zhang, ``Llms for knowledge graph construction and reasoning: Recent capabilities and future opportunities,'' \emph{arXiv preprint arXiv:2305.13168}, 2023.

\bibitem{Elmo}
M.~E. Peters, M.~Neumann, M.~Iyyer, M.~Gardner, C.~Clark, K.~Lee, and L.~Zettlemoyer, ``Deep contextualized word representations,'' in \emph{NAACL}, 2018, pp. 2227--2237.

\bibitem{generativeNER}
H.~Yan, T.~Gui, J.~Dai, Q.~Guo, Z.~Zhang, and X.~Qiu, ``A unified generative framework for various {NER} subtasks,'' in \emph{ACL}, 2021, pp. 5808--5822.

\bibitem{LDET}
Y.~Onoe and G.~Durrett, ``Learning to denoise distantly-labeled data for entity typing,'' in \emph{NAACL}, 2019, pp. 2407--2417.

\bibitem{BOX4Types}
Y.~Onoe, M.~Boratko, A.~McCallum, and G.~Durrett, ``Modeling fine-grained entity types with box embeddings,'' in \emph{ACL}, 2021, pp. 2051--2064.

\bibitem{ELQ}
B.~Z. Li, S.~Min, S.~Iyer, Y.~Mehdad, and W.~Yih, ``Efficient one-pass end-to-end entity linking for questions,'' in \emph{EMNLP}, 2020, pp. 6433--6441.

\bibitem{ReFinED}
T.~Ayoola, S.~Tyagi, J.~Fisher, C.~Christodoulopoulos, and A.~Pierleoni, ``Refined: An efficient zero-shot-capable approach to end-to-end entity linking,'' in \emph{NAACL}, 2022, pp. 209--220.

\bibitem{CR1}
M.~Joshi, O.~Levy, L.~Zettlemoyer, and D.~S. Weld, ``{BERT} for coreference resolution: Baselines and analysis,'' in \emph{EMNLP}, 2019, pp. 5802--5807.

\bibitem{SpanBERT}
M.~Joshi, D.~Chen, Y.~Liu, D.~S. Weld, L.~Zettlemoyer, and O.~Levy, ``Spanbert: Improving pre-training by representing and predicting spans,'' \emph{Trans. Assoc. Comput. Linguistics}, vol.~8, pp. 64--77, 2020.

\bibitem{CDLM}
A.~Caciularu, A.~Cohan, I.~Beltagy, M.~E. Peters, A.~Cattan, and I.~Dagan, ``{CDLM:} cross-document language modeling,'' in \emph{EMNLP}, 2021, pp. 2648--2662.

\bibitem{crossCR}
A.~Cattan, A.~Eirew, G.~Stanovsky, M.~Joshi, and I.~Dagan, ``Cross-document coreference resolution over predicted mentions,'' in \emph{ACL}, 2021, pp. 5100--5107.

\bibitem{CR-RL}
Y.~Wang, Y.~Shen, and H.~Jin, ``An end-to-end actor-critic-based neural coreference resolution system,'' in \emph{{IEEE} International Conference on Acoustics, Speech and Signal Processing, {ICASSP} 2021, Toronto, ON, Canada, June 6-11, 2021}, 2021, pp. 7848--7852.

\bibitem{sent-re1}
P.~Shi and J.~Lin, ``Simple {BERT} models for relation extraction and semantic role labeling,'' \emph{CoRR}, vol. abs/1904.05255, 2019.

\bibitem{Curriculum-RE}
S.~Park and H.~Kim, ``Improving sentence-level relation extraction through curriculum learning,'' \emph{CoRR}, vol. abs/2107.09332, 2021.

\bibitem{DREEAM}
Y.~Ma, A.~Wang, and N.~Okazaki, ``{DREEAM:} guiding attention with evidence for improving document-level relation extraction,'' in \emph{EACL}, 2023, pp. 1963--1975.

\bibitem{kumar2020building}
A.~Kumar, A.~Pandey, R.~Gadia, and M.~Mishra, ``Building knowledge graph using pre-trained language model for learning entity-aware relationships,'' in \emph{2020 IEEE International Conference on Computing, Power and Communication Technologies (GUCON)}.\hskip 1em plus 0.5em minus 0.4em\relax IEEE, 2020, pp. 310--315.

\bibitem{guo2021constructing}
Q.~Guo, Y.~Sun, G.~Liu, Z.~Wang, Z.~Ji, Y.~Shen, and X.~Wang, ``Constructing chinese historical literature knowledge graph based on bert,'' in \emph{Web Information Systems and Applications: 18th International Conference, WISA 2021, Kaifeng, China, September 24--26, 2021, Proceedings 18}.\hskip 1em plus 0.5em minus 0.4em\relax Springer, 2021, pp. 323--334.

\bibitem{melnyk2021grapher}
I.~Melnyk, P.~Dognin, and P.~Das, ``Grapher: Multi-stage knowledge graph construction using pretrained language models,'' in \emph{NeurIPS 2021 Workshop on Deep Generative Models and Downstream Applications}, 2021.

\bibitem{han2023pive}
J.~Han, N.~Collier, W.~Buntine, and E.~Shareghi, ``Pive: Prompting with iterative verification improving graph-based generative capability of llms,'' \emph{arXiv preprint arXiv:2305.12392}, 2023.

\bibitem{bosselut2019comet}
A.~Bosselut, H.~Rashkin, M.~Sap, C.~Malaviya, A.~Celikyilmaz, and Y.~Choi, ``Comet: Commonsense transformers for knowledge graph construction,'' in \emph{ACL}, 2019.

\bibitem{hao2022bertnet}
S.~Hao, B.~Tan, K.~Tang, H.~Zhang, E.~P. Xing, and Z.~Hu, ``Bertnet: Harvesting knowledge graphs from pretrained language models,'' \emph{arXiv preprint arXiv:2206.14268}, 2022.

\bibitem{west2022symbolic}
P.~West, C.~Bhagavatula, J.~Hessel, J.~Hwang, L.~Jiang, R.~Le~Bras, X.~Lu, S.~Welleck, and Y.~Choi, ``Symbolic knowledge distillation: from general language models to commonsense models,'' in \emph{NAACL}, 2022, pp. 4602--4625.

\bibitem{ribeiro-etal-2021-investigating}
L.~F.~R. Ribeiro, M.~Schmitt, H.~Sch{\"u}tze, and I.~Gurevych, ``Investigating pretrained language models for graph-to-text generation,'' in \emph{Proceedings of the 3rd Workshop on Natural Language Processing for Conversational AI}, 2021, pp. 211--227.

\bibitem{ke-etal-2021-jointgt}
P.~Ke, H.~Ji, Y.~Ran, X.~Cui, L.~Wang, L.~Song, X.~Zhu, and M.~Huang, ``{J}oint{GT}: Graph-text joint representation learning for text generation from knowledge graphs,'' in \emph{ACL Finding}, 2021, pp. 2526--2538.

\bibitem{li-etal-2021-shot-knowledge}
J.~Li, T.~Tang, W.~X. Zhao, Z.~Wei, N.~J. Yuan, and J.-R. Wen, ``Few-shot knowledge graph-to-text generation with pretrained language models,'' in \emph{ACL}, 2021, pp. 1558--1568.

\bibitem{colas-etal-2022-gap}
A.~Colas, M.~Alvandipour, and D.~Z. Wang, ``{GAP}: A graph-aware language model framework for knowledge graph-to-text generation,'' in \emph{Proceedings of the 29th International Conference on Computational Linguistics}, 2022, pp. 5755--5769.

\bibitem{jin-etal-2020-genwiki}
Z.~Jin, Q.~Guo, X.~Qiu, and Z.~Zhang, ``{G}en{W}iki: A dataset of 1.3 million content-sharing text and graphs for unsupervised graph-to-text generation,'' in \emph{Proceedings of the 28th International Conference on Computational Linguistics}, 2020, pp. 2398--2409.

\bibitem{chen-etal-2020-kgpt}
W.~Chen, Y.~Su, X.~Yan, and W.~Y. Wang, ``{KGPT}: Knowledge-grounded pre-training for data-to-text generation,'' in \emph{EMNLP}, 2020, pp. 8635--8648.

\bibitem{lukovnikov2019pretrained}
D.~Lukovnikov, A.~Fischer, and J.~Lehmann, ``Pretrained transformers for simple question answering over knowledge graphs,'' in \emph{The Semantic Web--ISWC 2019: 18th International Semantic Web Conference, Auckland, New Zealand, October 26--30, 2019, Proceedings, Part I 18}.\hskip 1em plus 0.5em minus 0.4em\relax Springer, 2019, pp. 470--486.

\bibitem{luo2020bert}
D.~Luo, J.~Su, and S.~Yu, ``A bert-based approach with relation-aware attention for knowledge base question answering,'' in \emph{IJCNN}.\hskip 1em plus 0.5em minus 0.4em\relax IEEE, 2020, pp. 1--8.

\bibitem{yasunaga-etal-2021-qa}
M.~Yasunaga, H.~Ren, A.~Bosselut, P.~Liang, and J.~Leskovec, ``{QA}-{GNN}: Reasoning with language models and knowledge graphs for question answering,'' in \emph{NAACL}, 2021, pp. 535--546.

\bibitem{hu2023empirical}
N.~Hu, Y.~Wu, G.~Qi, D.~Min, J.~Chen, J.~Z. Pan, and Z.~Ali, ``An empirical study of pre-trained language models in simple knowledge graph question answering,'' \emph{arXiv preprint arXiv:2303.10368}, 2023.

\bibitem{xu2021fusing}
Y.~Xu, C.~Zhu, R.~Xu, Y.~Liu, M.~Zeng, and X.~Huang, ``Fusing context into knowledge graph for commonsense question answering,'' in \emph{ACL}, 2021, pp. 1201--1207.

\bibitem{zhang2022drlk}
M.~Zhang, R.~Dai, M.~Dong, and T.~He, ``Drlk: Dynamic hierarchical reasoning with language model and knowledge graph for question answering,'' in \emph{EMNLP}, 2022, pp. 5123--5133.

\bibitem{hu2022empowering}
Z.~Hu, Y.~Xu, W.~Yu, S.~Wang, Z.~Yang, C.~Zhu, K.-W. Chang, and Y.~Sun, ``Empowering language models with knowledge graph reasoning for open-domain question answering,'' in \emph{EMNLP}, 2022, pp. 9562--9581.

\bibitem{zhang2022greaselm}
X.~Zhang, A.~Bosselut, M.~Yasunaga, H.~Ren, P.~Liang, C.~D. Manning, and J.~Leskovec, ``Greaselm: Graph reasoning enhanced language models,'' in \emph{ICLR}, 2022.

\bibitem{cao2022relmkg}
X.~Cao and Y.~Liu, ``Relmkg: reasoning with pre-trained language models and knowledge graphs for complex question answering,'' \emph{Applied Intelligence}, pp. 1--15, 2022.

\bibitem{huang2019knowledge}
X.~Huang, J.~Zhang, D.~Li, and P.~Li, ``Knowledge graph embedding based question answering,'' in \emph{WSDM}, 2019, pp. 105--113.

\bibitem{lin-etal-2019-kagnet}
B.~Y. Lin, X.~Chen, J.~Chen, and X.~Ren, ``{K}ag{N}et: Knowledge-aware graph networks for commonsense reasoning,'' in \emph{EMNLP-IJCNLP}, 2019, pp. 2829--2839.

\bibitem{wang2018dkn}
H.~Wang, F.~Zhang, X.~Xie, and M.~Guo, ``Dkn: Deep knowledge-aware network for news recommendation,'' in \emph{WWW}, 2018, pp. 1835--1844.

\bibitem{yang2015embedding}
B.~Yang, S.~W.-t. Yih, X.~He, J.~Gao, and L.~Deng, ``Embedding entities and relations for learning and inference in knowledge bases,'' in \emph{ICLR}, 2015.

\bibitem{xiong2018one}
W.~Xiong, M.~Yu, S.~Chang, X.~Guo, and W.~Y. Wang, ``One-shot relational learning for knowledge graphs,'' in \emph{EMNLP}, 2018, pp. 1980--1990.

\bibitem{wang2019logic}
P.~Wang, J.~Han, C.~Li, and R.~Pan, ``Logic attention based neighborhood aggregation for inductive knowledge graph embedding,'' in \emph{AAAI}, vol.~33, no.~01, 2019, pp. 7152--7159.

\bibitem{lin2015learning}
Y.~Lin, Z.~Liu, M.~Sun, Y.~Liu, and X.~Zhu, ``Learning entity and relation embeddings for knowledge graph completion,'' in \emph{Proceedings of the AAAI conference on artificial intelligence}, vol.~29, no.~1, 2015.

\bibitem{devlin2018bert}
J.~Devlin, M.-W. Chang, K.~Lee, and K.~Toutanova, ``Bert: Pre-training of deep bidirectional transformers for language understanding,'' \emph{arXiv preprint arXiv:1810.04805}, 2018.

\bibitem{CSProm-KG}
C.~Chen, Y.~Wang, A.~Sun, B.~Li, and L.~Kwok-Yan, ``Dipping plms sauce: Bridging structure and text for effective knowledge graph completion via conditional soft prompting,'' in \emph{ACL}, 2023.

\bibitem{KGC_analysis}
J.~Lovelace and C.~P. Ros{\'{e}}, ``A framework for adapting pre-trained language models to knowledge graph completion,'' in \emph{Proceedings of the 2022 Conference on Empirical Methods in Natural Language Processing, {EMNLP} 2022, Abu Dhabi, United Arab Emirates, December 7-11, 2022}, 2022, pp. 5937--5955.

\bibitem{Larger-Context-Tagging}
J.~Fu, L.~Feng, Q.~Zhang, X.~Huang, and P.~Liu, ``Larger-context tagging: When and why does it work?'' in \emph{Proceedings of the 2021 Conference of the North American Chapter of the Association for Computational Linguistics: Human Language Technologies, {NAACL-HLT} 2021, Online, June 6-11, 2021}, 2021, pp. 1463--1475.

\bibitem{Ptuning-v2}
X.~Liu, K.~Ji, Y.~Fu, Z.~Du, Z.~Yang, and J.~Tang, ``P-tuning v2: Prompt tuning can be comparable to fine-tuning universally across scales and tasks,'' \emph{CoRR}, vol. abs/2110.07602, 2021.

\bibitem{NER-as-DP}
J.~Yu, B.~Bohnet, and M.~Poesio, ``Named entity recognition as dependency parsing,'' in \emph{ACL}, 2020, pp. 6470--6476.

\bibitem{DiscontinuousNER}
F.~Li, Z.~Lin, M.~Zhang, and D.~Ji, ``A span-based model for joint overlapped and discontinuous named entity recognition,'' in \emph{ACL}, 2021, pp. 4814--4828.

\bibitem{Span-based-NER1}
C.~Tan, W.~Qiu, M.~Chen, R.~Wang, and F.~Huang, ``Boundary enhanced neural span classification for nested named entity recognition,'' in \emph{The Thirty-Fourth {AAAI} Conference on Artificial Intelligence, {AAAI} 2020, The Thirty-Second Innovative Applications of Artificial Intelligence Conference, {IAAI} 2020, The Tenth {AAAI} Symposium on Educational Advances in Artificial Intelligence, {EAAI} 2020, New York, NY, USA, February 7-12, 2020}, 2020, pp. 9016--9023.

\bibitem{Span-based-NER2}
Y.~Xu, H.~Huang, C.~Feng, and Y.~Hu, ``A supervised multi-head self-attention network for nested named entity recognition,'' in \emph{Thirty-Fifth {AAAI} Conference on Artificial Intelligence, {AAAI} 2021, Thirty-Third Conference on Innovative Applications of Artificial Intelligence, {IAAI} 2021, The Eleventh Symposium on Educational Advances in Artificial Intelligence, {EAAI} 2021, Virtual Event, February 2-9, 2021}, 2021, pp. 14\,185--14\,193.

\bibitem{Span-based-NER3}
J.~Yu, B.~Ji, S.~Li, J.~Ma, H.~Liu, and H.~Xu, ``{S-NER:} {A} concise and efficient span-based model for named entity recognition,'' \emph{Sensors}, vol.~22, no.~8, p. 2852, 2022.

\bibitem{parserNER1}
Y.~Fu, C.~Tan, M.~Chen, S.~Huang, and F.~Huang, ``Nested named entity recognition with partially-observed treecrfs,'' in \emph{AAAI}, 2021, pp. 12\,839--12\,847.

\bibitem{parserNER2}
C.~Lou, S.~Yang, and K.~Tu, ``Nested named entity recognition as latent lexicalized constituency parsing,'' in \emph{Proceedings of the 60th Annual Meeting of the Association for Computational Linguistics (Volume 1: Long Papers), {ACL} 2022, Dublin, Ireland, May 22-27, 2022}, 2022, pp. 6183--6198.

\bibitem{parserNER3}
S.~Yang and K.~Tu, ``Bottom-up constituency parsing and nested named entity recognition with pointer networks,'' in \emph{Proceedings of the 60th Annual Meeting of the Association for Computational Linguistics (Volume 1: Long Papers), {ACL} 2022, Dublin, Ireland, May 22-27, 2022}, 2022, pp. 2403--2416.

\bibitem{discontinuousNER1}
F.~Li, Z.~Lin, M.~Zhang, and D.~Ji, ``A span-based model for joint overlapped and discontinuous named entity recognition,'' in \emph{Proceedings of the 59th Annual Meeting of the Association for Computational Linguistics and the 11th International Joint Conference on Natural Language Processing, {ACL/IJCNLP} 2021, (Volume 1: Long Papers), Virtual Event, August 1-6, 2021}, 2021, pp. 4814--4828.

\bibitem{LRN}
Q.~Liu, H.~Lin, X.~Xiao, X.~Han, L.~Sun, and H.~Wu, ``Fine-grained entity typing via label reasoning,'' in \emph{Proceedings of the 2021 Conference on Empirical Methods in Natural Language Processing, {EMNLP} 2021, Virtual Event / Punta Cana, Dominican Republic, 7-11 November, 2021}, 2021, pp. 4611--4622.

\bibitem{MLMET}
H.~Dai, Y.~Song, and H.~Wang, ``Ultra-fine entity typing with weak supervision from a masked language model,'' in \emph{Proceedings of the 59th Annual Meeting of the Association for Computational Linguistics and the 11th International Joint Conference on Natural Language Processing, {ACL/IJCNLP} 2021, (Volume 1: Long Papers), Virtual Event, August 1-6, 2021}, 2021, pp. 1790--1799.

\bibitem{PL}
N.~Ding, Y.~Chen, X.~Han, G.~Xu, X.~Wang, P.~Xie, H.~Zheng, Z.~Liu, J.~Li, and H.~Kim, ``Prompt-learning for fine-grained entity typing,'' in \emph{Findings of the Association for Computational Linguistics: {EMNLP} 2022, Abu Dhabi, United Arab Emirates, December 7-11, 2022}, 2022, pp. 6888--6901.

\bibitem{DFET}
W.~Pan, W.~Wei, and F.~Zhu, ``Automatic noisy label correction for fine-grained entity typing,'' in \emph{Proceedings of the Thirty-First International Joint Conference on Artificial Intelligence, {IJCAI} 2022, Vienna, Austria, 23-29 July 2022}, 2022, pp. 4317--4323.

\bibitem{LITE}
B.~Li, W.~Yin, and M.~Chen, ``Ultra-fine entity typing with indirect supervision from natural language inference,'' \emph{Trans. Assoc. Comput. Linguistics}, vol.~10, pp. 607--622, 2022.

\bibitem{EL1}
S.~Broscheit, ``Investigating entity knowledge in {BERT} with simple neural end-to-end entity linking,'' \emph{CoRR}, vol. abs/2003.05473, 2020.

\bibitem{GENRE}
N.~D. Cao, G.~Izacard, S.~Riedel, and F.~Petroni, ``Autoregressive entity retrieval,'' in \emph{9th ICLR, {ICLR} 2021, Virtual Event, Austria, May 3-7, 2021}, 2021.

\bibitem{mGENRE}
N.~D. Cao, L.~Wu, K.~Popat, M.~Artetxe, N.~Goyal, M.~Plekhanov, L.~Zettlemoyer, N.~Cancedda, S.~Riedel, and F.~Petroni, ``Multilingual autoregressive entity linking,'' \emph{Trans. Assoc. Comput. Linguistics}, vol.~10, pp. 274--290, 2022.

\bibitem{EL2}
N.~D. Cao, W.~Aziz, and I.~Titov, ``Highly parallel autoregressive entity linking with discriminative correction,'' in \emph{Proceedings of the 2021 Conference on Empirical Methods in Natural Language Processing, {EMNLP} 2021, Virtual Event / Punta Cana, Dominican Republic, 7-11 November, 2021}, 2021, pp. 7662--7669.

\bibitem{LSTM-CR}
K.~Lee, L.~He, and L.~Zettlemoyer, ``Higher-order coreference resolution with coarse-to-fine inference,'' in \emph{NAACL}, 2018, pp. 687--692.

\bibitem{CR2}
T.~M. Lai, T.~Bui, and D.~S. Kim, ``End-to-end neural coreference resolution revisited: {A} simple yet effective baseline,'' in \emph{{IEEE} International Conference on Acoustics, Speech and Signal Processing, {ICASSP} 2022, Virtual and Singapore, 23-27 May 2022}, 2022, pp. 8147--8151.

\bibitem{CorefQA}
W.~Wu, F.~Wang, A.~Yuan, F.~Wu, and J.~Li, ``Corefqa: Coreference resolution as query-based span prediction,'' in \emph{Proceedings of the 58th Annual Meeting of the Association for Computational Linguistics, {ACL} 2020, Online, July 5-10, 2020}, 2020, pp. 6953--6963.

\bibitem{CR3}
T.~M. Lai, H.~Ji, T.~Bui, Q.~H. Tran, F.~Dernoncourt, and W.~Chang, ``A context-dependent gated module for incorporating symbolic semantics into event coreference resolution,'' in \emph{Proceedings of the 2021 Conference of the North American Chapter of the Association for Computational Linguistics: Human Language Technologies, {NAACL-HLT} 2021, Online, June 6-11, 2021}, 2021, pp. 3491--3499.

\bibitem{efficientCR1}
Y.~Kirstain, O.~Ram, and O.~Levy, ``Coreference resolution without span representations,'' in \emph{Proceedings of the 59th Annual Meeting of the Association for Computational Linguistics and the 11th International Joint Conference on Natural Language Processing, {ACL/IJCNLP} 2021, (Volume 2: Short Papers), Virtual Event, August 1-6, 2021}, 2021, pp. 14--19.

\bibitem{efficientCR2}
R.~Thirukovalluru, N.~Monath, K.~Shridhar, M.~Zaheer, M.~Sachan, and A.~McCallum, ``Scaling within document coreference to long texts,'' in \emph{Findings of the Association for Computational Linguistics: {ACL/IJCNLP} 2021, Online Event, August 1-6, 2021}, ser. Findings of {ACL}, vol. {ACL/IJCNLP} 2021, 2021, pp. 3921--3931.

\bibitem{Longformer}
I.~Beltagy, M.~E. Peters, and A.~Cohan, ``Longformer: The long-document transformer,'' \emph{CoRR}, vol. abs/2004.05150, 2020.

\bibitem{TRE}
C.~Alt, M.~H{\"{u}}bner, and L.~Hennig, ``Improving relation extraction by pre-trained language representations,'' in \emph{1st Conference on Automated Knowledge Base Construction, {AKBC} 2019, Amherst, MA, USA, May 20-22, 2019}, 2019.

\bibitem{BERT-MTB}
L.~B. Soares, N.~FitzGerald, J.~Ling, and T.~Kwiatkowski, ``Matching the blanks: Distributional similarity for relation learning,'' in \emph{ACL}, 2019, pp. 2895--2905.

\bibitem{RECENT}
S.~Lyu and H.~Chen, ``Relation classification with entity type restriction,'' in \emph{Findings of the Association for Computational Linguistics: {ACL/IJCNLP} 2021, Online Event, August 1-6, 2021}, ser. Findings of {ACL}, vol. {ACL/IJCNLP} 2021, 2021, pp. 390--395.

\bibitem{sent-re3}
J.~Zheng and Z.~Chen, ``Sentence-level relation extraction via contrastive learning with descriptive relation prompts,'' \emph{CoRR}, vol. abs/2304.04935, 2023.

\bibitem{docRE1}
H.~Wang, C.~Focke, R.~Sylvester, N.~Mishra, and W.~Y. Wang, ``Fine-tune bert for docred with two-step process,'' \emph{CoRR}, vol. abs/1909.11898, 2019.

\bibitem{HIN}
H.~Tang, Y.~Cao, Z.~Zhang, J.~Cao, F.~Fang, S.~Wang, and P.~Yin, ``{HIN:} hierarchical inference network for document-level relation extraction,'' in \emph{PAKDD}, ser. Lecture Notes in Computer Science, vol. 12084, 2020, pp. 197--209.

\bibitem{GLRE}
D.~Wang, W.~Hu, E.~Cao, and W.~Sun, ``Global-to-local neural networks for document-level relation extraction,'' in \emph{Proceedings of the 2020 Conference on Empirical Methods in Natural Language Processing, {EMNLP} 2020, Online, November 16-20, 2020}, 2020, pp. 3711--3721.

\bibitem{SIRE}
S.~Zeng, Y.~Wu, and B.~Chang, ``{SIRE:} separate intra- and inter-sentential reasoning for document-level relation extraction,'' in \emph{Findings of the Association for Computational Linguistics: {ACL/IJCNLP} 2021, Online Event, August 1-6, 2021}, ser. Findings of {ACL}, vol. {ACL/IJCNLP} 2021, 2021, pp. 524--534.

\bibitem{LSR}
G.~Nan, Z.~Guo, I.~Sekulic, and W.~Lu, ``Reasoning with latent structure refinement for document-level relation extraction,'' in \emph{ACL}, 2020, pp. 1546--1557.

\bibitem{GAIN}
S.~Zeng, R.~Xu, B.~Chang, and L.~Li, ``Double graph based reasoning for document-level relation extraction,'' in \emph{Proceedings of the 2020 Conference on Empirical Methods in Natural Language Processing, {EMNLP} 2020, Online, November 16-20, 2020}, 2020, pp. 1630--1640.

\bibitem{DocuNet}
N.~Zhang, X.~Chen, X.~Xie, S.~Deng, C.~Tan, M.~Chen, F.~Huang, L.~Si, and H.~Chen, ``Document-level relation extraction as semantic segmentation,'' in \emph{IJCAI}, 2021, pp. 3999--4006.

\bibitem{U-net}
O.~Ronneberger, P.~Fischer, and T.~Brox, ``U-net: Convolutional networks for biomedical image segmentation,'' in \emph{Medical Image Computing and Computer-Assisted Intervention - {MICCAI} 2015 - 18th International Conference Munich, Germany, October 5 - 9, 2015, Proceedings, Part {III}}, ser. Lecture Notes in Computer Science, vol. 9351, 2015, pp. 234--241.

\bibitem{ATLOP}
W.~Zhou, K.~Huang, T.~Ma, and J.~Huang, ``Document-level relation extraction with adaptive thresholding and localized context pooling,'' in \emph{AAAI}, 2021, pp. 14\,612--14\,620.

\bibitem{gardent-etal-2017-webnlg}
C.~Gardent, A.~Shimorina, S.~Narayan, and L.~Perez-Beltrachini, ``The {W}eb{NLG} challenge: Generating text from {RDF} data,'' in \emph{Proceedings of the 10th International Conference on Natural Language Generation}, 2017, pp. 124--133.

\bibitem{DBLP:conf/aaai/GuanWH19}
J.~Guan, Y.~Wang, and M.~Huang, ``Story ending generation with incremental encoding and commonsense knowledge,'' in \emph{AAAI}, 2019, pp. 6473--6480.

\bibitem{DBLP:conf/ijcai/ZhouYHZXZ18}
H.~Zhou, T.~Young, M.~Huang, H.~Zhao, J.~Xu, and X.~Zhu, ``Commonsense knowledge aware conversation generation with graph attention,'' in \emph{IJCAI}, 2018, pp. 4623--4629.

\bibitem{kale-rastogi-2020-text}
M.~Kale and A.~Rastogi, ``Text-to-text pre-training for data-to-text tasks,'' in \emph{Proceedings of the 13th International Conference on Natural Language Generation}, 2020, pp. 97--102.

\bibitem{DBLP:conf/aaai/LiuW0PY21}
Y.~Liu, Y.~Wan, L.~He, H.~Peng, and P.~S. Yu, ``{KG-BART:} knowledge graph-augmented {BART} for generative commonsense reasoning,'' in \emph{AAAI}, 2021, pp. 6418--6425.

\bibitem{mintz-etal-2009-distant}
M.~Mintz, S.~Bills, R.~Snow, and D.~Jurafsky, ``Distant supervision for relation extraction without labeled data,'' in \emph{ACL}, 2009, pp. 1003--1011.

\bibitem{saxena2020improving}
A.~Saxena, A.~Tripathi, and P.~Talukdar, ``Improving multi-hop question answering over knowledge graphs using knowledge base embeddings,'' in \emph{ACL}, 2020, pp. 4498--4507.

\bibitem{feng-etal-2020-scalable}
Y.~Feng, X.~Chen, B.~Y. Lin, P.~Wang, J.~Yan, and X.~Ren, ``Scalable multi-hop relational reasoning for knowledge-aware question answering,'' in \emph{EMNLP}, 2020, pp. 1295--1309.

\bibitem{yan2021large}
Y.~Yan, R.~Li, S.~Wang, H.~Zhang, Z.~Daoguang, F.~Zhang, W.~Wu, and W.~Xu, ``Large-scale relation learning for question answering over knowledge bases with pre-trained language models,'' in \emph{EMNLP}, 2021, pp. 3653--3660.

\bibitem{zhang2022subgraph}
J.~Zhang, X.~Zhang, J.~Yu, J.~Tang, J.~Tang, C.~Li, and H.~Chen, ``Subgraph retrieval enhanced model for multi-hop knowledge base question answering,'' in \emph{ACL (Volume 1: Long Papers)}, 2022, pp. 5773--5784.

\bibitem{jiang2023structgpt}
J.~Jiang, K.~Zhou, Z.~Dong, K.~Ye, W.~X. Zhao, and J.-R. Wen, ``Structgpt: A general framework for large language model to reason over structured data,'' \emph{arXiv preprint arXiv:2305.09645}, 2023.

\end{thebibliography}

\appendices
\section{Pros and Cons for LLMs and KGs}
\label{app:pros_and_cons}


\RE{In this section, we introduce the pros and cons of LLMs and KGs in detail. We summarize the pros and cons of LLMs and KGs in Fig. \ref{fig:LLM_vs_kg}, respectively.}

\RE{
\noindent\textbf{LLM pros.} 
\begin{itemize}
    \item \emph{General Knowledge} \cite{zhao2023survey}: LLMs pre-trained on large-scale corpora, which contain a large amount of general knowledge, such as commonsense knowledge \cite{li2022systematic} and factual knowledge \cite{petroni2019language}. Such knowledge can be distilled from LLMs and used for downstream tasks \cite{zheng2023large}.
    \item \emph{Language Processing} \cite{qiu2020pre}: LLMs have shown great performance in understanding natural language \cite{min2023recent}.  Therefore, LLMs can be used in many natural language processing tasks, such as question answering \cite{su2019generalizing}, machine translation \cite{lewis2020bart}, and text generation \cite{li2021pretrained}.
    \item \emph{Generalizability} \cite{yang2023harnessing}: LLMs enable great generalizability, which can be applied to various downstream tasks \cite{wei2021finetuned}. By providing few-shot examples \cite{brown2020language} or finetuning on multi-task data \cite{raffel2020exploring}, LLMs achieve great performance on many tasks. 
\end{itemize}
}

\RE{
\noindent\textbf{LLM cons.}
\begin{itemize}
    \item \emph{Implicit Knowledge} \cite{petroni2019language}: LLMs represent knowledge implicitly in their parameters. It is difficult to interpret or validate the knowledge obtained by LLMs.
    \item \emph{Hallucination} \cite{ji2023survey}: LLMs often experience hallucinations by generating content that while seemingly plausible but are factually incorrect \cite{zhang2023hallucination}. This problem greatly reduces the trustworthiness of LLMs in real-world scenarios.
    \item \emph{Indecisiveness} \cite{zhang2022survey}: LLMs perform reasoning by generating from a probability model, which is an indecisive process. The generated results are sampled from the probability distribution, which is difficult to control.
    \item \emph{Black-box} \cite{danilevsky2020survey}: LLMs are criticized for their lack of interpretability. It is unclear to know the specific patterns and functions LLMs use to arrive at predictions or decisions.
    \item \emph{Lacking Domain-specific/New Knowledge} \cite{wang2023robustness}: LLMs trained on general corpus might not be able to generalize well to specific domains or new knowledge due to the lack of domain-specific knowledge or new training data.
\end{itemize}
}

\RE{
\noindent\textbf{KG pros.}
\begin{itemize}
    \item \emph{Structural Knowledge} \cite{ji2021survey}: KGs store facts in a structural format (i.e., triples), which can be understandable by both humans and machines.
    \item \emph{Accuracy} \cite{vrandevcic2014wikidata}: Facts in KGs are usually manually curated or validated by experts, which are more accurate and dependable than those in LLMs.
    \item \emph{Decisiveness} \cite{hu2018state}: The factual knowledge in KGs is stored in a decisive manner. The reasoning algorithm in KGs is also deterministic, which can provide decisive results.
    \item \emph{Interpretability} \cite{zhang2021neural}: KGs are renowned for their symbolic reasoning ability, which provides an interpretable reasoning process that can be understood by humans.
    \item \emph{Domain-specific Knowledge} \cite{abu2021domain}: Many domains can construct their KGs by experts to provide precise and dependable domain-specific knowledge.
    \item \emph{Evolving Knowledge} \cite{Mitchell2015NeverEndingL}: The facts in KGs are continuously evolving. The KGs can be updated with new facts by inserting new triples and deleting outdated ones. 
\end{itemize}
}
\RE{
\noindent\textbf{KG cons.}
\begin{itemize}
    \item \emph{Incompleteness} \cite{zhong2023comprehensive}: KGs are hard to construct and often incomplete, which limits the ability of KGs to provide comprehensive knowledge.
    \item \emph{Lacking Language Understanding} \cite{bordes2013translating}: Most studies on KGs model the structure of knowledge, but ignore the textual information in KGs. The textual information in KGs is often ignored in KG-related tasks, such as KG completion \cite{yao2019kg} and KGQA \cite{jiang2023unikgqa}.
    \item \emph{Unseen Facts} \cite{luo2023normalizing}: KGs are dynamically changing, which makes it difficult to model unseen entities and represent new facts.
\end{itemize}
}
\end{document}